\documentclass[10pt]{article}
\usepackage[breakable]{tcolorbox}
\usepackage[english]{babel}
\usepackage[T2A]{fontenc}
\usepackage{paratype}
\usepackage{indentfirst}
\usepackage[default,scale=0.95]{opensans}
\usepackage{fancyhdr}
\usepackage{mathtools}
\usepackage{amssymb}
\usepackage{graphicx}

\usepackage[section]{placeins}
\usepackage[title]{appendix}
\usepackage{etoolbox}
\patchcmd{\appendices}{\quad}{. }{}{}

\usepackage[font=small,skip=5pt]{caption}

\usepackage{booktabs}
\usepackage{tabularx, ragged2e}
\usepackage{multirow}
\newcolumntype{C}{>{\Centering\arraybackslash}X}

\usepackage[colorlinks,urlcolor=blue, bookmarks=false]{hyperref}
\usepackage[a4paper,bindingoffset=0.2in,            left=1in,right=1in,top=1in,bottom=0.9in,            footskip=0.25in]{geometry}

\hypersetup{citecolor=black}
\usepackage[square,numbers]{natbib}
\let\emph\relax
\DeclareTextFontCommand{\emph}{\bfseries\em}

\usepackage{secdot}
\sectiondot{section}
\sectiondot{subsection}

\usepackage{mdframed}
\usepackage{lipsum}
\usepackage{amsthm}
\mdfsetup{linewidth=0.6pt}

\theoremstyle{definition}
\newmdtheoremenv{definition}{Definition}
\theoremstyle{theorem}
\newmdtheoremenv{proposition}{Proposition}

\newtcolorbox[auto counter]{algorithmbox}[2][]{
  breakable,
  before skip=20pt plus 2pt,after skip=20pt plus 2pt,
  colback=green!5,
  colframe=green!35!black,
  title={\bf Algorithm \thetcbcounter : #2},
  #1
}

\def\cdfeq{\mathrel{\stackrel{\rm c.d.f.}=}}
\def\cdfcoloneqq{\mathrel{\stackrel{\rm c.d.f.}\coloneqq}}
\newcommand{\R}{\mathbb{R}}
\newcommand{\E}{\mathbb{E}}
\newcommand{\Traj}{\mathcal{T}}
\newcommand{\Trans}{\mathbb{T}}
\newcommand{\A}{\mathcal{A}}
\newcommand{\St}{\mathcal{S}}
\newcommand{\Proba}{\mathcal{P}}
\newcommand{\eps}{\varepsilon}
\newcommand{\const}{\operatorname{const}}
\newcommand{\Loss}{\operatorname{Loss}}
\newcommand{\done}{\operatorname{done}}
\newcommand{\old}{\operatorname{old}}
\newcommand{\argmax}{\mathop{argmax}}

\newcommand{\KL}{\operatorname{KL}}

\begin{document}

%\begin{titlepage}
\thispagestyle{empty}

\begin{center}
\includegraphics[width=8cm, height=4cm]{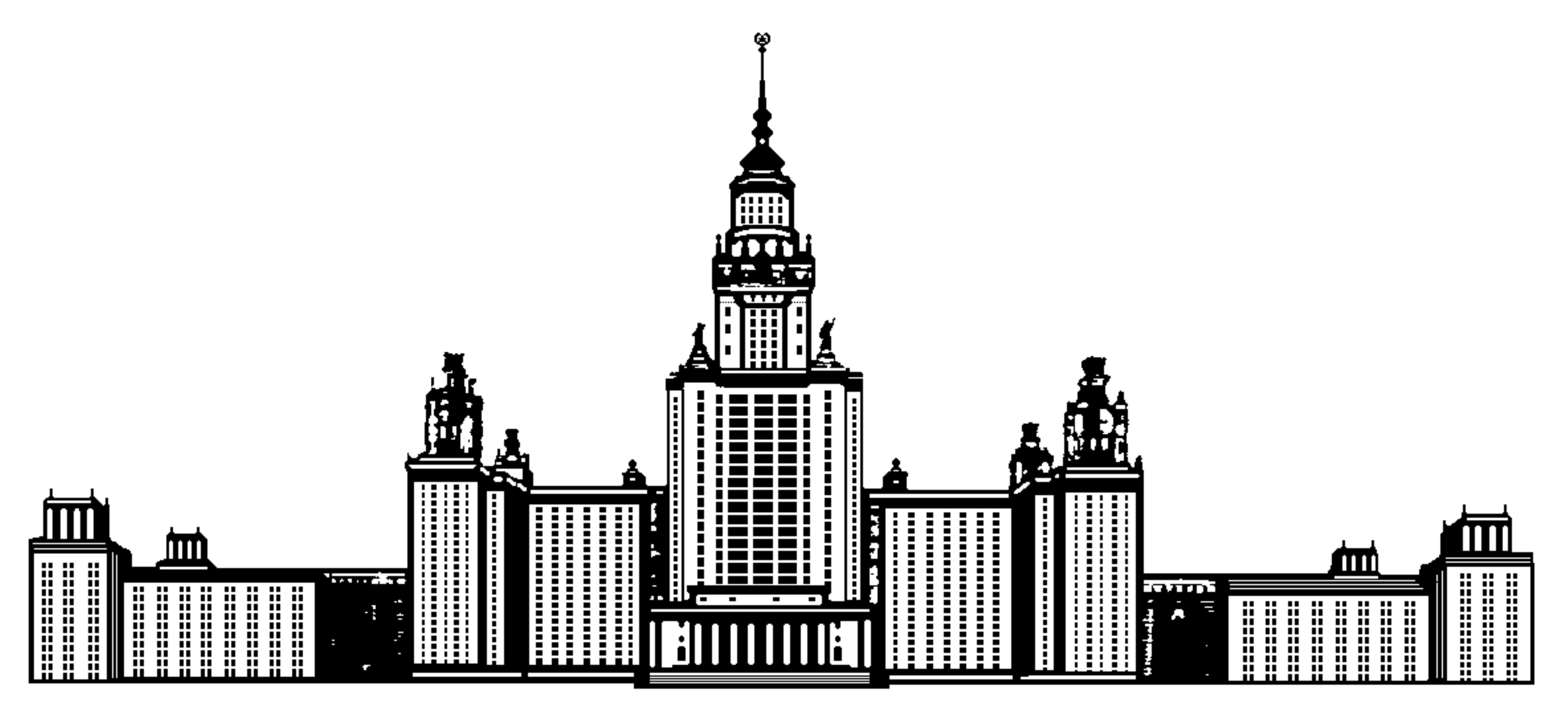}
\end{center}
\begin{center}
Moscow State University\\
\vspace{0.1 cm}
Faculty of Computational Mathematics and Cybernetics\\
\vspace{0.1 cm}
Department of Mathematical Methods of Forecasting

\vspace{3cm}

{\bf\LARGE Modern Deep Reinforcement Learning Algorithms}\\ \vspace{0.5cm}
%{\bf \Large Современные алгоритмы глубинного обучения с подкреплением} \\
%{\today}

\vspace{0.5cm}
{\bf Written by:}\\
Sergey Ivanov \\
\texttt{qbrick@mail.ru}

\vspace{0.5cm}
{\bf Scientific advisor:}\\
Alexander D'yakonov \\
\texttt{djakonov@mail.ru}

\end{center}

%\vspace{6.5cm}
\vspace{11cm}

\centerline {Moscow, 2019}

%\end{titlepage}

\newpage
\tableofcontents

\newpage
\begin{abstract}

Recent advances in Reinforcement Learning, grounded on combining classical theoretical results with Deep Learning paradigm, led to breakthroughs in many artificial intelligence tasks and gave birth to Deep Reinforcement Learning (DRL) as a field of research. In this work latest DRL algorithms are reviewed with a focus on their theoretical justification, practical limitations and observed empirical properties.

\end{abstract}

\mathversion{bold}

\newpage
\section{Introduction}

During the last several years Deep Reinforcement Learning proved to be a fruitful approach to many artificial intelligence tasks of diverse domains. Breakthrough achievements include reaching human-level performance in such complex games as Go \citep{silver2017mastering}, multiplayer Dota \citep{OpenAI_dota} and real-time strategy StarCraft II \citep{alphastarblog}. The generality of DRL framework allows its application in both discrete and continuous domains to solve tasks in robotics and simulated environments \citep{lillicrap2015continuous}.

Reinforcement Learning (RL) is usually viewed as general formalization of decision-making task and is deeply connected to dynamic programming, optimal control and game theory. \citep{sutton2018reinforcement} Yet its problem setting makes almost no assumptions about world model or its structure and usually supposes that environment is given to agent in a form of black-box. This allows to apply RL practically in all settings and forces designed algorithms to be adaptive to many kinds of challenges. Latest RL algorithms are usually reported to be transferable from one task to another with no task-specific changes and little to no hyperparameters tuning.

As an object of desire is a strategy, i.~e.~ a function mapping agent's observations to possible actions, reinforcement learning is considered to be a subfiled of machine learning. But instead of learning from data, as it is established in classical supervised and unsupervised learning problems, the agent learns from experience of interacting with environment. Being more "natural" model of learning, this setting causes new challenges, peculiar only to reinforcement learning, such as necessity of exploration integration and the problem of delayed and sparse rewards. The full setup and essential notation are introduced in section \ref{RLproblem}.

Classical Reinforcement Learning research in the last third of previous century developed an extensive theoretical core for modern algorithms to ground on. Several algorithms are known ever since and are able to solve small-scale problems when either environment states can be enumerated (and stored in the memory) or optimal policy can be searched in the space of linear or quadratic functions of state representation features. Although these restrictions are extremely limiting, foundations of classical RL theory underlie modern approaches. These theoretical fundamentals are discussed in sections \ref{TD} and \ref{pgt}--\ref{reinforce_section}.

Combining this framework with Deep Learning \citep{goodfellow2016deep} was popularized by Deep Q-Learning algorithm, introduced in \citep{mnih2013playing}, which was able to play any of 57 Atari console games without tweaking network architecture or algorithm hyperparameters. This novel approach was extensively researched and significantly improved in the following years. The principles of value-based direction in deep reinforcement learning are presented in section \ref{valuebasedDRL}.

One of the key ideas in the recent value-based DRL research is distributional approach, proposed in \citep{bellemare2017distributional}. Further extending classical theoretical foundations and coming with practical DRL algorithms, it gave birth to distributional reinforcement learning paradigm, which potential is now being actively investigated. Its ideas are described in section \ref{distributionalRL}.

Second main direction of DRL research is policy gradient methods, which attempt to directly optimize the objective function, explicitly present in the problem setup. Their application to neural networks involve a series of particular obstacles, which requested specialized optimization techniques. Today they represent a competitive and scalable approach in deep reinforcement learning due to their enormous parallelization potential and continuous domain applicability. Policy gradient methods are discussed in section \ref{policygradient}.

%Despite seemingly different nature, these two branches of RL algorithms are closely related and, in certain conditions, may be equivalent \citep{schulman2017equivalence}. Substantially alternative approaches are discussed in section \ref{discussion}. 

Despite the wide range of successes, current state-of-art DRL methods still face a number of significant drawbacks. As training of neural networks requires huge amounts of data, DRL demonstrates unsatisfying results in settings where data generation is expensive. Even in cases where interaction is nearly free (e.~g. in simulated environments), DRL algorithms tend to require excessive amounts of iterations, which raise their computational and wall-clock time cost. Furthermore, DRL suffers from random initialization and hyperparameters sensitivity, and its optimization process is known to be uncomfortably unstable \citep{irpan2018deep}. Especially embarrassing consequence of these DRL features turned out to be low reproducibility of empirical observations from different research groups \citep{henderson2018deep}. In section \ref{experiments}, we attempt to launch state-of-art DRL algorithms on several standard testbed environments and discuss practical nuances of their application. 

\newpage

\section{Reinforcement Learning problem setup}\label{RLproblem}

\subsection{Assumptions of RL setting}

Informally, the process of sequential decision-making proceeds as follows. The \emph{agent} is provided with some initial observation of environment and is required to choose some action from the given set of possibilities. The \emph{environment} responds by transitioning to another state and generating a \emph{reward signal} (scalar number), which is considered to be a ground-truth estimation of agent's performance. The process continues repeatedly with agent making choices of actions from observations and environment responding with next states and reward signals. The only goal of agent is to maximize the cumulative reward.

This description of learning process model already introduces several key assumptions. Firstly, the time space is considered to be discrete, as agent interacts with environment sequentially. Secondly, it is assumed that provided environment incorporates some reward function as supervised indicator of success. This is an embodiment of the \emph{reward hypothesis}, also referred to as \emph{Reinforcement Learning hypothesis}:

\begin{proposition}
\textbf{(Reward Hypothesis)} \citep{sutton2018reinforcement}\\
<<All of what we mean by goals and purposes can be well thought of as maximization of the expected value of the cumulative sum of a received scalar signal (reward).>>
\end{proposition}

Exploitation of this hypothesis draws a line between reinforcement learning and classical machine learning settings, supervised and unsupervised learning. Unlike unsupervised learning, RL assumes supervision, which, similar to labels in data for supervised learning, has a stochastic nature and represents a key source of knowledge. At the same time, no data or <<right answer>> is provided to training procedure, which distinguishes RL from standard supervised learning. Moreover, RL is the only machine learning task providing explicit objective function (cumulative reward signal) to maximize, while in supervised and unsupervised setting optimized loss function is usually constructed by engineer and is not <<included>> in data. The fact that reward signal is incorporated in the environment is considered to be one of the weakest points of RL paradigm, as for many real-life human goals introduction of this scalar reward signal is at the very least unobvious. 

For practical applications it is also natural to assume that agent's observations can be represented by some feature vectors, i.~e. elements of $\R^d$. The set of possible actions in most practical applications is usually uncomplicated and is either discrete (number of possible actions is finite) or can be represented as subset of $\R^m$ (almost always $[-1, 1]^m$ or can be reduced to this case)\footnote{this set is considered to be permanent for all states of environment without any loss of generality as if agent chooses invalid action the world may remain in the same state with zero or negative reward signal or stochastically select some valid action for him.}. RL algorithms are usually restricted to these two cases, but the mix of two (agent is required to choose both discrete and continuous quantities) can also be considered.   

The final assumption of RL paradigm is a \emph{Markovian property}:

\begin{proposition}
\textbf{(Markovian property)} \\
Transitions depend solely on previous state and the last chosen action and are independent of all previous interaction history.
\end{proposition}

Although this assumption may seem overly strong, it actually formalizes the fact that the world modeled by considered environment obeys some general laws. Giving that the agent knows the current state of the world and the laws, it is assumed that it is able to predict the consequences of his actions up to the internal stochasticity of these laws. In practice, both laws and complete state representation is unavailable to agent, which limits its forecasting capability. 

In the sequel we will work within the setting with one more assumption of \emph{full observability}. This simplification supposes that agent can observe complete world state, while in many real-life tasks only a part of observations is actually available. This restriction of RL theory can be removed by considering \emph{Partially observable Markov Decision Processes (PoMDP)}, which basically forces learning algorithms to have some kind of memory mechanism to store previously received observations. Further on we will stick to fully observable case. 

\subsection{Environment model}

Though the definition of \emph{Markov Decision Process (MDP)} varies from source to source, its essential meaning remains the same. The definition below utilizes several simplifications without loss of generality.\footnote{the reward function is often introduced as stochastic and dependent on action $a$, i.~e.~   $R(r \mid s, a) \colon \St \times \A \to \Proba(\R)$, while instead of fixed $s_0$ a distribution over $\St$ is given. Both extensions can be taken into account in terms of presented definition by extending the state space and incorporating all the uncertainty into transition probability $\Trans$.}

\begin{definition} 
\emph{Markov Decision Process} (MDP) is a tuple $(\St, \A, \Trans, r, s_0)$, where: 
\begin{itemize}
    \item $\St \subseteq \R^d$ --- arbitrary set, called the \emph{state space}.
    \item $\A$ --- a set, called the \emph{action space}, either
    \begin{itemize}
        \item \emph{discrete}: $|\A| < +\infty$, or
        \item \emph{continuous domain}: $\A = [-1, 1]^m$.
    \end{itemize}
    \item $\Trans$ --- \emph{transition probability} $p(s' \mid s, a)$, where $s, s' \in \St, a \in \A$.
    \item $r: \St \to \R$ --- \emph{reward function}.
    \item $s_0 \in \St$ --- starting state.
\end{itemize}
\end{definition}

It is important to notice that in the most general case the only things available for RL algorithm beforehand are $d$ (dimension of state space) and action space $\A$. The only possible way of collecting more information for agent is to interact with provided environment and observe $s_0$. It is obvious that the first choice of action $a_0$ will be probably random. While the environment responds by sampling $s_1 \sim p(s_1 \mid s_0, a_0)$, this distribution, defined in $\Trans$ and considered to be a part of MDP, may be unavailable to agent's learning procedure. What agent does observe is $s_1$ and reward signal $r_1 \coloneqq r(s_1)$ and it is the key information gathered by agent from interaction experience. 

\begin{definition} 
The tuple $\left( s_t, a_t, r_{t+1}, s_{t+1} \right)$ is called \emph{transition}. Several sequential transitions are usually referred to as \emph{roll-out}. Full track of observed quantities 
$$s_0, a_0, r_1, s_1, a_1, r_2, s_2, a_2, r_3, s_3, a_3 \dots$$ is called a \emph{trajectory}.
\end{definition}

In general case, the trajectory is infinite which means that the interaction process is neverending. However, in most practical cases the \emph{episodic property} holds, which basically means that the interaction will eventually come to some sort of an end\footnote{natural examples include the end of the game or agent's failure/success in completing some task.}. Formally, it can be simulated by the environment stucking in the last state with zero probability of transitioning to any other state and zero reward signal. Then it is convenient to reset the environment back to $s_0$ to initiate new interaction. One such interaction cycle from $s_0$ till reset, spawning one trajectory of some finite length $T$, is called an \emph{episode}. Without loss of generality, it can be considered that there exists a set of \emph{terminal states} $\St^+$, which mark the ends of interactions. By convention, transitions $(s_t, a_t, r_{t+1}, s_{t+1})$ are accompanied with binary flag $\done_{t+1} \in \{0, 1\}$, whether $s_{t+1}$ belongs to $\St^+$. As timestep $t$ at which the transition was gathered is usually of no importance, transitions are often denoted as $(s, a, r', s', \done)$ with primes marking the <<next timestep>>.

Note that the length of episode $T$ may vary between different interactions, but the episodic property holds if interaction is guaranteed to end after some finite time $T^{\max}$. If this is not the case, the task is called \emph{continuing}.

\subsection{Objective}

In reinforcement learning, the agent's goal is to maximize a cumulative reward. In episodic case, this reward can be expressed as a summation of all received reward signals during one episode and is called \emph{the return}:
\begin{equation}\label{return}
R \coloneqq \sum_{t=1}^T r_t    
\end{equation}

Note that this quantity is formally a random variable, which depends on agent's choices and the outcomes of environment transitions. As this stochasticity is an inevitable part of interaction process, the underlying distribution from which $r_t$ is sampled must be properly introduced to set rigorously the task of return maximization.

\begin{definition} 
Agent's algorithm for choosing $a$ by given current state $s$, which in general can be viewed as distribution $\pi(a \mid s)$ on domain $\A$, is called a \emph{policy} (strategy).
\end{definition}

\emph{Deterministic policy}, when the policy is represented by deterministic function $\pi \colon \St \to \A$, can be viewed as a particular case of \emph{stochastic policy} with degenerated policy $\pi(a \mid s)$, when agent's output is still a distribution with zero probability to choose an action other than $\pi(s)$. In both cases it is considered that agent sends to environment a sample $a \sim \pi(a \mid s)$.

Note that given some policy $\pi(a \mid s)$ and transition probabilities $\Trans$, the complete interaction process becomes defined from probabilistic point of view:

\begin{definition} 
For given MDP and policy $\pi$, the probability of observing $$s_0, a_0, s_1, a_1, s_2, a_2 \dots$$
is called \emph{trajectory distribution} and is denoted as $\Traj_{\pi}$:
$$\Traj_{\pi} \coloneqq \prod_{t=0} p(s_{t+1} \mid s_t, a_t)\pi(a_t \mid s_t)$$
\end{definition}

It is always substantial to keep track of what policy was used to collect certain transitions (roll-outs and episodes) during the learning procedure, as they are essentially samples from corresponding trajectory distribution. If the policy is modified in any way, the trajectory distribution changes either.

Now when a policy induces a trajectory distribution, it is possible to formulate a task of \emph{expected reward} maximization:

$$\E_{\Traj_{\pi}} \sum_{t=1}^T r_t \to \max_{\pi}$$

To ensure the finiteness of this expectation and avoid the case when agent is allowed to gather infinite reward, limit on absolute value of $r_t$ can be assumed:
$$|r_t| \le R^{\max}$$

Together with the limit on episode length $T^{\max}$ this restriction guarantees finiteness of optimal (maximal) expected reward. 

To extend this intuition to continuing tasks, the reward for each next interaction step is multiplied on some discount coefficient $\gamma \in [0,1)$, which is often introduced as part of MDP. This corresponds to the logic that with probability $1 - \gamma$ agent <<dies>> and does not gain any additional reward, which models the paradigm <<better now than later>>. In practice, this discount factor is set very close to 1.

\begin{definition} 
For given MDP and policy $\pi$ the \emph{discounted expected reward} is defined as 
$$J(\pi) \coloneqq \E_{\Traj_{\pi}} \sum_{t=0} \gamma^t r_{t+1}$$
\end{definition}

Reinforcement learning task is to find an \emph{optimal policy} $\pi^*$, which maximizes the discounted expected reward:
\begin{equation}\label{RLgoal}
    J(\pi) \to \max_{\pi}
\end{equation}

\subsection{Value functions}

Solving reinforcement learning task (\ref{RLgoal}) usually leads to a policy, that maximizes the expected reward not only for starting state $s_0$, but for any state $s \in \St$. This follows from the Markov property: the reward which is yet to be collected from some step $t$ does not depend on previous history and for agent staying at state $s$ the task of behaving optimal is equivalent to maximization of expected reward with current state $s$ as a starting state. This is the particular reason why many reinforcement learning algorithms do not seek only optimal policy, but additional information about usefulness of each state.

\begin{definition} 
For given MDP and policy $\pi$ the \emph{value function under policy $\pi$} is defined as 
$$V^\pi(s) \coloneqq \E_{\Traj_{\pi} \mid s_0 = s} \sum_{t=0} \gamma^t r_{t+1}$$
\end{definition}

This value function estimates how good it is for agent utilizing strategy $\pi$ to visit state $s$ and generalizes the notion of discounted expected reward $J(\pi)$ that corresponds to $V^{\pi}(s_0)$.

As value function can be induced by any policy, value function $V^{\pi^*}(s)$ under optimal policy $\pi^*$ can also be considered. By convention\footnote{though optimal policy may not be unique, the value functions under any optimal policy that behaves optimally from any given state (not only $s_0$) coincide. Yet, optimal policy may not know optimal behaviour for some states if it knows how to avoid them with probability 1.}, it is denoted as $V^*(s)$ and is called an \emph{optimal value function}.

Obtaining optimal value function $V^*(s)$ doesn't provide enough information to reconstruct some optimal policy $\pi^*$ due to unknown world dynamics, i.~e. transition probabilities. In other words, being blind to what state $s$ may be the environment's response on certain action in a given state makes knowing optimal value function unhelpful. This intuition suggests to introduce a similar notion comprising more information:   

\begin{definition} 
For given MDP and policy $\pi$ the \emph{quality function (Q-function) under policy $\pi$} is defined as 
$$Q^\pi(s, a) \coloneqq \E_{\Traj_{\pi} \mid s_0 = s, a_0 = a} \sum_{t=0} \gamma^t r_{t+1}$$
\end{definition}

It directly follows from the definitions that these two functions are deeply interconnected:
\begin{equation}\label{QV}
    Q^\pi(s, a) = \E_{s' \sim p(s' \mid s, a)} \left[ r(s') + \gamma V^\pi(s') \right]
\end{equation}
\begin{equation}\label{VQ}
    V^\pi(s) = \E_{a \sim \pi(a \mid s)} Q^\pi(s, a)
\end{equation}

The notion of \emph{optimal Q-function} $Q^{*}(s, a)$ can be introduced analogically. But, unlike value function, obtaining $Q^{*}(s, a)$ actually means solving a reinforcement learning task: indeed,

\begin{proposition}\label{piQ} 
If $Q^*(s, a)$ is a quality function under some optimal policy, then
$$\pi^*(s) = \argmax_a Q^*(s, a)$$
is an optimal policy.
\end{proposition}

This result implies that instead of searching for optimal policy $\pi^*$, an agent can search for optimal Q-function and derive the policy from it.

\begin{proposition} 
For any MDP existence of optimal policy leads to existence of deterministic optimal policy.
\end{proposition}

\subsection{Classes of algorithms}

Reinforcement learning algorithms are presented in a form of computational procedures specifying a strategy of collecting interaction experience and obtaining a policy with as higher $J(\pi)$ as possible. They rarely include a stopping criterion like in classic optimization methods as the stochasticity of given setting prevents any reasonable verification of optimality; usually the number of iterations to perform is determined by the amount of computational resources.  
All reinforcement learning algorithms can be roughly divided into four\footnote{in many sources evolutionary algorithms are bypassed in discussion as they do not utilize the structure of RL task in any way.} classes:
\begin{itemize}
    \item \emph{meta-heuristics}: this class of algorithms treats the task as black-box optimization with zeroth-order oracle. They usually generate a set of policies $\pi_1 \dots \pi_P$ and launch several episodes of interaction for each to determine best and worst policies according to average return. After that they try to construct more optimal policies using evolutionary or advanced random search techniques \citep{salimans2017evolution}. 
    \item \emph{policy gradient}: these algorithms directly optimize (\ref{RLgoal}), trying to obtain $\pi^*$ and no additional information about MDP, using approximate estimations of gradient with respect to policy parameters. They consider RL task as an optimization with stochastic first-order oracle and make use of interaction structure to lower the variance of gradient estimations. They will be discussed in sec. \ref{policygradient}. 
    \item \emph{value-based} algorithms construct optimal policy implicitly by gaining an approximation of optimal Q-function $Q^*(s, a)$ using dynamic programming. In DRL, Q-function is represented with neural network and an approximate dynamic programming is performed using reduction to supervised learning. This framework will be discussed in sec. \ref{valuebasedDRL} and \ref{distributionalRL}.
    \item \emph{model-based} algorithms exploit learned or given world dynamics, i.~e.~ distributions $p(s' \mid s, a)$ from $\Trans$. The class of algorithms to work with when the model is explicitly provided is represented by such algorithms as Monte-Carlo Tree Search; if not, it is possible to imitate the world dynamics by learning the outputs of black box from interaction experience \citep{kaiser2019model}.
\end{itemize}

\subsection{Measurements of performance}

Achieved performance (\emph{score}) from the point of average cumulative reward is not the only one measure of RL algorithm quality. When speaking of real-life robots, the required number of simulated episodes is always the biggest concern. It is usually measured in terms of interaction steps (where step is one transition performed by environment) and is referred to as \emph{sample efficiency}.

When the simulation is more or less cheap, RL algorithms can be viewed as a special kind of optimization procedures. In this case, the final performance of the found policy is opposed to required computational resources, measured by \emph{wall-clock time}. In most cases RL algorithms can be expected to find better policy after more iterations, but the amount of these iterations tend to be unjustified. 

The ratio between amount of interactions and required wall-clock time for one update of policy varies significantly for different algorithms. It is well-known that model-based algorithms tend to have the greatest sample-efficiency at the cost of expensive update iterations, while evolutionary algorithms require excessive amounts of interactions while providing massive resources for parallelization and reduction of wall-clock time. Value-based and policy gradient algorithms, which will be the focus of our further discussion, are known to lie somewhere in between.     

\newpage

\section{Value-based algorithms}\label{valuebasedDRL}

\subsection{Temporal Difference learning}\label{TD}

In this section we consider temporal difference learning algorithm \citep[Chapter~6]{sutton2018reinforcement}, which is a classical Reinforcement Learning method in the base of modern value-based approach in DRL.

The first idea behind this algorithm is to search for optimal Q-function $Q^*(s, a)$ by solving a system of recursive equations which can be derived by recalling interconnection between Q-function and value function (\ref{QV}):

\begin{align*}
    Q^\pi(s, a) &= \E_{s' \sim p(s' \mid s, a)} \left[ r(s') + \gamma V^\pi(s') \right] = \\
    = \{ \text{using (\ref{VQ})} \} &= \E_{s' \sim p(s' \mid s, a)} \left[ r(s') + \gamma \E_{a' \sim \pi(a' \mid s')} Q^\pi(s', a') \right]
\end{align*}

This equation, named \emph{Bellman equation}, remains true for value functions under any policies including optimal policy $\pi^*$:

\begin{equation}\label{QQ}
    Q^*(s, a) = \E_{s' \sim p(s' \mid s, a)} \left[ r(s') + \gamma \E_{a' \sim \pi(a' \mid s')} Q^*(s', a') \right]
\end{equation}

Recalling proposition \ref{piQ}, optimal (deterministic) policy can be represented as $\pi^*(s) = \argmax\limits_a \allowbreak Q^*(s, a)$. Substituting this for $\pi^*(s)$ in (\ref{QQ}), we obtain fundamental \emph{Bellman optimality equation}:

\begin{proposition}
\textbf{(Bellman optimality equation)}
\begin{equation}\label{Bellman}
 Q^*(s, a) = \E_{s' \sim p(s' \mid s, a)} \left[ r(s') + \gamma \max_{a'} Q^*(s', a') \right]   
\end{equation}
\end{proposition}

The straightforward utilization of this result is as follows. Consider the \emph{tabular case}, when both state space $\St$ and action space $\A$ are finite (and small enough to be listed in computer memory). Let us also assume for now that transition probabilities are available to training procedure. Then $Q^*(s, a): \St \times \A \to \R$ can be represented as a finite table with $|\St| |\A|$ numbers. In this case (\ref{Bellman}) just gives a set of $|\St| |\A|$ equations for this table to satisfy.

Addressing the values of the table as unknown variables, this system of equations can be solved using basic \emph{point iteration} method: let $Q^*_0(s, a)$ be initial arbitrary values of table (with the only exception that for terminal states $s \in \St^+$, if any, $Q^*_0(s, a) = 0$ for all actions $a$). On each iteration $t$ the table is updated by substituting current values of the table to the right side of equation until the process converges:
\begin{equation}\label{bellmanupdate}
Q^*_{t+1}(s, a) = \E_{s' \sim p(s' \mid s, a)} \left[ r(s') + \gamma \max_{a'} Q^*_{t}(s', a') \right]
\end{equation}

This straightforward approach of learning the optimal Q-function, named \emph{Q-learning}, has been extensively studied in classical Reinforcement Learning. One of the central results is presented in the following convergence theorem:

\begin{proposition}\label{BasicQlearningConvergence}
Let by $\mathcal{B}$ denote an operator $(\St \times \A \to \R) \to (\St \times \A \to \R)$, updating $Q^*_t$ as in (\ref{bellmanupdate}):
$$Q^*_{t+1} = \mathcal{B}Q^*_t$$
for all state-action pairs $s, a$.

Then $\mathcal{B}$ is a \emph{contraction mapping}, i.~.e. for any two tables $Q_1, Q_2 \in (\St \times \A \to \R)$
$$\|\mathcal{B}Q_1 - \mathcal{B}Q_2\|_{\infty} \le \gamma \|Q_1 - Q_2\|_{\infty}$$
Therefore, there is a unique fixed point of the system of equations (\ref{bellmanupdate}) and the point iteration method converges to it.
\end{proposition}

The contraction mapping property is actually of high importance. It demonstrates that the point iteration algorithm converges with exponential speed and requires small amount of iterations. As the true $Q^*$ is a fixed point of (\ref{Bellman}), the algorithm is guaranteed to yield a correct answer. The trick is that each iteration demands full pass across all state-action pairs and exact computation of expectations over transition probabilities.

In general case, these expectations can't be explicitly computed. Instead, agent is restricted to samples from transition probabilities gained during some interaction experience. \emph{Temporal Difference} (TD)\footnote{also known as TD(0) due to theoretical generalizations} algorithm proposes to collect this data using $\pi_t = \argmax\limits_a Q_t^*(s, a) \approx \pi^*$ and after each gathered transition $(s_t, a_t, r_{t+1}, s_{t+1})$ update only one cell of the table:
\begin{equation}\label{Qlearningupdate}
Q^*_{t+1}(s, a) = \begin{cases}
(1 - \alpha_t)Q^*_{t}(s, a) + \alpha_t\left[ r_{t+1} + \gamma \max\limits_{a'} Q^*_{t}(s_{t+1}, a') \right] \quad &\text{if\,} s = s_t, a = a_t \\
Q^*_t(s, a) \quad &\text{else}
\end{cases}
\end{equation}
where $\alpha_t \in (0, 1)$ plays the role of exponential smoothing parameter for estimating expectation $\E_{s' \sim p(s' \mid s_t, a_t)} ( \cdot )$ from samples.

Two key ideas are introduced in the update formula (\ref{Qlearningupdate}): exponential smoothing instead of exact expectation computation and cell by cell updates instead of updating full table at once. Both are required to settle Q-learning algorithm for online application.

As the set $\St^+$ of terminal states in online setting is usually unknown beforehand, a slight modification of update (\ref{Qlearningupdate}) is used. If observed next state $s'$ turns out to be terminal (recall the convention to denote this by flag $\done$), its value function is known to be equal to zero: 
$$V^*(s') = \max\limits_{a'} Q^*(s', a') = 0$$

This knowledge is embedded in the update rule (\ref{Qlearningupdate}) by multiplying $\max\limits_{a'} Q^*_{t}(s_{t+1}, a')$ on $(1 - \done_{t+1})$. For the sake of shortness, this factor is often omitted but should be always present in implementations.

Second important note about formula (\ref{Qlearningupdate}) is that it can be rewritten in the following equivalent way:

\begin{equation}\label{Qlearningupdate2}
Q^*_{t+1}(s, a) = \begin{cases}
Q^*_{t}(s, a) + \alpha_t\left[ r_{t+1} + \gamma \max\limits_{a'} Q^*_{t}(s_{t+1}, a') - Q^*_{t}(s, a) \right] \quad &\text{if\,} s = s_t, a = a_t \\
Q^*_t(s, a) \quad &\text{else}
\end{cases}
\end{equation}

The expression in the brackets, referred to as \emph{temporal difference}, represents a difference between Q-value $Q^*_{t}(s, a)$ and its one-step approximation $r_{t+1} + \gamma \max\limits_{a'} Q^*_{t}(s_{t+1}, a')$, which must be zero in expectation for true optimal Q-function.

The idea of exponential smoothing allows us to formulate first practical algorithm which can work in the tabular case with unknown world dynamics:

\begin{algorithmbox}[label = TDalgorithm]{Temporal Difference algorithm}
\textbf{Hyperparameters:} $\alpha_t \in (0, 1)$

\vspace{0.3cm}
Initialize $Q^*(s, a)$ arbitrary \\
\textbf{On each interaction step:}
\begin{enumerate}
    \item select $a = \argmax\limits_a Q^*(s, a)$
    \item observe transition $(s, a, r', s', \done)$
    \item update table:
    $$Q^*(s, a) \gets Q^*(s, a) + \alpha_t\left[ r' + (1 - \done)\gamma \max_{a'} Q^*(s', a') - Q^*(s, a)\right]$$
\end{enumerate}
\end{algorithmbox}

It turns out that under several assumptions on state visitation during interaction process this procedure holds similar properties in terms of convergence guarantees, which are stated by the following theorem: 

\begin{proposition}\label{TDconvergence} \citep{watkins1992q}
Let's define
$$e_t(s, a) = \begin{cases}
\alpha_t \quad (s, a) \, \text{is updated on step } t \\
0 \quad \text{otherwise}
\end{cases}$$

Then if for every state-action pair $(s, a)$
$$\sum_t^{+\infty} e_t(s, a) = \infty \quad \sum_t^{+\infty} e_t(s, a)^2 < \infty$$

the algorithm \ref{TDalgorithm} converges to optimal $Q^*$ with probability 1.
\end{proposition}

This theorem states that basic policy iteration method can be actually applied online in the way proposed by TD algorithm, but demands <<enough exploration>> from the strategy of interacting with MDP during training. Satisfying this demand remains a unique and common problem of reinforcement learning.

The widespread kludge is \emph{$\eps$-greedy strategy} which basically suggests to choose random action instead of $a = \argmax\limits_a Q^*(s, a)$ with probability $\eps_t$. The probability $\eps_t$ is usually set close to 1 during first interaction iterations and scheduled to decrease to a constant close to 0. This heuristic makes agent visit all states with non-zero probabilities independent of what current approximation $Q^*(s, a)$ suggests. 

The main practical issue with Temporal Difference algorithm is that it requires table $Q^*(s,a)$ to be explicitly stored in memory, which is impossible for MDP with high state space complexity. This limitation substantially restricted its applicability until its combination with deep neural network was proposed.

\subsection{Deep Q-learning (DQN)}\label{DQN}

Utilization of neural nets to model either a policy or a Q-function frees from constructing task-specific features and opens possibilities of applying RL algorithms to complex tasks, e.~g. tasks with images as input. Video games are classical example of such tasks where raw pixels of screen are provided as state representation and, correspondingly, as input to either policy or Q-function. 

Main idea of Deep Q-learning \cite{mnih2013playing} is to adapt Temporal Difference algorithm so that update formula (\ref{Qlearningupdate2}) would be equivalent to gradient descent step for training a neural network to solve a certain regression task. Indeed, it can be noticed that the exponential smoothing parameter $\alpha_t$ resembles learning rate of first-order gradient optimization procedures, while the exploration conditions from theorem \ref{TDconvergence} look identical to restrictions on learning rate of stochastic gradient descent.

The key hint is that (\ref{Qlearningupdate2}) is actually a gradient descent step in the parameter space of the table functions family:
$$Q^*(s, a, \theta) = \theta^{s, a}$$
where all $\theta^{s, a}$ form a vector of parameters $\theta \in \R^{|\St| |\A|}$.

To unravel this fact, it is convenient to introduce some notation from regression tasks. First, let's denote by $y$ the \emph{target} of our regression task, i.~e. the quantity that our model is trying to predict:
\begin{equation}\label{target}
y(s, a) \coloneqq r(s') + \gamma \max\limits_{a'} Q^*(s', a', \theta)
\end{equation}
where $s'$ is a sample from $p(s' \mid s, a)$ and $s, a$ is input data. In this notation (\ref{Qlearningupdate2}) is equivalent to:
$$\theta_{t+1} = \theta_t + \alpha_t\left[ y(s, a) - Q^*(s, a, \theta_t) \right]e^{s, a}$$
where we multiplied scalar value $\alpha_t\left[ y(s, a) - Q^*(s, a, \theta_t) \right]$ on the following vector $e^{s, a}$
\begin{equation*}
e^{s, a}_{i, j} \coloneqq \begin{cases}
1 \qquad (i, j) = (s, a) \\
0 \qquad (i, j) \ne (s, a)
\end{cases}
\end{equation*}
to formulate an update of only one component of $\theta$ in a vector form. By this we transitioned to update in parameter space using $Q^*(s, a, \theta) = \theta^{s, a}$. Remark that for table functions family the derivative of $Q^*(s, a, \theta)$ by $\theta$ for given input $s, a$ is its one-hot encoding, i.~e. exactly $e^{s, a}$:
\begin{equation}\label{tableQfuncgrad}
\frac{\partial Q^* (s, a, \theta)}{\partial \theta} = e^{s, a}
\end{equation}

The statement now is that this formula is a gradient descent update for regression with input $s, a$, target $y(s, a)$ and MSE loss function:
\begin{equation}\label{mse}
\Loss(y(s, a), Q^*(s, a, \theta_t)) = \left( Q^*(s, a, \theta_t) - y(s, a) \right) ^2
\end{equation}

Indeed:
\begin{align*}
\theta_{t+1} &= \theta_t + \alpha_t\left[ y(s, a) - Q^*(s, a, \theta_t) \right]e^{s, a} = \\
\{ \text{(\ref{mse})} \} &= \theta_t - \alpha_t \frac{\partial \Loss(y, Q^*(s, a, \theta_t))}{\partial Q^*} e^{s, a} \\
\{ \text{(\ref{tableQfuncgrad})} \} &= \theta_t - \alpha_t \frac{\partial \Loss(y, Q^*(s, a, \theta_t))}{\partial Q^*} \frac{\partial Q^* (s, a, \theta_t)}{\partial \theta} = \\
\{ \text{chain rule} \}&= \theta_t - \alpha_t \frac{\partial \Loss(y, Q^*(s, a, \theta_t))}{\partial \theta}
\end{align*}

The obtained result is evidently a gradient descent step formula to minimize MSE loss function with target (\ref{target}):

\begin{equation}\label{DQNupdate}
    \theta_{t+1} = \theta_t - \alpha_t \frac{\partial \Loss(y, Q^*(s, a, \theta_t))}{\partial \theta}
\end{equation}

It is important that dependence of $y$ from $\theta$ is ignored during gradient computation (otherwise the chain rule application with $y$ being dependent on $\theta$ is incorrect). On each step of temporal difference algorithm new target $y$ is constructed using current Q-function approximation, and a new regression task with this target is set. For this fixed target one MSE optimization step is done according to (\ref{DQNupdate}), and on the next step a new regression task is defined. Though during each step the target is considered to represent some ground truth like it is in supervised learning, here it provides a direction of optimization and because of this reason is sometimes called a \emph{guess}.  

Notice that representation (\ref{DQNupdate}) is equivalent to standard TD update (\ref{Qlearningupdate2}) with all theoretical results remaining while the parametric family $Q(s, a, \theta)$ is a table functions family. At the same time, (\ref{DQNupdate}) can be formally applied to any parametric function family including neural networks. It must be taken into account that this transition is not rigorous and all theoretical guarantees provided by theorem \ref{TDconvergence} are lost at this moment.

Further on we assume that optimal Q-function is approximated with neural network $Q^*_{\theta}(s, a)$ with parameters $\theta$. Note that for discrete action space case this network may take only $s$ as input and output $|\A|$ numbers representing $Q^*_{\theta}(s, a_1) \dots Q^*_{\theta}(s, a_{|\A|})$, which allows to find an optimal action in a given state $s$ with a single forward pass through the net. Therefore target $y$ for given transition $(s, a, r', s', \done)$ can be computed with one forward pass and optimization step can be performed in one more forward\footnote{in implementations it is possible to combine $s$ and $s'$ in one batch and perform these two forward passes <<at once>>.} and one backward pass. 

Small issue with this straightforward approach is that, of course, it is impractical to train neural networks with batches of size 1. In \citep{mnih2013playing} it is proposed to use \emph{experience replay} to store all collected transitions $(s, a, r', s', \done)$ as data samples and on each iteration sample a batch of standard for neural networks training size. As usual, the loss function is assumed to be an average of losses for each transition from the batch. This utilization of previously experienced transitions is legit because TD algorithm is known to be an \emph{off-policy} algorithm, which means it can work with arbitrary transitions gathered by any agent's interaction experience. One more important benefit from experience replay is \emph{sample decorrelation} as  consecutive transitions from interaction are often similar to each other since agent usually locates at the particular part of MDP. 

Though empirical results of described algorithm turned out to be promising, the behaviour of $Q^*_{\theta}$ values indicated the instability of learning process. Reconstruction of target after each optimization step led to so-called \emph{compound error} when approximation error propagated from the close-to-terminal states to the starting in avalanche manner and could lead to guess being $10^6$ and more times bigger than the true $Q^*$ value. To address this problem, \citep{mnih2013playing} introduced a kludge known as \emph{target network}, which basic idea is to solve fixed regression problem for $K > 1$ steps, i.~.e. recompute target every $K$-th step instead of each.

To avoid target recomputation for the whole experience replay, the copy of neural network $Q^*_{\theta}$ is stored, called the target network. Its architecture is the same while weights $\theta^{-}$ are a copy of $Q^*_{\theta}$ from the moment of last target recomputation\footnote{alternative, but more computationally expensive option, is to update target network weights on each step using exponential smoothing} and its main purpose is to generate targets $y$ for given current batch.

Combining all things together and adding $\eps$-greedy strategy to facilitate exploration, we obtain classic DQN algorithm:

\begin{algorithmbox}[label = DQNalgorithm]{Deep Q-learning (DQN)}
\textbf{Hyperparameters:} $B$ --- batch size, $K$ --- target network update frequency, $\eps(t) \in (0, 1]$ --- greedy exploration parameter, $Q^*_\theta$ --- neural network, SGD optimizer.

\vspace{0.3cm}
Initialize weights of $\theta$ arbitrary \\
Initialize $\theta^- \gets \theta$ \\
\textbf{On each interaction step:}
\begin{enumerate}
    \item select $a$ randomly with probability $\varepsilon(t)$, else $a = \argmax\limits_a Q^*_\theta(s, a)$
    \item observe transition $(s, a, r', s', \done)$
    \item add observed transition to experience replay
    \item sample batch of size $B$ from experience replay
    \item for each transition $T$ from the batch compute target:
    $$y(T) = r(s') + \gamma \max\limits_{a'} Q^*(s', a', \theta^-)$$
    \item compute loss:
    $$\Loss = \frac{1}{B}\sum_T \left( Q^*(s, a, \theta) - y(T) \right) ^2$$
    \item make a step of gradient descent using $\frac{\partial \Loss }{\partial \theta}$
    \item if $t \mod K = 0$: $\theta^- \gets \theta$
\end{enumerate}
\end{algorithmbox}

\subsection{Double DQN}\label{DDQN}

Although target network successfully prevented $Q^*_\theta$ from unbounded growth and empirically stabilized learning process, the values of $Q^*_\theta$ on many domains were evident to tend to overestimation. The problem is presumed to reside in max operation in target construction formula (\ref{target}): 
$$y = r(s') + \gamma \max\limits_{a'} Q^*(s', a', \theta^-)$$
During this estimation $\max$ shifts Q-value estimation towards either to those actions that led to high reward due to luck or to the actions with overestimating approximation error. 

The solution proposed in \citep{van2016deep} is based on idea of separating \emph{action selection} and \emph{action evaluation} to carry out each of these operations using its own approximation of $Q^*$:
\begin{align*}
\max\limits_{a'} Q^*(s', a', \theta^-) = Q^*(s', \argmax\limits_{a'} Q^*(s', a', \theta^-), \theta^-) \approx \\
\approx Q^*(s', \argmax\limits_{a'} Q^*(s', a', \theta^-_1), \theta^-_2)
\end{align*}

The simplest, but expensive, implementation of this idea is to run two independent DQN (<<Twin DQN>>) algorithms and use the twin network to evaluate actions:
$$y_1 = r(s') + \gamma Q_1^*(s', \argmax\limits_{a'} Q_2^*(s', a', \theta^-_2), \theta^-_1)$$
$$y_2 = r(s') + \gamma Q_2^*(s', \argmax\limits_{a'} Q_1^*(s', a', \theta^-_1), \theta^-_2)$$
Intuitively, each Q-function here may prefer lucky or overestimated actions, but the other Q-function judges them according to its own luck and approximation error, which may be as underestimating as overestimating. Ideally these two DQNs should not share interaction experience to achieve that, which makes such algorithm twice as expensive both in terms of computational cost and sample efficiency.

Double DQN \citep{van2016deep} is more compromised option which suggests to use current weights of network $\theta$ for action selection and target network weights $\theta^-$ for action evaluation, assuming that when the target network update frequency $K$ is big enough these two networks are sufficiently different:
$$y = r(s') + \gamma Q^*(s', \argmax\limits_{a'} Q^*(s', a', \theta), \theta^-)$$

\subsection{Dueling DQN}\label{DueDQN}

Another issue with DQN algorithm \ref{DQNalgorithm} emerges when a huge part of considered MDP consists of states of low optimal value $V^*(s)$, which is an often case. The problem is that when the agent visits unpromising state instead of lowering its value $V^*(s)$ it remembers only low pay-off for performing some action $a$ in it by updating $Q^*(s, a)$. This leads to regular returns to this state during future interactions until all actions prove to be unpromising and all $Q^*(s, a)$ are updated. The problem gets worse when the cardinality of action space is high or there are many similar actions in action space. 

One benefit of deep reinforcement learning is that we are able to facilitate generalization across actions by specifying the architecture of neural network. To do so, we need to encourage the learning of $V^*(s)$ from updates of $Q^*(s, a)$. The idea of \emph{dueling architecture} \citep{wang2015dueling} is to incorporate approximation of $V^*(s)$ explicitly in computational graph. For that purpose we need the definition of advantage function:

\begin{definition} 
For given MDP and policy $\pi$ the \emph{advantage function under policy $\pi$} is defined as 
\begin{equation}\label{advantage}
A^\pi(s, a) \coloneqq Q^\pi(s, a) - V^\pi(s)
\end{equation}
\end{definition}

Advantage function is evidently interconnected with Q-function and value function and actually shows the relative advantage of selecting action $a$ comparing to average performance of the policy. If for some state $A^\pi(s, a) > 0$, then modifying $\pi$ to select $a$ more often in this particular state will lead to better policy as its average return will become bigger than initial $V^\pi(s)$. This follows from the following property of arbitrary advantage function:
\begin{align}\label{advantagerestriction}
\begin{split}
\mathbb{E}_{a \sim \pi(a \mid s)}A^\pi(s, a) &= \mathbb{E}_{a \sim \pi(a \mid s)} \left[ Q^\pi(s, a) - V^\pi(s) \right] = \\
&= \mathbb{E}_{a \sim \pi(a \mid s)} Q^\pi(s, a) - V^\pi(s) = \\
\{ \text{using (\ref{VQ})} \}&= V^\pi(s) - V^\pi(s) = 0
\end{split}
\end{align}

Definition of optimal advantage function $A^*(s, a)$ is analogous and allows us to reformulate $Q^*(s, a)$ in terms of $V^*(s)$ and $A^*(s, a)$:
\begin{equation}\label{naivedueling}
Q^*(s, a) = V^*(s) + A^*(s, a)
\end{equation}

Straightforward utilization of this decomposition is following: after several feature extracting layers the network is joined with two heads, one outputting single scalar $V^*(s)$ and one outputting $|\A|$ numbers $A^*(s, a)$ like it was done in DQN for Q-function. After that this scalar value estimation is added to all components of $A^*(s, a)$ in order to obtain $Q^*(s, a)$ according to (\ref{naivedueling}). The problem with this naive approach is that due to (\ref{advantagerestriction}) advantage function can not be arbitrary and must hold the property (\ref{advantagerestriction}) for $Q^*(s, a)$ to be identifiable.

This restriction (\ref{advantagerestriction}) on advantage function can be simplified for the case when optimal policy is induced by optimal Q-function:
\begin{align*}
\begin{split}
0 &= \mathbb{E}_{a \sim \pi^*(a \mid s)} Q^*(s, a) - V^*(s) = \\ 
&= Q^*(s, \argmax_a Q^*(s, a)) - V^*(s) = \\
&= \max_a Q^*(s, a) - V^*(s) = \\
&= \max_a \left[ Q^*(s, a) - V^*(s)\right] = \\
&= \max_a A^*(s, a)
\end{split}
\end{align*}

This condition can be easily satisfied in computational graph by subtracting $\max\limits_a A^*(s, a)$ from advantage head. This will be equivalent to the following formula of dueling DQN:
\begin{equation}\label{dueling}
Q^*(s, a) = V^*(s) + A^*(s, a) - \max_a A^*(s, a)    
\end{equation}

%Nothing else except network architecture is changed in dueling DQN model.

The interesting nuance of this improvement is that after evaluation on Atari-57 authors discovered that substituting max operation in (\ref{dueling}) with averaging across actions led to better results (while usage of unidentifiable formula (\ref{naivedueling}) led to poor performance). Although gradients can be backpropagated through both operation and formula (\ref{dueling}) seems theoretically justified, in practical implementations averaging instead of maximum is widespread.

\subsection{Noisy DQN}\label{NoisyDQN}

By default, DQN algorithm does not concern the exploration problem and is always augmented with $\eps$-greedy strategy to force agent to discover new states. This baseline exploration strategy suffers from being extremely hyperparameter-sensitive as early decrease of $\eps(t)$ to close to zero values may lead to stucking in local optima, when agent is unable to explore new options due to imperfect $Q^*$, while high values of $\eps(t)$ force agent to behave randomly for excessive amount of episodes, which slows down learning. In other words, $\eps$-greedy strategy transfers responsibility to solve \emph{exploration-exploitation} trade-off on engineer.

The key reason why $\eps$-greedy exploration strategy is relatively primitive is that exploration priority does not depend on current state. Intuitively, the choice whether to exploit knowledge by selecting approximately optimal action or to explore MDP by selecting some other depends on how explored the current state $s$ is. Discovering a new part of state space after any amount of interaction probably indicates that random actions are good to try there, while close-to-initial states will probably be sufficiently explored after several first episodes. 

In $\eps$-greedy strategy agent selects action using deterministic $Q^*(s, a, \theta)$ and only afterwards injects state-independent noise in a form of $\eps(t)$ probability of choosing random action. \emph{Noisy networks} \citep{fortunato2017noisy} were proposed as a simple extension of DQN to provide state-dependent and parameter-free exploration by injecting noise of trainable volume to all (or most\footnote{usually it is not injected in very first layers responsible for feature extraction like convolutional layers in networks for images as input.}) nodes in computational graph.

Let a linear layer with $m$ inputs and $n$ outputs in q-network perform the following computation:
$$y(x) = Wx + b$$
where $x \in \R^m$ is input, $W \in \R^{n \times m}$ --- weights matrix, $b \in \R^m$ --- bias. In noisy layers it is proposed to substitute deterministic parameters with samples from $\mathcal{N}(\mu, \sigma)$ where $\mu, \sigma$ are trained with gradient descent\footnote{using standard reparametrization trick}. On the forward pass through the noisy layer we sample $\varepsilon_W \sim \mathcal{N}(0, I_{nm \times nm}), \varepsilon_b \sim \mathcal{N}(0, I_{n \times n})$ and then compute
\begin{align*}
W &= (\mu_W + \sigma_W \odot \varepsilon_W)\\
b &= (\mu_b + \sigma_b \odot \varepsilon_b)\\
y(x) &= Wx + b   
\end{align*}
where $\odot$ denotes element-wise multiplication, $\mu_W, \sigma_W \in \R^{n \times m}, \mu_b, \sigma_b \in \R^n$ --- trainable parameters of the layer. Note that the number of parameters for such layers is doubled comparing to ordinary layers.

As the output of q-network now becomes a random variable, loss value becomes a random variable too. Like in similar models for supervised learning, on each step an expectation of loss function over noise is minimized:
$$\mathbb{E}_\varepsilon \Loss(\theta, \varepsilon) \to \min_\theta$$
The gradient in this setting can be estimated using Monte-Carlo:
$$\nabla_\theta \mathbb{E}_\varepsilon \Loss(\theta, \varepsilon) = \mathbb{E}_\varepsilon \nabla_\theta \Loss(\theta, \varepsilon) \approx \nabla_\theta \Loss(\theta, \varepsilon) \quad \varepsilon \sim \mathcal{N}(0, I)$$

It can be seen that amount of noise actually inflicting output of network may vary for different inputs, i.~e. for different states. There are no guarantees that this amount will reduce as the interaction proceeds; the behaviour of average magnitude of noise injected in the network with time is reported to be extremely sensitive to initialization of $\sigma_W, \sigma_b$ and vary from MDP to MDP.

One technical issue with noisy layers is that on each pass an excessive amount (by the number of network parameters) of noise samples is required. This may substantially reduce computational efficiency of forward pass through the network. For optimization purposes it is proposed to obtain noise for weights matrices in the following way: sample just $n + m$ noise samples $\varepsilon^1_W \sim \mathcal{N}(0, I_{m \times m}), \varepsilon^2_W \sim \mathcal{N}(0, I_{n \times n})$ and acquire matrix noise in a factorized form:
$$\varepsilon_W = f(\varepsilon^1_W) f(\varepsilon^2_W)^T$$
where $f$ is a scaling function, e.~g. $f(x) = \operatorname{sign}(x)\sqrt{|x|}$. The benefit of this procedure is that it requires $m + n$ samples instead of $mn$, but sacrifices the interlayer independence of noise.

\subsection{Prioritized experience replay}\label{PER}

In DQN each batch of transitions is sampled from experience replay using uniform distribution, treating collected data as equally prioritized. In such scheme states for each update come from the same distribution as they come from interaction experience (except that they become decorellated), which agrees with TD algorithm as the basement of DQN.

Intuitively observed transitions vary in their importance. At the beginning of training most guesses tend to be more or less random as they rely on arbitrarily initialized $Q^*_\theta$ and the only source of trusted information are transitions with non-zero received reward, especially near terminal states where $V^*_\theta(s')$ is known to be equal to 0. In the midway of training, most of experience replay is filled with the memory of interaction within well-learned part of MDP while the most crucial information is contained in transitions where agent explored new promising areas and gained novel reward yet to be propagated through Bellman equation. All these significant transitions are drowned in collected data and rarely appear in sampled batches.

The central idea of prioritized experience replay \citep{schaul2015prioritized} is that priority of some transition $T = (s, a, r', s', \done)$ is proportional to temporal difference:
\begin{equation}\label{priority}
    \rho(T) \coloneqq y(T) - Q^*(s, a, \theta) = \sqrt{\Loss(y(T), Q^*(s, a, \theta))}
\end{equation}
Using these priorities as proxy of transition importances, sampling from experience replay proceeds using following probabilities:
$$\mathcal{P}(T) \propto \rho(T)^\alpha$$
where hyperparameter $\alpha \in \R^+$ controls the degree to which the sampling weights are sparsified: the case $\alpha = 0$ corresponds to uniform sampling distribution while $\alpha = +\infty$ is equivalent to greedy sampling of transitions with highest priority.

The problem with (\ref{priority}) claim is that each transition's priority changes after each network update. As it is impractical to recalculate loss for the whole data after each step, some simplifications must be put up with. The straightforward option is to update priority only for sampled transitions in the current batch. New transitions can be added to experience replay with highest priority, i.~e. $\max\limits_{T} \rho(T)$\footnote{which can be computed online with $\mathcal{O}(1)$ complexity}.

Second debatable issue of prioritized replay is that it actually substitutes loss function of DQN updates, which assumed uniform sampling of visited states to ensure they come from state visitation distribution:
$$\mathbb{E}_{T \sim \operatorname{Uniform}} \Loss(T) \to \min_\theta$$
While it is not clear what distribution is better to sample from to ensure exploration restrictions of theorem \ref{TDconvergence}, prioritized experienced replay changes this distribution in uncontrollable way. Despite its fruitfulness at the beginning and midway of training process, this distribution shift may destabilize learning close to the end and make algorithm stuck with locally optimal policy. Since formally this issue is about estimating an expectation over one probability with preference to sample from another one, the standard technique called \emph{importance sampling} can be used as countermeasure:
\begin{align*}
\mathbb{E}_{T \sim \operatorname{Uniform}} \Loss(T) &= \sum_{i=0}^M \frac{1}{M} \Loss(T_i) = \\ 
&= \sum_{i=0}^M \mathcal{P}(T_i) \frac{1}{M\mathcal{P}(T_i)} \Loss(T_i) = \\
&= \mathbb{E}_{T \sim \mathcal{P}(T)} \frac{1}{M\mathcal{P}(T)} \Loss(T) 
\end{align*}
where $M$ is a number of transitions stored in experience replay memory. Importance sampling implies that we can avoid distribution shift that introduces undesired bias by making smaller gradient updates for significant transitions which now appear in the batches with higher frequency. The price for bias elimination is that importance sampling weights lower prioritization effect by slowing down learning of highlighted new information. 

This duality resembles trade-off between bias and variance, but important moment here is that distribution shift does not cause any seeming issues at the beginning of training when agent behaves close to random and do not produce valid state visitation distribution anyway. The idea proposed in \citep{schaul2015prioritized} based on this intuition is to anneal the importance sampling weights so they correct bias properly only towards the end of training procedure.
$$
\Loss^{\operatorname{prioritizedER}} = \mathbb{E}_{T \sim \mathcal{P}(T)} \left( \frac{1}{B\mathcal{P}(T)} \right)^{\beta(t)} \Loss(T) 
$$
where $\beta(t) \in [0, 1]$ and approaches 1\footnote{often it is initialized by a constant close to 0 and is linearly increased until it reaches 1} as more interaction steps are executed. If $\beta(t)$ is set to 0, no bias correction is held, while $\beta(t) = 1$ corresponds to unbiased loss function, i.~e. equivalent to sampling from uniform distribution.

The most significant and obvious drawback of prioritized experience replay approach is that it introduces additional hyperparameters. Although $\alpha$ represents one number, algorithm's behaviour may turn out to be sensitive to its choosing, and $\beta(t)$ must be designed by engineer as some scheduled motion from something near 0 to 1, and its well-turned selection may require inaccessible knowledge about how many steps it will take for algorithm to <<warm up>>.

\subsection{Multi-step DQN}\label{NstepDQN}

One more widespread modification of Q-learning in RL community is substituting one-step approximation present in Bellman optimality equation (\ref{Bellman}) with $N$-step:

\begin{proposition}
\textbf{($N$-step Bellman optimality equation)}
\begin{equation}\label{NstepBellman}
 Q^*(s_0, a_0) = \E_{\Traj_{\pi^*} \mid s_0, a_0} \left[ \sum_{t=1}^N \gamma^{t-1}r(s_t) + \gamma^N \max_{a_N} Q^*(s_N, a_N) \right]   
\end{equation}
\end{proposition}

Indeed, definition of $Q^*(s, a)$ consists of average return and can be viewed as making $T^{\max}$ steps from state $s_0$ after selecting action $a_0$, while vanilla Bellman optimality equation represents $Q^*(s, a)$ as reward from one next step in the environment and estimation of the rest of trajectory reward recursively. $N$-step Bellman equation (\ref{NstepBellman}) generalizes these two opposites.

All the same reasoning as for DQN can be applied to $N$-step Bellman equation to obtain $N$-step DQN algorithm, which only modification appears in target computation:
\begin{equation}\label{Nsteptarget}
y(s_0, a_0) = \sum_{t=1}^N \gamma^{t-1}r(s_t) + \gamma^N \max_{a_N} Q^*(s_N, a_N, \theta)
\end{equation}
To perform this computation, we are required to obtain for given state $s$ and $a$ not only one next step, but $N$ steps. To do so, instead of transitions $N$-step roll-outs are stored, which can be done by precomputing following tuples:
$$T = \left( s, \, a, \, \sum_{n=1}^N \gamma^{n-1}r^{(n)}, \, s^{(N)}, \, \done \right)$$
where $r^{(n)}$ is the reward received in $n$ steps after visitation of considered state $s$, $s^{(N)}$ is state visited in $N$ steps, and $\done$ is a flag whether the episode ended during $N$-step roll-out\footnote{all $N$-step roll-outs must be considered including those terminated at $k$-th step for $k < N$.}. All other aspects of algorithm remain the same in practical implementations, and the case $N = 1$ corresponds to standard DQN.

The goal of using $N > 1$ is to accelerate propagation of reward from terminal states backwards through visited states to $s_0$ as less update steps will be required to take into account freshly observed reward and optimize behaviour at the beginning of episodes. The price is that formula (\ref{Nsteptarget}) includes an important trick: to calculate such target, for second (and following) step action $a'$ must be sampled from $\pi^*$ for Bellman equation (\ref{NstepBellman}) to remain true. In other words, application of $N$-step Q-learning is theoretically improper when behaviour policy differs from $\pi^*$. Note that we do not face this problem in the case $N = 1$ in which we are required to sample only from transition probability $p(s' \mid s, a)$ for given state-action pair $s, a$.

Even considering $\pi^* \approx \argmax\limits_a Q^*(s, a, \theta)$, where $Q^*$ is our current approximation of $\pi^*$, makes $N$-step DQN an \emph{on-policy} algorithm when for every state-action pair $s, a$ it is preferable to sample target using the closest approximation of $\pi^*$ available. This questions usage of experience replay or at the very least encourages to limit its capacity to store only $M^{\max}$ newest transitions with $M^{\max}$ being relatively not very big.

To see the negative effect of $N$-step DQN, consider the following toy example. Suppose agent makes a mistake on the second step after $s$ and ends episode with huge negative reward. Then in the case $N > 2$ each time the roll-out starting with this $s$ is sampled in the batch, the value of $Q^*(s, a, \theta)$ will be updated with this received negative reward even if $Q^*(s', \cdot, \theta)$ already learned not to repeat this mistake again.

Yet empirical results in many domains demonstrate that raising $N$ from 1 to 2-3 may result in substantial acceleration of training and positively affect the final performance. On the contrary, the theoretical groundlessness of this approach explains its negative effects when $N$ is set too big. 

\newpage

\section{Distributional approach for value-based methods}\label{distributionalRL}

\subsection{Theoretical foundations}

The setting of RL task inherently carries internal stochasticity of which agent has no substantial control. Sometimes intelligent behaviour implies taking risks with severe chance of low episode return. All this information resides in the distribution of return $R$ (\ref{return}) as random variable.

While value-based methods aim at learning expectation of this random variable as it is the quantity we actually care about, in distributional approach \citep{bellemare2017distributional} it is proposed to learn the whole distribution of returns. It further extends the information gathered by algorithm about MDP towards model-based case in which the whole MDP is imitated by learning both reward function $r(s)$ and transitions $\Trans$, but still restricts itself only to reward and doesn't intend to learn world model.

In this section we discuss some theoretical extensions of temporal difference ideas in the case when expectations on both sides of Bellman equation (\ref{QQ}) and Bellman optimality equation (\ref{Bellman}) are taken away.

The central object of study in Q-learning was Q-function, which for given state and action returns the expectation of reward. To rewrite Bellman equation not in terms of expectations, but in terms of the whole distributions, we require a corresponding notation.

\begin{definition} 
For given MDP and policy $\pi$ the \emph{value distribution of policy $\pi$} is a random variable defined as 
$$Z^\pi(s, a) \coloneqq \sum_{t=0} \gamma^t r_{t+1} \, \mathrel{\Big|} \, s_0 = s, a_0 = a $$
\end{definition}

Note that $Z^\pi$ just represents a random variable which is taken expectation of in definition of $Q$-function:
$$Q^\pi(s, a) = \mathbb{E}_{\Traj_\pi} Z^\pi(s, a)$$

Using this definition of value distribution, Bellman equation can be rewritten to extend the recursive connection between adjacent states from expectations of returns to the whole distributions of returns:
\begin{proposition}
\textbf{(Distributional Bellman Equation)} \citep{bellemare2017distributional}
\begin{equation}\label{distributionalBellman}
 Z^\pi(s, a) \cdfeq r(s') + \gamma Z^\pi(s', a') \, \mathrel{\big|} \, s' \sim p(s' \mid s, a), a' \sim \pi(a' \mid s') 
\end{equation}
\end{proposition}
Here we used some auxiliary notation: by $\cdfeq$ we mean that cumulative distribution functions of two random variables to the right and left are equal almost everywhere. Such equations are called \emph{recursive distributional equations} and are well-known in theoretical probability theory\footnote{to get familiar with this notion, consider this basic example: $$X_1 \cdfeq \frac{X_2}{\sqrt{2}} + \frac{X_3}{\sqrt{2}}$$
where $X_1, X_2, X_3$ are random variables coming from $\mathcal{N}(0, \sigma^2)$.}. By using $\mid$ we describe a sampling procedure for the random variable to the right side of equation: for given $s, a$ next state $s'$ is sampled from transition probability, then $a'$ is sampled from given policy, then random variable $Z^\pi(s', a')$ is sampled to calculate a resulting sample $r(s') + \gamma Z^\pi(s', a')$.

While the space of Q-functions $Q^\pi(s, a) \in \St \times \A \to \R$ is finite, the space of value distributions is a space of mappings from state-action pair to continuous distributions:
$$Z^\pi(s, a) \in \St \times \A \to \mathcal{P}(\R)$$
and it is important to notice that even in the table-case when state and action spaces are finite, the space of value distributions is essentially infinite. Crucial moment for us will be that convergence properties now depend on chosen metric\footnote{in finite spaces it is true that convergence in one metric guarantees convergence to the same point for any other metric.}.

The choice of metric in $\St \times \A \to \mathcal{P}(\R)$ represents the same issue as in the space of continuous random variables $\mathcal{P}(\R)$: if we choose a metric in the latter, we can construct one in the former:
\begin{proposition}\label{constructmetric}
If $d(X, Y)$ is a metric in the space $\mathcal{P}(\R)$, then
$$\overline{d}(Z_1, Z_2) \coloneqq \sup_{s \in \St, a \in \A} d(Z_1(s, a), Z_2(s, a))$$
is a metric in the space $\St \times \A \to \mathcal{P}(\R)$.
\end{proposition}

The particularly interesting for us example of metric in $\mathcal{P}(\R)$ will be Wasserstein metric, which concerns only random variables with bounded moments, so we will additionally assume that for all state-action pairs $s, a$
$$\mathbb{E} Z^\pi(s, a)^p \le +\infty$$
are finite for $p \ge 1$.

\begin{proposition} 
For $1 \le p \le +\infty$ for two random variables $X, Y$ on continuous domain with $p$-th bounded moments and cumulative distribution functions $F_X$ and $F_Y$ correspondingly a \emph{Wasserstein distance}
$$W_p(X, Y) \coloneqq \left( \int\limits_0^1 \left| F_X^{-1}(\omega) - F_Y^{-1}(\omega) \right| ^p d\omega \right)^{\frac{1}{p}}$$
$$W_\infty(X, Y) \coloneqq \sup_{\omega \in [0,1]} \left| F_X^{-1}(\omega) - F_Y^{-1}(\omega) \right| $$
is a metric in the space of random variables with $p$-th bounded moments.
\end{proposition}

Thus we can conclude from proposition \ref{constructmetric} that maximal form of Wasserstein metric
\begin{equation}\label{maximalWasserstein}
\overline{W}_p(Z_1, Z_2) = \sup_{s \in \St, a \in \A} W_p(Z_1(s, a), Z_2(s, a))
\end{equation}
is a metric in the space of value distributions.

We now concern convergence properties of point iteration method to solve (\ref{distributionalBellman}) in order to obtain $Z^\pi$ for given policy $\pi$, i.~e. solve the task of \emph{policy evaluation}. For that purpose we initialize $Z_0^\pi(s, a)$ arbitrarily\footnote{here we consider value distributions from theoretical point of view, assuming that we are able to explicitly store a table of $|\St| |\A|$ continuous distributions without any approximations.} and perform the following updates for all state-action pairs $s, a$:
\begin{equation}\label{distrpolicyevalupdate}
    Z^\pi_{t+1}(s, a) \cdfcoloneqq r(s') + \gamma Z^\pi_t(s', a')
\end{equation}
Here we assume that we are able to compute the distribution of random variable on the right side knowing $\pi$, all transition probabilities $\Trans$, the distribution of $Z^\pi_t$ and reward function. The question whether the sequence $\{Z^\pi_t\}$ converges to $Z^\pi$ can be given a detailed answer:

\begin{proposition}\label{DistributionalPolicyEvaluationConvergence} \citep{bellemare2017distributional}
Denote by $\mathcal{B}$ the following operator $\left( \St \times \A \to \mathcal{P}(\R) \right) \to \left( \St \times \A \to \mathcal{P}(\R) \right)$, updating $Z^\pi_t$ as in (\ref{distrpolicyevalupdate}):
$$Z^\pi_{t+1} = \mathcal{B}Z^\pi_t$$
for all state-action pairs $s, a$.

Then $\mathcal{B}$ is a contraction mapping in $\overline{W}_p$ (\ref{maximalWasserstein}) for $1 \le p \le +\infty$, i.e. for any two value distributions $Z_1, Z_2$
$$\overline{W}_p(\mathcal{B} Z_1, \mathcal{B} Z_2) \le \gamma \overline{W}_p(Z_1, Z_2)$$
Hence there is a unique fixed point of system of equations (\ref{distributionalBellman}) and the point iteration method converges to it.
\end{proposition}

One more curious theoretical result is that $\mathcal{B}$ is in general not a contraction mapping for such distances as Kullback-Leibler divergence, Total Variation distance and Kolmogorov distance\footnote{one more metric for which the contraction property was shown is Cramer metric:
$$l_2(X, Y) = \left(\int\limits_\R \left(F_X(\omega) - F_Y(\omega) \right)^2 d\omega \right)^{\frac{1}{2}}$$ where $F_X, F_Y$ are c.d.f. of random variables $X, Y$ correspondingly.}. It shows that metric selection indeed influences convergence rate.

Similar to traditional value functions, we can define \emph{optimal value distribution} $Z^*(s, a)$. Substituting\footnote{to perform this step validly, a clarification concerning $\argmax$ operator definition must be given. The choice of action $a$ returned by this operator in the cases when several actions lead to the same maximal average returns must not depend on $Z$, as this choice affects higher moments of resulted distribution. To overcome this issue, for example, in the case of finite action space all actions can be enumerated and the optimal action with the lowest index is returned by operator.} $\pi^*(s) = \argmax\limits_a \mathbb{E}_{\Traj_{\pi^*}} Z^*(s, a)$ into (\ref{distributionalBellman}), we obtain distributional Bellman optimality equation:
\begin{proposition}
\textbf{(Distributional Bellman optimality equation)}
\begin{equation}\label{optdistributionalBellman}
 Z^*(s, a) \cdfeq r(s') + \gamma Z^*(s', \argmax\limits_{a'} \mathbb{E}_{\Traj_{\pi^*}} Z^*(s', a')) \, \mathrel{\big|} \, s' \sim p(s' \mid s, a) 
\end{equation}
\end{proposition}

Now we concern the same question whether the point iteration method of solving (\ref{optdistributionalBellman}) leads to solution $Z^*$ and whether it is a contraction mapping for some metric. The answer turns out to be negative.

\begin{proposition}\label{divergence} \citep{bellemare2017distributional}
Point iteration for solving (\ref{optdistributionalBellman}) may diverge.
\end{proposition}

Level of impact of this result is not completely clear. Point iteration for (\ref{optdistributionalBellman}) preserves means of distributions, i.~e. it will eventually converge to $Q^*(s, a)$ with all theoretical guarantees from classical Q-learning. The reason behind divergence theorems hides in the rest of distributions like other moments and situations when equivalent (in terms of average return) actions may lead to different higher moments. 

\subsection{Categorical DQN}\label{CategoricalDQN}

There are obvious obstacles for practical application of distributional Q-learning following from complication of working with arbitrary continuous distributions. Usually we are restricted to approximations inside some family of parametric distributions, so we have to perform a projection step on each iteration.  

Second matter in combining distributional Q-learning with deep neural networks is to take into account that only samples from $p(s' \mid s, a)$ are available for each update. To provide a distributional analog of temporal difference algorithm \ref{Qlearningupdate2}, some analog of exponential smoothing for distributional setting must be proposed.

Categorical DQN \citep{bellemare2017distributional} (also referred as c51) provides straightforward design of practical distributional algorithm. While DQN was a resemblance of temporal difference algorithm, Categorical DQN attempts to follow the logic of DQN.

The concept is as following. The neural network with parameters $\theta$ in this setting takes as input $s \in \St$ and for each action $a$ outputs parameters $\zeta_\theta(s, a)$ of distributions of random variable $Z^*_\theta(s, a)$. As in DQN, experience replay can be used to collect observed transitions and sample a batch for each update step. For each transition $T = (s, a, r', s', \done)$ in the batch a guess is computed:
\begin{equation}\label{distributionaltarget}
y(T) \cdfcoloneqq r' + (1 - \done)\gamma Z^*_\theta \left( s', \argmax_{a'} \mathbb{E} Z^*_\theta(s', a') \right)     
\end{equation}
Note that expectation of $Z^*_\theta(s', a')$ is computed explicitly using the form of chosen parametric family of distributions and outputted parameters $\zeta_\theta(s', a')$, as is the distribution of random variable $r' + (1 - \done)\gamma Z^*_\theta(s', a')$. In other words, in this setting guess $y(T)$ is also a continuous random variable, distribution of which can be constructed only approximately. As both target and model output are distributions, it is reasonable to design loss function in a form of some divergence $\mathcal{D}$ between $y(T)$ and $Z^*_\theta(s, a)$:
\begin{equation}\label{distributionalloss}
\Loss(\theta) = \mathbb{E}_T \mathcal{D} \left( y(T) \parallel Z^*_\theta(s, a) \right)    
\end{equation}
$$\theta_{t+1} = \theta_t - \alpha \frac{\partial \Loss(\theta_t)}{\partial \theta}$$

The particular choice of this divergence must be made with concern that $y(T)$ is a <<sample>> from a full one-step approximation of $Z^*_\theta$ which includes transition probabilities:
\begin{equation}\label{yfull}
y^{\operatorname{full}}(s, a) \cdfcoloneqq \sum_{s' \in \St} p(s' \mid s, a)y(s, a, r(s'), s', \done(s'))    
\end{equation}
This form is precisely the right side of distributional Bellman optimality equation as we just incorporated intermediate sampling of $s'$ into the value of random variable. In other words, if transition probabilities $\Trans$ were known, the update could be made using distribution of $y^{\operatorname{full}}$ as a target.
$$\Loss^{\operatorname{full}}(\theta) = \mathbb{E}_{s, a} \mathcal{D}(y^{\operatorname{full}}(s, a) \parallel Z^*_\theta(s, a))$$

This motivates to choose $\KL(y(T) \parallel Z^*_\theta(s,a))$ (specifically with this order of arguments) as $\mathcal{D}$ to exploit the following property (we denote by $p_X$ a p.d.f. pf random variable $X$):
\begin{align*}
\nabla_\theta \mathbb{E}_T \KL(y^{\operatorname{full}}(s, a) \parallel Z^*_\theta(s, a)) &= \nabla_\theta \left[ \mathbb{E}_T \int_\R -p_{y^{\operatorname{full}}(s, a)}(\omega) \log p_{Z^*_\theta(s, a))}(\omega) d\omega + \const (\theta) \right] = \\
\{ \text{using (\ref{yfull})} \}&= \nabla_\theta \mathbb{E}_T \int_\R \mathbb{E}_{s' \sim p(s' \mid s, a)}-p_{y(T)}(\omega) \log p_{Z^*_\theta(s, a))}(\omega) d\omega = \\
\{ \text{taking expectation out} \} &= \nabla_\theta \mathbb{E}_T \mathbb{E}_{s' \sim p(s' \mid s, a)} \int_\R -p_{y(T)}(\omega) \log p_{Z^*_\theta(s, a))}(\omega) d\omega = \\
&= \nabla_\theta \mathbb{E}_T \mathbb{E}_{s' \sim p(s' \mid s, a)} \KL \left( y(T) \parallel Z^*_\theta(s, a) \right)
\end{align*}

This property basically states that gradient of loss function (\ref{distributionalloss}) with $\KL$ as $\mathcal{D}$ is an unbiased (Monte-Carlo) estimation of gradient of $\KL$-divergence for <<full>> distribution (\ref{yfull}), which resembles the employment of exponential smoothing in temporal difference learning. For many other divergences, including Wasserstein metric, same statement is not true, so their utilization in described online setting will lead to biased gradients and all theory-grounded intuition that algorithm moves in the right direction becomes distinctively lost. Moreover, $\KL$-divergence is known to be one of the easiest divergences to work with due to its nice smoothness properties and wide prevalence in many deep learning pipelines.

Described above motivation to choose $\KL$-divergence as an actual objective for minimization is contradictory. Theoretical analysis of distributional Q-learning, specifically theorem \ref{DistributionalPolicyEvaluationConvergence}, though concerning policy evaluation other than optimal $Z^*$ search, explicitly hints that the process converges exponentially fast for Wasserstein metric, while even for precisely made updates in terms of $\KL$-divergence we are not guaranteed to get any closer to true solution.

More <<practical>> defect of $\KL$-divergence is that it demands two comparable distributions to share the same domain. This means that by choosing $\KL$-divergence we pledge to guarantee that $y(T)$ and $Z^*_\theta(s, a)$ in (\ref{distributionalloss}) have coinciding support. This emerging restriction seems limiting even beforehand as for episodic MDP value distribution in terminal states is obviously degenerated (their support consists of one point ${r(s)}$ which is given all probability mass) which means that our value distribution approximation is basically ensured to never be precise.

In Categorical DQN, as follows from the name, the family of distributions is chosen to be categorical on the fixed support $\{z_0, z_1 \dots z_{A-1}\}$ where $A$ is number of \emph{atoms}. As no prior information about MDP is given, the basic choice of this support is uniform grid from some $V_{\min} \in \R$ to $V^{\max} \in \R$:
$$z_i = V_{\min} + \frac{i}{A - 1}(V_{\max} - V_{\min}), \quad i \in {0, 1, \dots A - 1}$$
These bounds, though, must be chosen carefully as they implicitly assume
$$V_{\min} \le Z^*(s, a) \le V_{\max}$$
and if these inequalities are not tight, the approximation will obviously become poor.

Therefore the neural network outputs $A$ numbers, summing into 1, to represent arbitrary distribution on this support:
$$\zeta_i(s, a, \theta) \coloneqq \mathcal{P}(Z^*_\theta(s, a) = z_i)$$
Within this family of distributions, computation of expectation, greedy action selection and $KL$-divergence is trivial. One problem hides in target formula (\ref{distributionaltarget}): while we can compute distribution $y(T)$, its support may in general differ from $\{z_0 \dots z_{A-1}\}$. To avoid the issue of disjoint supports, a \emph{projection step} must be done to find the closest to target distribution within the chosen family\footnote{to project a categorical distribution with support $\{v_0, v_1 \dots v_{A-1}\}$ on categorical distributions with support $\{z_0, z_1 \dots z_{A-1}\}$ one can just find for each $v_i$ the closest two atoms $z_j \le v_i \le z_{j+1}$ and split all probability mass for $v_i$ between $z_j$ and $z_{j+1}$ proportional to closeness. If $v_i < z_0$, then all its probability mass is given to $z_0$, same with upper bound.}. Therefore the resulting target used in the loss function is
\begin{equation*}
y(T) \cdfcoloneqq \Pi_C \left[ r' + (1 - \done)\gamma Z^*_\theta \left( s', \argmax_{a'} \mathbb{E} Z^*_\theta(s', a') \right) \right]   
\end{equation*}
where $\Pi_C$ is projection operator.

The resulting practical algorithm, named c51 after categorical distributions with $A = 51$ atoms, inherits ideas of experience replay, $\eps$-greedy exploration and target network from DQN. Empirically, though, usage of target network remains an open question as the chosen family of distributions restricts value approximation from unbounded growth by <<clipping>> predictions at $z_{A - 1}$ and $z_0$, yet it is still considered slightly improving performance.

\begin{algorithmbox}[label = c51algorithm]{Categorical DQN (c51)}
\textbf{Hyperparameters:} $B$ --- batch size, $V_{\max}, V_{\min}, A$ --- parameters of support, $K$ --- target network update frequency, $\eps(t) \in (0, 1]$ --- greedy exploration parameter, $\zeta^*$ --- neural network, SGD optimizer.

\vspace{0.3cm}
Initialize weights $\theta$ of neural net $\zeta^*$ arbitrary \\
Initialize $\theta^- \gets \theta$ \\
Precompute support grid $z_i = V_{\min} + \frac{i}{A - 1}(V_{\max} - V_{\min})$ \\
\textbf{On each interaction step:}
\begin{enumerate}
    \item select $a$ randomly with probability $\varepsilon(t)$, else $a = \argmax\limits_a \sum_{i} z_i\zeta^*_i(s, a, \theta)$
    \item observe transition $(s, a, r', s', \done)$
    \item add observed transition to experience replay
    \item sample batch of size $B$ from experience replay
    \item for each transition $T$ from the batch compute target:
    $$\mathcal{P}(y(T) = r' + \gamma z_i) = \zeta^*_i\left( s', \argmax_{a'} \sum_{i} z_i \zeta^*_i(s', a', \theta^-), \theta^- \right) $$
    \item project $y(T)$ on support $\{ z_0, z_1 \dots z_{A-1} \}$
    \item compute loss:
    $$\Loss = \frac{1}{B}\sum_T \KL(y(T) \parallel Z^*(s, a, \theta))$$
    \item make a step of gradient descent using $\frac{\partial \Loss }{\partial \theta}$
    \item if $t \mod K = 0$: $\theta^- \gets \theta$
\end{enumerate}
\end{algorithmbox}

\subsection{Quantile Regression DQN (QR-DQN)}\label{QRDQN}

Categorical DQN discovered a gap between theory and practice as $\KL$-divergence, used in practical algorithm, is theoretically unjustified. Theorem \ref{DistributionalPolicyEvaluationConvergence} hints that the true divergence we should care about is actually Wasserstein metric, but it remained unclear how it could be optimized using only samples from transition probabilities $\Trans$.

In \citep{dabney2018distributional} it was discovered that selecting another family of distributions to approximate $Z^*_\theta(s, a)$ will reduce Wasserstein minimization task to the search for quantiles of specific distributions. The latter can be done in online setting using \emph{quantile regression} technique. This led to alternative distributional Q-learning algorithm named Quantile Regression DQN (QR-DQN).

The basic idea is to <<swap>> fixed support and learned probabilities of Categorical DQN. We will now consider the family with fixed probabilities for $A$-atomed categorical distribution with arbitrary support $\{ \zeta^*_0(s, a, \theta), \zeta^*_1(s, a, \theta), \dots , \zeta^*_{A - 1}(s, a, \theta)\}$. Again, we will assume all probabilities to be equal given the absence of any prior knowledge; namely, our distribution family is now
$$Z^*_\theta(s, a) \sim \operatorname{Uniform}\left( \zeta^*_0(s, a, \theta), \dots , \zeta^*_{A - 1}(s, a, \theta) \right)$$
In this setting neural network outputs $A$ arbitrary real numbers that represent the support of uniform categorical distribution\footnote{Note that target distribution is now guaranteed to remain within this distribution family as multiplying on $\gamma$ just shrinks the support and adding $r'$ just shifts it. We assume that if some atoms of the support coincide, the distribution is still $A$-atomed categorical; for example, for degenerated distribution (like in the case of terminal states) $\zeta^*_0(s, a, \theta) = \zeta^*_1(s, a, \theta) = \dots = \zeta^*_{A - 1}(s, a, \theta)$. This shows that projection step heuristic is not needed for this particular choice of distribution family.}, where $A$ is the number of atoms and the only hyperparameter to select.

For table-case setting, on each step of point iteration we desire to update the cell for given state-action pair $s, a$ with full distribution of random variable to the right side of (\ref{optdistributionalBellman}). If we are limited to store only $A$ atoms of the support, the true distribution must be projected on the space of $A$-atomed categorical distributions. Consider now this task of projecting some given random variable with c.d.f. $F(\omega)$ in terms of Wasserstein distance. Specifically, we will be interested in minimizing $\mathcal{W}_1$-distance for $p = 1$ as the theorem \ref{DistributionalPolicyEvaluationConvergence} states the contraction property for all $1 \le p \le +\infty$ and we are free to choose any:
\begin{equation}\label{Wassersteinminimization}
\int_0^1 \left| F^{-1}(\omega) - U^{-1}_{z_0, z_1 \dots z_{A-1}}(\omega) \right| d\omega \to \min_{z_0, z_1 \dots z_{A-1}}
\end{equation}
where $U_{z_0, z_1 \dots z_{A-1}}$ is c.d.f. for uniform categorical distribution on given support. Its inverse, also known as \emph{quantile function}, has a following simple form:
$$U^{-1}_{z_0, z_1 \dots z_{A-1}}(\omega) = \begin{cases}
z_0 \quad &0 \le \omega < \frac{1}{A} \\
z_1 \quad &\frac{1}{A} \le \omega < \frac{2}{A} \\
\vdots \\
z_{A - 1} \quad &\frac{A - 1}{A} \le \omega < 1
\end{cases}$$
Substituting this into (\ref{Wassersteinminimization})
\begin{equation*}
\sum_{i = 0}^{A - 1} \int_{\frac{i}{A}}^{\frac{i + 1}{A}} \left| F^{-1}(\omega) - z_i \right| d\omega \to \min_{z_0, z_1 \dots z_{A-1}}
\end{equation*}
splits the optimization of Wasserstein into $A$ independent tasks that can be solved separately:
\begin{equation}\label{w1minim}
\int_{\frac{i}{A}}^{\frac{i + 1}{A}} \left| F^{-1}(\omega) - z_i \right| d\omega \to \min_{z_i}
\end{equation}

\begin{proposition}\label{quantilesisdesire} \citep{dabney2018distributional}
Let's denote
$$\tau_i \coloneqq \frac{\frac{i}{A} + \frac{i + 1}{A}}{2}$$
Then every solution for (\ref{w1minim}) satisfies $F(z_i) = \tau_i$, i.~e. it is $\tau_i$-th quantile of c.~d.~f. $F$.
\end{proposition}

The result \ref{quantilesisdesire} states that we require only $A$ specific quantiles of random variable to the right side of Bellman equation\footnote{It can be proved that for table-case policy evaluation algorithm which stores in each cell not expectations of reward (as in Q-learning) but $A$ quantiles updated according to distributional Bellman equation (\ref{distributionalBellman}) using theorem \ref{quantilesisdesire} converges to quantiles of $Z^*(s, a)$ in Wasserstein metric for $1 \le p \le +\infty$ and its update operator is a contraction mapping in $\mathcal{W}_{\infty}$.}. Hence the last thing to do to design a practical algorithm is to develop a procedure of unbiased estimation of quantiles for the random variable on the right side of distribution Bellman optimality equation (\ref{optdistributionalBellman}).

Quantile regression is the standard technique to estimate the quantiles of \emph{empirical distribution} (i.~.e. distribution that is represented by finite amount of i.~i.~d. samples from it). Recall from machine learning that the constant solution optimizing l1-loss is median, i.~.e. $\frac{1}{2}$-th quantile. This fact can be generalized to arbitrary quantiles:
\begin{proposition}
\textbf{(Quantile Regression)} \citep{koenker1978regression}
Let's define loss as
$$\Loss(c, X) = \begin{cases}
\tau (c - X) \quad &c \ge X \\
(1 - \tau) (X - c) \quad &c < X \\
\end{cases}$$
Then solution for
\begin{equation}\label{QR}
\mathbb{E}_X \Loss(c, X) \to \min_{c \in \R}
\end{equation}
is $\tau$-th quantile of distribution of $X$.
\end{proposition}

As usual in the case of neural networks, it is impractical to optimize (\ref{QR}) until convergence on each iteration for each of $A$ desired quantiles $\tau_i$. Instead just one step of gradient optimization is made and the outputs of neural network $\zeta^*_i(s, a, \theta)$, which play the role of $c$ in formula (\ref{QR}), are moved towards the quantile estimation via backpropagation. In other words, (\ref{QR}) sets a loss function for network outputs; the losses for different quantiles are summed up. The resulting loss is
\begin{equation}\label{QRloss}
    \Loss^{\operatorname{QR}}(s, a, \theta) = \sum_{i=0}^{A - 1} \mathbb{E}_{s' \sim p(s' \mid s, a)} \mathbb{E}_{y \sim y(T)} \left( \tau - \mathbb{I}[\zeta^*_i(s, a, \theta) < y] \right) \left( \zeta^*_i(s, a, \theta) - y \right)
\end{equation}
where $\mathbb{I}$ denotes an indicator function. The expectation over $y \sim y(T)$ for given transition can be computed in closed form: indeed, $y(T)$ is also an $A$-atomed categorical distribution with support $\{r' + \gamma \zeta_0^*(s', a'), \dots, r' + \gamma \zeta_{A - 1}^*(s', a')\}$, where 
$$a' = \argmax\limits_{a'} \mathbb{E} Z^*(s', a', \theta) = \argmax\limits_{a'} \frac{1}{A}\sum_i \zeta_i^*(s', a', \theta)$$
and expectation over transition probabilities, as always, is estimated using Monte-Carlo by sampling transitions from experience replay.

\begin{algorithmbox}[label = QRDQNalgorithm]{Quantile Regression DQN (QR-DQN)}
\textbf{Hyperparameters:} $B$ --- batch size, $A$ --- number of atoms, $K$ --- target network update frequency, $\eps(t) \in (0, 1]$ --- greedy exploration parameter, $\zeta^*$ --- neural network, SGD optimizer.

\vspace{0.3cm}
Initialize weights $\theta$ of neural net $\zeta^*$ arbitrary \\
Initialize $\theta^- \gets \theta$ \\
Precompute mid-quantiles $\tau_i = \frac{\frac{i}{A} + \frac{i+1}{A}}{2}$ \\
\textbf{On each interaction step:}
\begin{enumerate}
    \item select $a$ randomly with probability $\varepsilon(t)$, else $a = \argmax\limits_a \frac{1}{A}\sum_{i} \zeta^*_i(s, a, \theta)$
    \item observe transition $(s, a, r', s', \done)$
    \item add observed transition to experience replay
    \item sample batch of size $B$ from experience replay
    \item for each transition $T$ from the batch compute the support of target distribution:
    $$y(T)_j = r' + \gamma \zeta^*_j\left( s', \argmax_{a'} \frac{1}{A}\sum_{i} \zeta^*_i(s', a', \theta^-), \theta^- \right) $$
    \item compute loss:
    $$\Loss = \frac{1}{BA}\sum_T \sum_i \sum_j \left( \tau_i - \mathbb{I}[\zeta^*_i(s, a, \theta) < y(T)_j] \right) \left( \zeta^*_i(s, a, \theta) - y(T)_j \right)$$
    \item make a step of gradient descent using $\frac{\partial \Loss }{\partial \theta}$
    \item if $t \mod K = 0$: $\theta^- \gets \theta$
\end{enumerate}
\end{algorithmbox}

\subsection{Rainbow DQN}\label{rainbow}

Success of Deep Q-learning encouraged a full-scale research of value-based deep reinforcement learning by studying various drawbacks of DQN and developing auxiliary extensions. In many articles some extensions from previous research were already considered and embedded in compared algorithms during empirical studies.

In Rainbow DQN \citep{hessel2018rainbow}, seven Q-learning-based ideas are united in one procedure with ablation studies held whether all these incorporated extensions are essentially necessary for resulted RL algorithm:
\begin{itemize}
    \item DQN (sec. \ref{DQN})
    \item Double DQN (sec. \ref{DDQN})
    \item Dueling DQN (sec. \ref{DueDQN})
    \item Noisy DQN (sec. \ref{NoisyDQN})
    \item Prioritized Experience Replay (sec. \ref{PER})
    \item Multi-step DQN (sec. \ref{NstepDQN})
    \item Categorical\footnote{Quantile Regression can be considered instead} DQN (sec. \ref{CategoricalDQN})
\end{itemize}

There is little ambiguity on how these ideas can be combined; we will discuss several non-straightforward circumstances and provide the full algorithm description after.

To apply prioritized experience replay in distributional setting, the measure of transition importance must be provided. The main idea is inherited from ordinary DQN where priority is just loss for this transition:
$$\rho(T) \coloneqq \Loss(y(T), Z^*(s, a, \theta)) = \KL(y(T) \parallel Z^*(s, a, \theta))$$

To combine noisy networks with double DQN heuristic, it is proposed to resample noise on each forward pass through the network and through its copy for target computation. This decision implies that action selection, action evaluation and network utilization are independent and stochastic (for exploration cultivation) steps.

The one snagging combination here is categorical DQN and dueling DQN. To merge these ideas, we need to model advantage $A^*(s, a, \theta)$ in distributional setting. In Rainbow this is done straightforwardly: the network has two heads, value stream $v(s, \theta)$ outputting $A$ real values and advantage stream $a(s, a, \theta)$ outputting $A \times |\A|$ real values. Then these streams are integrated using the same formula (\ref{dueling}) with the only exception being softmax applied across atoms dimension to guarantee that output is categorical distribution:
\begin{equation}\label{rainbowdueling}
\zeta^*_i(s, a, \theta) \propto \operatorname{exp} \left( v(s, \theta)_i + a(s, a, \theta)_i - \frac{1}{|\A|}\sum_a a(s, a, \theta)_i \right)
\end{equation}
Combining lack of intuition behind this integration formula with usage of mean instead of theoretically justified max makes this element of Rainbow the most questionable. During the ablation studies it was discovered that dueling architecture is the only component that can be removed without noticeable loss of performance. All other ingredients are believed to be crucial for resulting algorithm as they address different problems.

\begin{algorithmbox}[label = rainbowalg]{Rainbow DQN}
\textbf{Hyperparameters:} $B$ --- batch size, $V_{\max}, V_{\min}, A$ --- parameters of support, $K$ --- target network update frequency, $N$ --- multi-step size, $\alpha$ --- degree of prioritized experience replay, $\beta(t)$ --- importance sampling bias correction for prioritized experience replay, $\zeta^*$ --- neural network, SGD optimizer.

\vspace{0.3cm}
Initialize weights $\theta$ of neural net $\zeta^*$ arbitrary \\
Initialize $\theta^- \gets \theta$ \\
Precompute support grid $z_i = V_{\min} + \frac{i}{A - 1}(V_{\max} - V_{\min})$ \\
\textbf{On each interaction step:}
\begin{enumerate}
    \item select $a = \argmax\limits_a \sum_{i} z_i\zeta^*_i(s, a, \theta, \eps), \eps \sim \mathcal{N}(0, I)$
    \item observe transition $(s, a, r', s', \done)$
    \item construct $N$-step transition $T = \left( s, a, \sum_{n=0}^N \gamma^n r^{(n+1)}, s^{(N)}, \done \right)$ and add it to experience replay with priority $\max_T \rho(T)$
    \item sample batch of size $B$ from experience replay using probabilities $\mathcal{P}(T) \propto \rho(T)^\alpha$
    \item compute weights for the batch (where $M$ is the size of experience replay memory)
    $$w(T) = \left( \frac{1}{M\mathcal{P}(T)} \right)^{\beta(t)}$$
    \item for each transition $T = (s, a, \bar{r}, \bar{s}, \done )$ from the batch compute target (detached from computational graph to prevent backpropagation):
    $$\eps_1, \eps_2 \sim \mathcal{N}(0, I)$$
    $$\mathcal{P}(y(T) = \bar{r} + \gamma^N z_i) = \zeta^*_i\left( \bar{s}, \argmax_{\bar{a}} \sum_{i} z_i \zeta^*_i(\bar{s}, \bar{a}, \theta, \eps_1), \theta^-, \eps_2 \right)$$
    \item project $y(T)$ on support $\{ z_0, z_1 \dots z_{A-1} \}$
    \item update transition priorities
    $$\rho(T) \gets \KL(y(T) \parallel Z^*(s, a, \theta, \eps)), \eps \sim \mathcal{N}(0, I)$$
    \item compute loss:
    $$\Loss = \frac{1}{B}\sum_T w(T) \rho(T) $$
    \item make a step of gradient descent using $\frac{\partial \Loss }{\partial \theta}$
    \item if $t \mod K = 0$: $\theta^- \gets \theta$
\end{enumerate}
\end{algorithmbox}
 
\newpage

\section{Policy Gradient algorithms}\label{policygradient}

\subsection{Policy Gradient theorem}\label{pgt}

Alternative approach to solving RL task is direct optimization of objective
\begin{equation}\label{policygradientgoal}
J(\theta) = \E_{\Traj \sim \pi_\theta} \sum_{t=1} \gamma^{t-1}r_t \to \max_{\theta}
\end{equation}as a function of $\theta$. Policy gradient methods provide a framework how to construct an efficient optimization procedure based on stochastic first-order optimization within RL setting.

We will assume that $\pi_\theta(a \mid s)$ is a stochastic policy parameterized with $\theta \in \Theta$. It turns out, that if $\pi$ is differentiable by $\theta$, then so is our goal (\ref{policygradientgoal}). We now proceed to discuss the technique of derivative calculation which is based on employment of \emph{log-derivative trick}:
\begin{proposition}
For arbitrary distribution $\pi(a)$ parameterized by $\theta$:
\begin{equation}\label{logderivative}
\nabla_\theta \pi(a) = \pi(a) \nabla_\theta \log \pi(a)
\end{equation}
\end{proposition}

In most general form, this trick allows us to derive the gradient of expectation of an arbitrary function $f(a, \theta): \A \times \Theta \to \R$, differentiable by $\theta$, with respect to some distribution $\pi_\theta(a)$, also parameterized by $\theta$:
\begin{align*}
    \nabla_\theta \E_{a \sim \pi_\theta(a)} f(a, \theta) &= \nabla_\theta \int_{\A} \pi_\theta(a) f(a, \theta) da = \\
    &= \int_{\A} \nabla_\theta \left[ \pi_\theta(a) f(a, \theta) \right] da = \\
    \{ \text{product rule} \}&= \int_{\A} \left[ \nabla_\theta \pi_\theta(a) f(a, \theta) + \pi_\theta(a) \nabla_\theta f(a, \theta) \right] da = \\
    &= \int_{\A} \nabla_\theta \pi_\theta(a) f(a, \theta) da + \E_{\pi_\theta(a)} \nabla_\theta f(a, \theta) = \\
    \{ \text{log-derivative trick (\ref{logderivative})} \}&= \int_{\A} \pi_\theta(a) \nabla_\theta \log \pi_\theta(a) f(a, \theta) da + \E_{\pi_\theta(a)} \nabla_\theta f(a, \theta) = \\
    &= \E_{\pi_\theta(a)} \nabla_\theta \log \pi_\theta(a) f(a, \theta) + \E_{\pi_\theta(a)} \nabla_\theta f(a, \theta)
\end{align*}

This technique can be applied sequentially (to expectations over $\pi_\theta(a_0 \mid s_0)$, $\pi_\theta(a_1 \mid s_1)$ and so on) to obtain the gradient $\nabla_\theta J(\theta)$.
\begin{proposition}
\textbf{(Policy Gradient Theorem)} \citep{sutton2000policy}
For any MDP and differentiable policy $\pi_\theta$ the gradient of objective (\ref{policygradientgoal}) is
\begin{equation}\label{gradient}
\nabla_{\theta} J(\theta) = \E_{\Traj \sim \pi_\theta} \sum_{t=0} \gamma^t \nabla_{\theta} \log \pi_{\theta}(a_t \mid s_t) Q^\pi(s_t, a_t)
\end{equation}
\end{proposition}

For future references, we require another form of formula (\ref{gradient}), which provides another point of view. For this purpose, let us define a discounted state visitation frequency:
\begin{definition}\label{dsvf}
For given MDP and given policy $\pi$ its \emph{discounted state visitation frequency} is defined by
$$d_\pi(s) \coloneqq (1 - \gamma)\sum_{t = 0} \gamma^t \mathcal{P}(s_t = s)$$
where $s_t$ are taken from trajectories $\Traj$ sampled using given policy $\pi$.
\end{definition}

Discounted state visitation frequencies, if normalized, represent a marginalized probability for agent to land in a given state $s$\footnote{the $\gamma^t$ weighting in this definition is often introduced to incorporate the same reduction of contribution of later states in the whole gradient according to (\ref{gradient}). Similar notation is sometimes used for state visitation frequency without discount.}. It is rarely attempted to be learned, but it assists theoretical study by allowing us to rewrite expectations over trajectories with separated intrinsic and extrinsic randomness of the decision making process:
\begin{equation}\label{gradientalt}
\nabla_{\theta} J(\theta) = \E_{s \sim d_\pi(s)} \E_{a \sim \pi(a \mid s)}\nabla_{\theta} \log \pi_{\theta}(a \mid s) Q^\pi(s, a)   
\end{equation}
This form is equivalent to (\ref{gradient}) as sampling a trajectory and going through all visited states with weights $\gamma^t$ induces the same distribution as defined in $d_\pi(s)$.

Now, although we acquired an explicit form of objective's gradient, we are able to compute it only approximately, using Monte-Carlo estimation for expectations via sampling one or several trajectories. Second form of gradient (\ref{gradientalt}) reveals that it is possible to use roll-outs of trajectories without waiting for episode ending, as the states for the roll-outs come from the same distribution as they would for complete episode trajectories\footnote{in practice and in most policy gradients algorithms, sampling roll-outs never include $\gamma^t$ weights, which formally corresponds to estimating gradient using incorrect equation (<<approximation>>):
$$\nabla_{\theta} J(\theta) \approx \E_{\Traj \sim \pi_\theta} \sum_{t=0} \nabla_{\theta} \log \pi_{\theta}(a_t \mid s_t) Q^\pi(s_t, a_t)$$
which differs from the correct version (\ref{gradient}) in ignoring $\gamma^t$ multiplier. On the one hand, it equalizes the contribution of different terms and agrees with intuition, but on the other hand such gradient estimation does not imply optimization of any reasonable objective and breaks the idea of straightforward gradient ascent \citep{nota2019policy}.
}. The essential thing is that exactly the policy $\pi(\theta)$ must be used for sampling to obtain unbiased Monte-Carlo estimation (otherwise state visitation frequency $d_\pi(s)$ is different). These features are commonly underlined by notation $\E_\pi$, which is a shorter form of $\E_{s \sim d_\pi(s)} \E_{a \sim \pi(a \mid s)}$. When convenient, we will use it to reduce the gradient to a shorter form:
\begin{equation}\label{gradientshort}
\nabla_{\theta} J(\theta) = \E_{\pi(\theta)}\nabla_{\theta} \log \pi_{\theta}(a \mid s) Q^\pi(s, a)
\end{equation}

Second important thing worth mentioning is that $Q^\pi(s, a)$ is essentially present in the gradient. Remark that it is never available to the algorithm and must also be somehow estimated.

\subsection{REINFORCE}\label{reinforce_section}

REINFORCE \citep{williams1992simple} provides a straightforward approach to approximately calculate the gradient (\ref{gradient}) in episodic case using Monte-Carlo estimation: $N$ games are played and Q-function under policy $\pi$ is approximated with corresponding return:
$$Q^\pi(s, a) = \E_{\Traj \sim \pi_\theta \mid s, a} R(\Traj) \approx R(\Traj), \quad \Traj \sim \pi_\theta \mid s, a$$

The resulting formula is therefore the following:
\begin{equation}\label{reinforce}
\nabla_{\theta} J(\theta) \approx \frac{1}{N}\sum_{\Traj}^N \sum_{t=0} \left[ \gamma^t \nabla_{\theta} \log \pi_{\theta}(a_t \mid s_t) \left( \sum_{t' = t} \gamma^{t' - t}r_{t' + 1}\right) \right]
\end{equation}
This estimation is unbiased as both approximation of $Q^\pi$ and approximation of expectation over trajectories are done using Monte-Carlo. Given that estimation of gradient is unbiased, stochastic gradient ascent or more advanced stochastic optimization techniques are known to converge to local optimum. 

From theoretical point of view REINFORCE can be applied straightforwardly for any parametric family $\pi_{\theta}(a \mid s)$ including neural networks. Yet the enormous time required for convergence and the problem of stucking in local optimums make this naive approach completely impractical.

The main source of problems is believed to be the \emph{high variance} of gradient estimation (\ref{reinforce}), as the convergence rate of stochastic gradient descent directly depends on the variance of gradient estimation. 

The standard technique of variance reduction is an introduction of \emph{baseline}. The idea is to add some term that will not affect the expectation, but may affect the variance. One such baseline can be derived using following reasoning: for any distribution it is true that $\int\limits_\A \pi_\theta(a \mid s) da = 1$. Taking the gradient $\nabla_\theta$ from both sides, we obtain: 
\begin{align*}
0 &= \int_\A \nabla_\theta \pi_\theta(a \mid s) da = \\
\{\text{log-derivative trick (\ref{logderivative})}\} &= \int_\A \pi_\theta(a \mid s) \nabla_\theta \log \pi_\theta(a \mid s) da = \\
& = \E_{\pi_\theta(a \mid s)} \nabla_\theta \log \pi_\theta(a \mid s)
\end{align*}
Multiplying this expression on some constant, we can scale this baseline:
$$\E_{\pi_\theta(a \mid s)} \const(a) \nabla_\theta \log \pi_\theta(a \mid s) = 0$$
Notice that the constant here must be independent of $a$, but may depend on $s$. Application of this technique to our case provides the following result\footnote{this result can be generalized by introducing different baselines for estimation of different components of $\nabla_\theta J(\theta)$.}:
\begin{proposition} 
For any arbitrary function $b(s) \colon \St \to \R$, called \emph{baseline}:
$$\nabla_{\theta} J(\theta) = \E_{\Traj \sim \pi_\theta} \sum_{t=0} \gamma^t \nabla_{\theta} \log \pi_{\theta}(a_t \mid s_t) \left( Q^\pi(s_t, a_t) - b(s_t) \right)$$
\end{proposition}

Selection of the baseline is up to us as long as it does not depend on actions $a_t$. The intent is to choose it in a way that minimizes the variance.

It is believed that high variance of (\ref{reinforce}) originates from multiplication of $Q^\pi(s, a)$, which may have arbitrary scale (e.~.g. in a range $[100, 200]$) while $\nabla_\theta \log \pi_\theta(a_t \mid s_t)$ naturally has varying signs\footnote{this follows, for example, from baseline derivation.}. To reduce the variance, the baseline must be chosen so that absolute values of expression inside the expectation are shifted towards zero. Wherein the optimal baseline is provided by the following theorem:
\begin{proposition}
The solution for
$$\mathbb{V}_{\Traj \sim \pi_\theta} \sum_{t=0} \gamma^t \nabla_{\theta} \log \pi_{\theta}(a_t \mid s_t) \left( Q^\pi(s_t, a_t) - b(s_t) \right) \to \min_{b(s)}$$
is given by
\begin{equation}\label{optimalbaseline}
    b(s) = \frac{\E_{a \sim \pi_\theta(a \mid s)} \gamma^t \|\nabla_{\theta} \log \pi_{\theta}(a \mid s)\|_2^2 Q^\pi(s, a)}{\E_{a \sim \pi_\theta(a \mid s)} \|\nabla_{\theta} \log \pi_{\theta}(a \mid s)\|_2^2}
\end{equation}
\end{proposition}
As can be seen, optimal baseline calculation involves expectations which again can only be computed (in most cases) using Monte-Carlo (both for numerator and denominator). For that purpose, for every visited state $s$ estimations of $Q^\pi(s, a)$ are needed for all (or some) actions $a$, as otherwise estimation of baseline will coincide with estimation of $Q^\pi(s, a)$ and collapse gradient to zero. Practical utilization of result (\ref{optimalbaseline}) is to consider a constant baseline independent of $s$ with similar optimal form:
$$b = \frac{\E_{\Traj \sim \pi_\theta} \sum_{t = 0} \gamma^t \|\nabla_{\theta} \log \pi_{\theta}(a_t \mid s_t)\|_2^2 Q^\pi(s_t, a_t)}{\E_{\Traj \sim \pi_\theta} \sum_{t = 0} \|\nabla_{\theta} \log \pi_{\theta}(a_t \mid s_t)\|_2^2}$$

Utilization of some kind of baseline, not necessarily optimal, is known to significantly reduce the variance of gradient estimation and is an essential part of any policy gradient method. The final step to make this family of algorithms applicable when using deep neural networks is to reduce variance of $Q^\pi$ estimation by employing RL task structure like it was done in value-based methods.

\subsection{Advantage Actor-Critic (A2C)}\label{A2C}

Suppose that in optimal baseline formula (\ref{optimalbaseline}) it happens that $\|\nabla_{\theta} \log \pi_{\theta}(a \mid s)\|_2^2 = \const(a)$. Though in reality this is actually not true, under this circumstance the optimal baseline formula significantly reduces and unravels a close-to-optimal but simple form of baseline:
$$b(s) = \gamma^t \E_{a \sim \pi_\theta(a \mid s)} Q^\pi(s, a) = \gamma^t V^{\pi}(s)$$

Substituting this baseline into gradient formula (\ref{gradientshort}) and recalling the definition of advantage function (\ref{advantage}), the gradient can now be rewritten as follows:
\begin{equation}\label{ACgradient}
\nabla_{\theta} J(\theta) = \E_{\pi(\theta)} \nabla_{\theta} \log \pi_{\theta}(a \mid s) A^\pi(s, a)
\end{equation}

This representation of gradient is used as the basement for most policy gradient algorithms as it offers lower variance while selecting the baseline expressed in terms of value functions which can be efficiently learned similar to how it was done in value-based methods. Such algorithms are usually named Actor-Critic as they consist of two neural networks: $\pi_\theta(a \mid s)$, representing a policy, called an \emph{actor}, and $V^\pi_\phi(s)$ with parameters $\phi$, approximately estimating actor's performance, called a \emph{critic}. Note that the choice of value function to learn can be arbitrary; it is possible to learn $Q^\pi$ or $A^\pi$ instead, as all of them are deeply interconnected. Value function $V^\pi$ is chosen as the simplest one since it depends only on state and thus is hoped to be easier to learn. 

Having a critic $V^\pi_\phi(s)$, Q-function can be approximated in a following way:
$$Q^\pi(s, a) \approx r' + \gamma V^\pi(s') \approx r' + \gamma V^\pi_\phi(s')$$
First approximation is done using Monte-Carlo, while second approximation inevitably introduces bias. Important thing to notice is that at this moment our gradient estimation stops being unbiased and all theoretical guarantees of converging are once again lost.

Advantage function therefore can be obtained according to the definition:
\begin{equation}\label{onestepadvantage}
A^\pi(s, a) = Q^\pi(s, a) - V^\pi(s) \approx r' + \gamma V^\pi_\phi(s') - V^\pi_\phi(s)
\end{equation}
Note that biased estimation of baseline doesn't make gradient estimation biased by itself, as baseline can be an arbitrary function of state. All bias introduction happens inside the approximation of $Q^\pi$. It is possible to use critic only for baseline, which allows complete avoidance of bias, but then the only way to estimate $Q^\pi$ is via playing several games and using corresponding returns, which suffers from higher variance and low sample efficiency.

The logic behind training procedure for the critic is taken from value-based methods: for given policy $\pi$ its value function can be obtained using point iteration for solving
$$V^\pi(s) = \mathbb{E}_{a \sim \pi(a \mid s)}\mathbb{E}_{s' \sim p(s' \mid s, a)} \left[r' + \gamma V^\pi(s') \right]$$
Similar to DQN, on each update a target is computed using current approximation
$$y = r' + \gamma V^\pi_\phi(s')$$
and then MSE is minimized to move values of $V^\pi_\phi(s)$ towards the guess.

Notice that to compute the target for critic we require samples from the policy $\pi$ which is being evaluated. Although actor evolves throughout optimization process, we assume that one update of policy $\pi$ does not lead to significant change of true $V^\pi$ and thus our critic, which approximates value function for older version of policy, is close enough to construct the target. But if samples from, for example, old policy are used to compute the guess, the step of critic update will correspond to learning the value function for old policy other than current. Essentially, this means that both actor and critic training procedures require samples from current policy $\pi$, making Actor-Critic algorithm \emph{on-policy} by design. Consequently, samples that were collected on previous update iterations become useless and can be forgotten. This is the key reason why policy gradient algorithms are usually less sample-efficient than value-based.

Now as we have an approximation of value function, advantage estimation can be done using one-step transitions (\ref{onestepadvantage}). As the procedure of training an actor, i.~.e. gradient estimation (\ref{ACgradient}), also does not demand sampling the whole trajectory, each update now requires only a small roll-out to be sampled. The amount of transitions in the roll-out corresponds to the size of mini-batch.

The problem with roll-outs is that the data is obviously not i.~i.~d., which is crucial for training networks. In value-based methods, this problem was solved with experience replay, but in policy gradient algorithms it is essential to collect samples from scratch after each update of the networks parameters. The practical solution for simulated environments is to launch several instances of environment (for example, on different cores of multiprocessor) in parallel threads and have several parallel interactions. After several steps in each environment, the batch for update is collected by uniting transitions from all instances and one synchronous\footnote{there is also an asynchronous modification of advantage actor critic algorithm (A3C) which accelerates the training process by storing a copy of network for each thread and performing weights synchronization from time to time.} update of networks parameters $\theta$ and $\phi$ is performed.

One more optimization that can be done is to partially share weights of networks $\theta$ and $\phi$. It is justified as first layers of both networks correspond to basic features extraction and these features are likely to be the same for optimal policy and value function. While it reduces the number of training parameters almost twice, it might destabilize learning process as the scales of gradient (\ref{ACgradient}) and gradient of critic's MSE loss may be significantly different, so they should be balanced with additional hyperparameter.  

\begin{algorithmbox}[label = A2Calgorithm]{Advantage Actor-Critic (A2C)}
\textbf{Hyperparameters:} $B$ --- batch size, $V^*_\phi$ --- critic neural network, $\pi_\theta$ --- actor neural network, $\alpha$ --- critic loss scaling, SGD optimizer.

\vspace{0.3cm}
Initialize weights $\theta, \phi$ arbitrary \\
\textbf{On each step:}
\begin{enumerate}
    \item obtain a roll-out of size $B$ using policy $\pi(\theta)$
    \item for each transition $T$ from the roll-out compute advantage estimation:
    $$A^\pi(T) = r' + \gamma V^\pi_\phi(s') - V^\pi_\phi$$
    \item compute target (detached from computational graph to prevent backpropagation):
    $$y(T) = r' + \gamma V^\pi_\phi(s')$$
    \item compute critic loss:
    $$\Loss = \frac{1}{B}\sum_T \left( y(T) - V^\pi_\phi \right) ^2$$
    \item compute critic gradients:
    $$\nabla^{\operatorname{critic}} = \frac{\partial \Loss}{\partial \phi}$$
    \item compute actor gradient:
    $$\nabla^{\operatorname{actor}} = \frac{1}{B}\sum_T \nabla_\theta \log \pi_\theta(a \mid s)A^\pi(T)$$
    \item make a step of gradient descent using $\nabla^{\operatorname{actor}} + \alpha \nabla^{\operatorname{critic}}$
\end{enumerate}
\end{algorithmbox}

\subsection{Generalized Advantage Estimation (GAE)}\label{GAE}

There is a design dilemma in Advantage Actor Critic algorithm concerning the choice whether to use the critic to estimate $Q^\pi(s, a)$ and introduce bias into gradient estimation or to restrict critic employment only for baseline and cause higher variance with necessity of playing the whole episodes for each update step.

Actually, the range of possibilities is wider. Since Actor-Critic is an on-policy algorithm by design, we are free to use $N$-step approximations instead of one-step: using
$$Q^\pi(s, a) \approx \sum_{n = 0}^{N - 1} \gamma^n r^{(n + 1)} + \gamma^N V^\pi \left( s^{(N)} \right)$$
we can define $N$-step advantage estimator as
$$A^\pi_{(N)}(s, a) \coloneqq \sum_{n = 0}^{N - 1} \gamma^n r^{(n + 1)} + \gamma^N V^\pi_\phi \left( s^{(N)} \right) - V^\pi_\phi(s)$$
For $N = 1$ this estimation corresponds to Actor-Critic one-step estimation with high bias and low variance. For $N = \infty$ it yields the estimator with critic used only for baseline with no bias and high variance. Intermediate values correspond to something in between. Note that to use $N$-step advantage estimation we have to perform $N$ steps of interaction after given state-action pair.

Usually finding a good value for $N$ as hyperparameter is difficult as its <<optimal>> value may float throughout the learning process. In Generalized Advantage Estimation (GAE) \citep{schulman2015high} it is proposed to construct an ensemble out of different $N$-step advantage estimators using exponential smoothing with some hyperparameter $\lambda$:
\begin{equation}\label{GAEadest}
A^\pi_{\operatorname{GAE}}(s, a) \coloneqq (1 - \lambda) \left( A^\pi_{(1)}(s, a) + \lambda A^\pi_{(2)}(s, a) + \lambda^2 A^\pi_{(3)}(s, a) + \dots \right)
\end{equation}

Here the parameter $\lambda \in [0, 1]$ allows smooth control over bias-variance trade-off: $\lambda = 0$ corresponds to Actor-Critic with higher bias and lower variance while $\lambda \to 1$ corresponds to REINFORCE with no bias and high variance. But unlike $N$ as hyperparameter, it uses mix of different estimators in intermediate case.

GAE proved to be a convenient way how more information can be obtained from collected roll-out in practice. Instead of waiting for episode termination to compute (\ref{GAEadest}) we may use <<truncated>> GAE which ensembles only those $N$-step advantage estimators that are available:
\begin{equation*}
A^\pi_{\operatorname{trunc. GAE}}(s, a) \coloneqq \frac{ A^\pi_{(1)}(s, a) + \lambda A^\pi_{(2)}(s, a) + \lambda^2 A^\pi_{(3)}(s, a) + \dots + \lambda^{N - 1} A^\pi_{(N)}(s, a)}{1 + \lambda + \lambda^2 + \dots + \lambda^{N - 1}}
\end{equation*}
Note that the amount $N$ of available estimators may be different for different transitions from roll-out: if we performed $K$ steps of interaction in some instance of environment starting from some state-action pair $s, a$, we can use $N = K$ step estimators; for next state-action pair $s', a'$ we have only $N = K - 1$ transitions and so on, while the last state-action pair $s^{N - 1}, a^{N - 1}$ can be estimated only using $A^\pi_{(1)}$ as only $N = 1$ following transition is available. Although different transitions are estimated with different precision (leading to different bias and variance), this approach allows to use all available information for each transition and utilize multi-step approximations without dropping last transitions of roll-outs used only for target computation.  

\subsection{Natural Policy Gradient (NPG)}\label{NPG}

In this section we discuss the motivation and basic principles behind the idea of natural gradient descent, which we will require for future references.

The standard gradient descent optimization method is known to be extremely sensitive to the choice of parametrization. Suppose we attempt to solve the following optimization task:
$$f(q) \to \min_q$$
where $q$ is a distribution and $F$ is arbitrary differentiable function. We often restrict $q$ to some parametric family and optimize similar objective, but with respect to some vector of parameters $\theta$ as unknown variable:
$$f(q_\theta) \to \min_{\theta}$$
Classic example of such problem is maximum likelihood task when we try to fit the parameters of our model to some observed data. The problem is that when using standard gradient descent both the convergence rate and overall performance of optimization method substantially depend on the choice of parametrization $q_\theta$. The problem holds even if we fix specific distribution family as many distribution families allow different parametrizations.

To see why gradient descent is parametrization-sensitive, consider the model which is used at some current point $\theta_k$ to determine the direction of next optimization step:
$$\begin{cases}
f(q_{\theta_k}) + \langle \nabla_\theta f(q_{\theta_k}), \delta \theta \rangle \to \min\limits_{\delta \theta} \\
\|\delta \theta\|_2^2 < \alpha_k
\end{cases}$$
where $\alpha_k$ is learning rate at step $k$. Being first-order method, gradient descent constructs a <<model>> which approximates $F$ locally around $\theta_k$ using first-order Taylor expansion and employs standard Euclidean metric to determine a region of trust for this model. Then this surrogate task is solved analytically to obtain well-known update formula:
$$\delta \theta \propto -\nabla_\theta f(q_{\theta_k})$$

The issue arises from reliance on Eucliden metric in the space of parameters. In most parametrizations, small changes in parameters space do not guarantee small change in distribution space and vice versa: some small changes in distribution may demand big steps in parameters space\footnote{classic example is that $\mathcal{N}(0, 100)$ is similar to $\mathcal{N}(1, 100)$ while $\mathcal{N}(0, 0.1)$ is completely different from $\mathcal{N}(1, 0.1)$, although Euclidean distance in parameter space is the same for both pairs.}.

Natural gradient proposes to use another metric, which achieves invariance to parametrization of distribution $q$ using the properties of Fisher matrix:
\begin{definition}
For distribution $q_\theta$ \emph{Fisher matrix} $F_q(\theta)$ is defined as
$$F_q(\theta) \coloneqq \mathbb{E}_{x \sim q} \nabla_\theta \log q_\theta(x) (\nabla_\theta \log q_\theta(x))^T$$
\end{definition}

Note that Fisher matrix depends on parametrization. Yet for any parametrization it is guaranteed to be positive semi-definite by definition. Moreover, it induces a so-called \emph{Riemannian metric}\footnote{in Euclidean space the general form of scalar product is $\langle x, y \rangle \coloneqq x^TGy$, where $G$ is fixed positive semi-definite matrix. The metric induced by this scalar product is correspondingly $d(x, y)^2 \coloneqq  (y - x)^T G (y - x)$. The difference in Riemannian space is that $G$, called \emph{metric tensor}, depends on $x$, so the relative distance may vary for different points. It is used to describe the distances between points on manifolds and holds important properties which Fisher matrix inherits as metric tensor for distribution space.} in the space of parameters:
$$d(\theta_1, \theta_2)^2 \coloneqq (\theta_2 - \theta_1)^T F_q(\theta_1) (\theta_2 - \theta_1)$$

In natural gradient descent it is proposed to use this metric instead of Euclidean:
$$\begin{cases}
f(q_{\theta_k}) + \langle \nabla_\theta f(q_{\theta_k}), \delta \theta \rangle \to \min\limits_{\delta \theta} \\
\delta \theta^T F_q(\theta_k) \delta \theta < \alpha_k
\end{cases}$$
This surrogate task can be solved analytically to obtain the following optimization direction:
\begin{equation}\label{naturalgradient}
\delta \theta \propto -F_q(\theta_k)^{-1} \nabla_\theta f(q_{\theta_k})
\end{equation}
The direction of gradient descent is corrected by Fisher matrix which concerns the scale across different axes. This direction, specified by $F_q(\theta_k)^{-1} \nabla_\theta f(q_{\theta_k})$, is called \emph{natural gradient}.

Let's discuss why this new metric really provides us invariance to distribution parametrization. We already obtained natural gradient for $q$ being parameterized by $\theta$ (\ref{naturalgradient}). Assume that we have another parametrization $q_\nu$. These new parameters $\nu$ are somehow related to $\theta$; we suppose there is some functional dependency $\theta(\nu)$, which we assume to be differentiable with jacobian $J$. In this notation:
\begin{equation}\label{jacobian}
\delta \theta = J \delta \nu \text{,} \qquad J_{ij} \coloneqq \frac{\partial \theta_i}{\partial \nu_j}
\end{equation}

The central property of Fisher matrix, which provides the desired invariance, is the following:
\begin{proposition}
If $\theta = \theta(\nu)$ with jacobian $J$, then reparametrization formula for Fisher matrix is
\begin{equation}\label{Fisherreparam}
F_q(\nu) = J^T F_q(\theta) J
\end{equation}
\end{proposition}

Now it can be derived that natural gradient for parametrization with $\nu$ is the same as for $\theta$. If we want to calculate natural gradient in terms of $\nu$, then our step is, according to (\ref{jacobian}):
\begin{align*}
\delta \theta = J \delta \nu &= \\
\{ \text{natural gradient in terms of $\nu$} \} & \propto J F_q(\nu_k)^{-1} \nabla_\nu f(q_{\nu_k}) = \\
\{ \text{Fisher matrix reparametrization (\ref{Fisherreparam})} \} &= J\left( J^TF_q(\theta_k)J \right) ^{-1} \nabla_\nu f(q_{\nu_k}) \\
\{ \text{chain rule} \} &= J\left( J^TF_q(\theta_k)J \right) ^{-1} \nabla_\nu \theta(\nu_k)^T \nabla_\theta f(q_{\theta_k}) = \\
&= JJ^{-1}F_q(\theta_k)^{-1}J^{-T} J^T \nabla_\theta f(q_{\theta_k}) = \\
&= F_q(\theta_k)^{-1} \nabla_\theta f(q_{\theta_k})
\end{align*}
which can be seen to be the same as in (\ref{naturalgradient}).

Application of natural gradient descent in DRL setting is complicated in practice. Theoretically, the only change that must be done is scaling of gradient using inverse Fisher matrix (\ref{naturalgradient}). Yet, Fisher matrix requires $n^2$ memory and $\mathcal{O}(n^3)$ computational costs for inversion where $n$ is the number of parameters. For neural networks this causes the same complications as the application of second-order optimization methods.

K-FAC optimization method \citep{martens2015optimizing} provides a specific approximation form of Fisher matrix for neural networks with linear layers which can be efficiently computed, stored and inverted. Usage of K-FAC approximation allows to compute natural gradient directly using (\ref{naturalgradient}).

\subsection{Trust-Region Policy Optimization (TRPO)}

The main drawback of Actor-Critic algorithm is believed to be the abandonment of experience that was used for previous updates. As the number of updates required is usually huge, this is considered to be a substantial loss of information. Yet, it is not clear how this information can be effectively used for newer updates.

Suppose we want to make an update of $\pi(\theta)$, but using samples collected by some $\pi^{\old}$. The straightforward approach is importance sampling technique, which naive application to gradient formula (\ref{ACgradient}) yields the following result:
$$
\nabla_{\theta} J(\theta) = \E_{\Traj \sim \pi^{\old}} \frac{\mathcal{P}(\Traj \mid \pi(\theta))}{\mathcal{P}(\Traj \mid \pi^{\old})} \sum_{t=0} \nabla_{\theta} \log \pi_{\theta}(a_t \mid s_t) A^\pi(s_t, a_t)
$$
The emerged importance sampling weight is actually computable as transition probabilities cross out:
$$\frac{\mathcal{P}(\Traj \mid \pi(\theta))}{\mathcal{P}(\Traj \mid \pi^{\old})} = \frac{\prod_{t=1} \pi_\theta(a_t \mid s_t)}{\prod_{t=1} \pi^{\old}(a_t \mid s_t)}$$
The problem with this coefficient is that it tends either to be exponentially small or to explode. Even with some heuristic normalization of coefficients the batch gradient would become dominated by one or several transitions and destabilize the training procedure by introducing even more variance.

Notice that application of importance sampling to another representation of gradient (\ref{gradientshort}) yields seemingly different result:
\begin{equation}\label{policygradientISalt}
\nabla_{\theta} J(\theta) = \E_{\pi^{\old}} \frac{d_{\pi(\theta)}(s)}{d_{\pi^{\old}}(s)} \frac{\pi_\theta(a \mid s)}{\pi^{\old}(a \mid s)} \nabla_{\theta} \log \pi_{\theta}(a \mid s) A^\pi(s, a)
\end{equation}
Here we avoided common for the whole trajectories importance sampling weights by using the definition of state visitation frequencies. But this result is even less practical as these frequencies are unknown to us.

The first key idea behind the theory concerning this problem is that may be these importance sampling coefficients behave more stable if the policies $\pi^{\old}$ and $\pi(\theta)$ are in some terms <<close>>. Intuitively, in this case $\frac{d_{\pi(\theta)}(s)}{d_{\pi^{\old}}(s)}$ of formula (\ref{policygradientISalt}) is close to 1 as state visitation frequencies are similar, and the remained importance sampling coefficient becomes acceptable in practice. And if some two policies are similar, their values of our objective (\ref{RLgoal}) are probably close too.

\allowdisplaybreaks

For any two policies, $\pi$ and $\pi^{\old}$:
\begin{align*}
J(\pi) - J(\pi^{\old}) &= \mathbb{E}_{\mathcal{T} \sim \pi} \sum_{t=0} \gamma^t r(s_t) - J(\pi^{\old}) = \\
&= \mathbb{E}_{\mathcal{T} \sim \pi} \sum_{t=0} \gamma^t r(s_t) - V^{\pi^{\old}}(s_0) = \\
&= \mathbb{E}_{\mathcal{T} \sim \pi} \left[\sum_{t=0} \gamma^t r(s_t) - V^{\pi^{\old}}(s_0) \right] = \\
\{ \text{trick $\textstyle \sum_{t = 0}^\infty \left( a_{t+1} - a_t \right) = -a_0$\footnotemark} \} &= \mathbb{E}_{\mathcal{T} \sim \pi} \left[\sum_{t=0} \gamma^t r(s_t) + \sum_{t=0} \left[ \gamma^{t+1} V^{\pi^{\old}}(s_{t+1}) - \gamma^t V^{\pi^{\old}}(s_t) \right] \right] = \\
\{ \text{regroup} \} &= \mathbb{E}_{\mathcal{T} \sim \pi} \sum_{t=0} \gamma^t \left( r(s_t) + \gamma V^{\pi^{\old}}(s_{t+1}) - V^{\pi^{\old}}(s_t) \right) = \\
\{ \text{by definition (\ref{QV})} \} &= \mathbb{E}_{\mathcal{T} \sim \pi} \sum_{t=0} \gamma^t \left( Q^{\pi^{\old}}(s_t, a_t) - V^{\pi^{\old}}(s_t) \right) \\
\{ \text{by definition (\ref{advantage})} \} &= \mathbb{E}_{\mathcal{T} \sim \pi} \sum_{t=0} \gamma^t A^{\pi^{\old}}(s_t, a_t)
\end{align*}
\footnotetext{and if MDP is episodic, for terminal states $V^{\pi^{\old}}(s_T) = 0$ by definition.}

The result obtained above is often referred to as \emph{relative policy performance identity} and is actually very interesting: it states that we can substitute reward with advantage function of arbitrary policy and that will shift the objective by the constant.

Using the discounted state visitation frequencies definition \ref{dsvf}, relative policy performance identity can be rewritten as
\begin{equation*}
J(\pi) - J(\pi^{\old}) = \frac{1}{1 - \gamma}\mathbb{E}_{s \sim d_\pi(s)} \mathbb{E}_{a \sim \pi(a \mid s)} A^{\pi^{\old}}(s, a)   
\end{equation*}

Now assume we want to optimize parameters $\theta$ of policy $\pi$ while using data collected by $\pi^{\old}$: applying importance sampling in the same manner:
\begin{equation*}
J(\pi_\theta) - J(\pi^{\old}) = \frac{1}{1 - \gamma}\mathbb{E}_{s \sim d_{\pi^{\old}}(s)} \frac{d_{\pi_\theta}(s)}{d_{\pi^{\old}}(s)} \mathbb{E}_{a \sim \pi^{\old}(a \mid s)} \frac{\pi_\theta(a \mid s)}{\pi^{\old}(a \mid s)} A^{\pi^{\old}}(s, a)   
\end{equation*}

As we have in mind the idea of $\pi^{\old}$ being close to $\pi_\theta$, the question is how well this identity can be approximated if we assume $d_{\pi_\theta}(s) = d_{\pi^{\old}}(s)$. Under this assumption:
\begin{equation*}
J(\pi_\theta) - J(\pi^{\old}) \approx L_{\pi^{\old}}(\theta) \coloneqq \frac{1}{1 - \gamma}\mathbb{E}_{s \sim d_{\pi^{\old}}(s)} \mathbb{E}_{a \sim \pi^{\old}(a \mid s)} \frac{\pi_\theta(a \mid s)}{\pi^{\old}(a \mid s)} A^{\pi^{\old}}(s, a)   
\end{equation*}

The point is that interaction using $\pi^{\old}$ corresponds to sampling from the expectations presented in $L_{\pi^{\old}}(\theta)$:
$$L_{\pi^{\old}}(\theta) = \mathbb{E}_{\pi^{\old}} \frac{\pi_\theta(a \mid s)}{\pi^{\old}(a \mid s)} A^{\pi^{\old}}(s, a)$$

The approximation quality of $L_{\pi^{\old}}(\theta)$ can be described by the following theorem:
\begin{proposition}\label{Lapproximationestimation}\citep{schulman2015trust}
\begin{equation*} 
\left| J(\pi_\theta) - J(\pi^{\old}) - L_{\pi^{\old}}(\theta) \right| \le C \max_s \KL(\pi^{\old} \parallel \pi_\theta)[s]
\end{equation*}
where $C$ is some constant and $\KL(\pi^{\old} \parallel \pi_\theta)[s]$ is a shorten notation for $\KL(\pi^{\old}(a \mid s) \parallel \pi_\theta(a \mid s))$.
\end{proposition}

There is an important corollary of proposition \ref{Lapproximationestimation}:
$$J(\pi_\theta) - J(\pi^{\old}) \ge L_{\pi^{\old}}(\theta) - C \max_s \KL(\pi^{\old} \parallel \pi_\theta)[s]$$
which not only states that expression on the right side represents a \emph{lower bound}, but also that the optimization procedure
\begin{equation}\label{lowerbound}
\theta_{k+1} = \argmax_{\theta} \left[ L_{\pi_{\theta_k}}(\theta) - C \max_s \KL(\pi_{\theta_k} \parallel \pi_\theta)[s] \right]
\end{equation}
will yield a policy with guaranteed monotonic improvement\footnote{the maximum of lower bound is non-negative as its value for $\theta = \theta_k$ equals zero, which causes $J(\pi_{k+1}) - J(\pi_k) \ge 0$.}. 

In practice there are several obstacles which preserve us from obtaining such procedure. First of all, our advantage function estimation is never precise. Secondly, it is hard to estimate precise value of constant $C$. One last obstacle is that it is not clear how to calculate $\KL$-divergence in its maximal form (with $\max$ taken across all states).

In Trust-Region policy optimization \citep{schulman2015trust} the idea of practical algorithm, approximating procedure (\ref{lowerbound}), is analyzed. To address the last issue, the naive approximation is proposed to substitute $\max$ with averaging across states\footnote{the distribution from which the states come is set to be $d_{\pi^{\old}}(s)$ for convenience as this is the distribution from which they come in $L_{\pi^{\old}}(\theta)$.}:
$$\max_s \KL(\pi^{\old} \parallel \pi_\theta)[s] \approx \E_{s \sim d_{\pi^{\old}}(s)} \KL(\pi^{\old} \parallel \pi_\theta)[s]$$

The second step of TRPO is to rewrite the task of unconstrained minimization (\ref{lowerbound}) in equivalent constrained (<<trust-region>>) form\footnote{the unconstrained objective is Lagrange function for constrained form.} to incorporate the unknown constant $C$ into learning rate:
\begin{equation}\label{trustregion}
\begin{cases}
L_{\pi^{\old}}(\theta) \to \max\limits_\theta \\
\E_{s \sim d(s \mid \pi^{\old})} \KL(\pi^{\old} \parallel \pi_\theta)[s] < C
\end{cases}
\end{equation}
Note that this rewrites an update iteration in terms of optimization methods: while $L_{\pi^{\old}}(\theta)$ is an approximation of true objective $J(\pi_\theta) - J(\pi^{\old})$, the constraint sets the region of trust to the surrogate. Remark that constraint is actually a divergence in policy space, i.~e. it is very similar to a metric in the space of distributions while the surrogate is a function of the policy and depends on parameters $\theta$ only through $\pi_\theta$.

To solve the constrained problem (\ref{trustregion}), the technique from convex optimization is used. Assume that $\pi^{\old}$ is a current policy and we want to update its parameters $\theta_k$. Then the objective of (\ref{trustregion}) is modeled using first-order Taylor expansion around $\theta_k$ while constraint is modeled using second-order \footnote{as first-order term is zero.} Taylor approximation:
$$\begin{cases}
L_{\pi^{\old}}(\theta_k + \delta \theta) \approx \langle \nabla_\theta \left. L_{\pi^{\old}}(\theta) \right|_{\theta_k} , \delta \theta \rangle \to \max\limits_{\delta \theta} \\
\E_{s \sim d(s \mid \pi^{\old})} \KL(\pi^{\old} \parallel \pi_{\theta_k + \delta \theta}) \approx \frac{1}{2}\E_{s \sim d(s \mid \pi^{\old})} \delta \theta^T \left. \nabla_\theta^2 \KL(\pi^{\old} \parallel \pi_\theta) \right|_{\theta_k} \delta \theta < C
\end{cases}
$$

It turns out, that this model is equivalent to natural policy gradient, discussed in sec. \ref{NPG}:
\begin{proposition}
$$\left. \nabla_\theta^2 \KL(\pi_\theta \parallel \pi^{\old})[s]  \right|_{\theta_k} = F_{\pi(a \mid s)}(\theta)$$
\end{proposition}
so $\KL$-divergence constraint can be approximated with metric induced by Fisher matrix. Moreover, the gradient of surrogate function is
\begin{align*}
\left. \nabla_\theta L_{\pi^{\old}}(\theta)  \right|_{\theta_k} &= \mathbb{E}_{\pi^{\old}} \frac{\left. \nabla_\theta \pi_\theta(a \mid s)  \right|_{\theta_k}}{\pi^{\old}(a \mid s)} A^{\pi^{\old}}(s, a) = \\
\\ \{ \pi^{\old} = \pi_{\theta_k} \} &= \mathbb{E}_{\pi^{\old}} \nabla_\theta \log \pi_{\theta_k}(a \mid s) A^{\pi^{\old}}(s, a)
\end{align*}
which is exactly an Actor-Critic gradient. Therefore the formula of update step is given by
$$\delta \theta \propto -F_\pi(\theta)^{-1} \nabla_\theta L_{\pi^{\old}}(\theta)$$
where $\nabla_\theta L_{\pi^{\old}}(\theta)$ coincides with standard policy gradient, and $F_\pi(\theta)$ is hessian of $\KL$-divergence:
$$F_\pi(\theta) \coloneqq \E_{s \sim d_{\pi^{\old}}(s)} \left. \nabla_\theta^2 \KL(\pi^{\old} \parallel \pi_\theta) \right|_{\theta_k}$$

In practical implementations $\KL$-divergence can be Monte-Carlo estimated using collected roll-out. The size of roll-out must be significantly bigger than in Actor-Critic to achieve sufficient precision of hessian estimation. Then to obtain a direction of optimization step the following system of linear equations
$$F_\pi(\theta) \delta \theta = -\nabla_\theta L_{\pi^{\old}}(\theta)$$
is solved using a \emph{conjugate gradients} method which is able to work with Hessian-vector multiplication procedure instead of requiring to calculate $F_\pi(\theta)$ explicitly.

TRPO also accompanies the update step with a \emph{line-search} procedure which dynamically adjusts step length using standard backtracking heuristic. As TRPO intuitively seeks for policy improvement on each step, the idea is to check whether the lower bound (\ref{lowerbound}) is positive after the biggest step allowed according to $\KL$-constraint and reduce the step size until it becomes positive.

Unlike Actor-Critic, TRPO performs extremely expensive complicated update steps but requires relatively small number of iterations in return. Of course, due to many approximations done, the overall procedure is only a resemblance of theoretically-justified iterations (\ref{lowerbound}) providing improvement guarantees.

\subsection{Proximal Policy Optimization (PPO)}\label{PPO}

Proximal Policy Optimization \citep{schulman2017proximal} proposes alternative heuristic way of performing lower bound (\ref{lowerbound}) optimization which demonstrated encouraging empirical results.

PPO still substitutes $\max\limits_s \KL$ on average, but leaves the surrogate in unconstrained form, suggesting to treat unknown constant $C$ as a hyperparameter:
\begin{equation}\label{lowerboundunconstrained}
\mathbb{E}_{\pi^{\old}} \left[ \frac{\pi_\theta(a \mid s)}{\pi^{\old}(a \mid s)} A^{\pi^{\old}}(s, a) - C \KL(\pi^{\old} \parallel \pi_\theta)[s] \right] \to \max_\theta
\end{equation}

The naive idea would be to straightforwardly optimize (\ref{lowerboundunconstrained}) as it is equivalent to solving the constraint trust-region task (\ref{trustregion}). To avoid Hessian-involved computations, one possible option is just to perform one step of first-order gradient optimization of (\ref{lowerboundunconstrained}). Such algorithm was empirically discovered to perform poorly as importance sampling coefficients $\frac{\pi_\theta(a \mid s)}{\pi^{\old}(a \mid s)}$ tend to unbounded growth.

In PPO it is proposed to cope with this problem in a simple old-fashioned way: by clipping. Let's denote by
$$r(\theta) \coloneqq \frac{\pi_\theta(a \mid s)}{\pi^{\old}(a \mid s)}$$
an importance sampling weight and by
$$r^{\operatorname{clip}}(\theta) \coloneqq \mathop{clip}(r(\theta), 1 - \epsilon, 1 + \epsilon)$$
its clipped version where $\epsilon \in (0, 1)$ is a hyperparameter. Then the clipped version of lower bound is:
\begin{equation}\label{PPOlowerbound}
\mathbb{E}_{\pi^{\old}} \left[ \min \left( r(\theta) A^{\pi^{\old}}(s, a), r^{\operatorname{clip}}(\theta) A^{\pi^{\old}}(s, a)\right) - C \KL(\pi^{\old} \parallel \pi_\theta)[s] \right] \to \max_\theta
\end{equation}

Here the minimum operation is introduced to guarantee that the surrogate objective remains a lower bound. Thus the clipping at $1 + \epsilon$ may occur only in the case if advantage is positive while clipping at $1 - \epsilon$ may occur if advantage is negative. In both cases, clipping represents a penalty for importance sampling weight $r(\theta)$ being too far from 1.

The overall procedure suggested by PPO to optimize the <<stabilized>> version of lower bound (\ref{PPOlowerbound}) is the following. A roll-out is collected using current policy $\pi^{\old}$ with some parameters $\theta$. Then the batches of typical size (as for Actor-Critic methods) are sampled from collected roll-out and several steps of SGD optimization of (\ref{PPOlowerbound}) proceed with respect to policy parameters $\theta$. During this process the policy $\pi^{\old}$ is considered to be fixed and new interaction steps are not performed, while in implementations there is no need to store old weights $\theta_k$ since everything required from $\pi^{\old}$ is to collect transitions and remember the probabilities $\pi^{\old}(a \mid s)$. The idea is that during these several steps we may use transitions from the collected roll-out several times. Similar alternative is to perform several epochs of training by passing through roll-out several times, as it is often done in deep learning.

Interesting fact discovered by the authors of PPO during ablation studies is that removing $\KL$-penalty term doesn't affect the overall empirical performance. That is why in many implementations PPO does not include $\KL$-term at all, making the final surrogate objective have a following form:
\begin{equation}\label{PPOobjective}
\mathbb{E}_{\pi^{\old}} \min \left( r(\theta) A^{\pi^{\old}}(s, a), r^{\operatorname{clip}}(\theta) A^{\pi^{\old}}(s, a)\right) \to \max_\theta
\end{equation}
Note that in this form the surrogate is not generally a lower bound and <<improvement guarantees>> intuition is lost.

\begin{algorithmbox}[label = PPOalgorithm]{Proximal Policy Optimization (PPO)}
\textbf{Hyperparameters:} $B$ --- batch size, $R$ --- rollout size, $\operatorname{n\_epochs}$ --- number of epochs, $\eps$ --- clipping parameter, $V^*_\phi$ --- critic neural network, $\pi_\theta$ --- actor neural network, $\alpha$ --- critic loss scaling, SGD optimizer.

\vspace{0.3cm}
Initialize weights $\theta, \phi$ arbitrary \\
\textbf{On each step:}
\begin{enumerate}
    \item obtain a roll-out of size $R$ using policy $\pi(\theta)$, storing action probabilities as $\pi^{\old}(a \mid s)$.
    \item for each transition $T$ from the roll-out compute advantage estimation (detached from computational graph to prevent backpropagation):
    $$A^\pi(T) = r' + \gamma V^\pi_\phi(s') - V^\pi_\phi$$
    \item perform $\operatorname{n\_epochs}$ passes through roll-out using batches of size $B$; \textbf{for each batch:}
    \begin{itemize}
    \item compute critic target (detached from computational graph to prevent backpropagation):
    $$y(T) = r' + \gamma V^\pi_\phi(s')$$
    \item compute critic loss:
    $$\Loss = \frac{1}{B}\sum_T \left( y(T) - V^\pi_\phi \right) ^2$$
    \item compute critic gradients:
    $$\nabla^{\operatorname{critic}} = \frac{\partial \Loss}{\partial \phi}$$
    \item compute importance sampling weights:
    $$r_\theta(T) = \frac{\pi_\theta(a \mid s)}{\pi^{\old}(a \mid s)}$$
    \item compute clipped importance sampling weights:
    $$r_\theta^{\operatorname{clip}}(T) = \mathop{clip}(r_\theta(T), 1 - \epsilon, 1 + \epsilon)$$
    \item compute actor gradient:
    $$\nabla^{\operatorname{actor}} = \frac{1}{B}\sum_T \nabla_\theta \min \left( r_\theta(T)A^\pi(T), r_\theta^{\operatorname{clip}}(T)A^\pi(T) \right)$$
    \item make a step of gradient descent using $\nabla^{\operatorname{actor}} + \alpha \nabla^{\operatorname{critic}}$
    \end{itemize}
\end{enumerate}
\end{algorithmbox}

\newpage

\section{Experiments}\label{experiments}

\subsection{Setup}

We performed our experiments using custom implementation of discussed algorithms attempting to incorporate best features from different official and unofficial sources and unifying all algorithms in a single library interface. The full code is available at \href{https://github.com/FortsAndMills/Learning-Reinforcement-Learning/tree/master/LRL}{our github}. 

While custom implementation might not be the most efficient, it hinted us several ambiguities in algorithms which are resolved differently in different sources. We describe these nuances and the choices made for our experiments in appendix \ref{implemdetails}.

For each environment we launch several algorithms to train the network with the same architecture with the only exception being the head which is specified by the algorithm (see table \ref{headstable}).

\begin{table}[h]
\setlength\extrarowheight{2pt}
\centering
\begin{tabular}{|c|c|}
\hline
DQN & Linear transformation to $|\A|$ arbitrary real values \\\hline
\multirow{3}{*}{Dueling} & First head: linear transformation to $|\A|$ arbitrary real values \\
& Second head: linear transformations to an arbitrary scalar \\
& Aggregated using dueling architecture formula (\ref{dueling}) \\\hline
Categorical & $|\A|$ linear transformations with softmax to $A$ values \\\hline
\multirow{3}{*}{Dueling Categorical} & First head: linear transformation to $|\A|$ arbitrary real values \\
& Second head: $|\A|$ linear transformations to $A$ arbitrary real values \\
& Aggregated using dueling architecture formula (\ref{rainbowdueling}) \\\hline
Quantile & $|\A|$ linear transformations to $A$ arbitrary real values \\\hline
\multirow{3}{*}{Dueling Quantile} & First head: linear transformation to $|\A|$ arbitrary real values \\
& Second head: $|\A|$ linear transformations to $A$ arbitrary real values \\
& Aggregated using dueling architecture formula (\ref{rainbowdueling}) without softmax \\\hline
\multirow{2}{*}{A2C / PPO} & Actor head: linear transformation with softmax to $|\A|$ values \\
& Critic head: linear transformation to scalar value \\\hline
\end{tabular}
\caption{Heads used for different algorithms. Here $|\A|$ is the number of actions and $A$ is the chosen number of atoms.}
\label{headstable}
\end{table}

For noisy networks all fully-connected layers in the feature extractor and in the head are substituted with noisy layers, doubling the number of their trained parameters. Both usage of noisy layers and the choice of the head influences the total number of parameters trained by the algorithm.

As practical tuning of hyperparameters is computationally consuming activity, we set all hyperparameters to their recommended values while trying to share the values of common hyperparameters among algorithms without affecting overall performance. 

We choose to give each algorithm same amount of interaction steps to provide the fair comparison of their sample efficiency. Thus the wall-clock time, number of episodes played and the number of network parameters updates varies for different algorithms.

\subsection{Cartpole}

Cartpole from OpenAI Gym \citep{brockman2016openai} is considered to be one of the simplest environments for DRL algorithms testing. The state is described with 4 real numbers while action space is two-dimensional discrete.

The environment rewards agent with +1 each tick until the episode ends. Poor action choices lead to early termination. The game is considered solved if agent holds for 200 ticks, therefore 200 is maximum reward in this environment.

In our first experiment we launch algorithms for 10\,000 interaction steps to train a neural network on the Cartpole environment. The network consists of two fully-connected hidden layers with 128 neurons and an algorithm-specific head. We used ReLU for activations. The results of a single launch are provided\footnote{we didn't tune hyperparameters for each of the algorithms, so the configurations used might not be optimal.} in table \ref{cartpoleresults}.

\begin{table}[h]
\setlength\extrarowheight{2pt}
\centering
\begin{tabular}{lrrr}
\toprule
{} &  Reached 200 &  Average reward &  Average FPS \\
\midrule
%Categorical Backwards DQN                          &         32.0 &          134.95 &    74.02 \\
%Backwards DQN                                      &         31.0 &          132.77 &    97.24 \\
Double DQN                                         &         23.0 &          126.17 &    95.78 \\
Dueling Double DQN                                 &         27.0 &          121.78 &    62.65 \\
DQN                                                &         33.0 &          116.27 &   101.53 \\
Categorical DQN                                    &         28.0 &          110.87 &    74.95 \\
Prioritized Double DQN                             &         37.0 &          110.52 &    85.58 \\
Categorical Prioritized Double DQN                 &         46.0 &          104.86 &    66.00 \\
Quantile Prioritized Double DQN                    &         42.0 &          100.76 &    68.62 \\
Categorical DQN with target network                &         44.0 &           96.08 &    73.92 \\
Quantile Double DQN                                &         54.0 &           93.14 &    75.40 \\
Quantile DQN                                       &         70.0 &           88.12 &    77.93 \\
Categorical Double DQN                             &         42.0 &           81.25 &    70.90 \\
Noisy Quantile Prioritized Dueling DQN             &         86.0 &           74.13 &    21.41 \\
Twin DQN                                           &         57.0 &           71.14 &    52.51 \\
Noisy Double DQN                                   &         67.0 &           71.06 &    31.81 \\
Noisy Prioritized Double DQN                       &         94.0 &           67.34 &    30.72 \\
Quantile Regression Rainbow                        &        106.0 &           67.11 &    21.54 \\
Rainbow                                            &         91.0 &           64.01 &    20.35 \\
Noisy Quantile Prioritized Double DQN              &        127.0 &           63.01 &    28.27 \\
Noisy Categorical Prioritized Double DQN           &         63.0 &           62.04 &    27.81 \\
PPO with GAE                                       &        144.0 &           53.06 &   390.53 \\
Noisy Prioritized Dueling Double DQN               &        180.0 &           47.52 &    22.56 \\
PPO                                                &        184.0 &           45.19 &   412.88 \\
Noisy Categorical Prioritized Dueling Double DQN   &        428.0 &           22.09 &    20.63 \\
A2C                                                &           - &           12.30 &  1048.64 \\
%Quantile Backwards DQN                             &           - &           12.11 &    82.36 \\
A2C with GAE                                       &           - &           11.50 &   978.00 \\
\bottomrule
\end{tabular}
\caption{Results on Cartpole for different algorithms: number of episode when the highest score of 200 was reached, average reward across all played episodes and average number of frames processed in a second (FPS).}
\label{cartpoleresults}
\end{table}

\subsection{Pong}

We used Atari Pong environment from OpenAI Gym \citep{brockman2016openai} as our main testbed to study the behaviour of the following algorithms:
\begin{itemize}
    \item DQN --- Deep Q-learning (sec. \ref{DQN})
    \item c51 --- Categorical DQN (sec. \ref{CategoricalDQN})
    \item QR-DQN --- Quantile Regression DQN (sec. \ref{QRDQN})
    \item Rainbow (sec. \ref{rainbow})
    \item A2C --- Advantage Actor Critic (sec. \ref{A2C}) extended with GAE (sec. \ref{GAE})
    \item PPO --- Proximal Policy Optimization (sec. \ref{PPO}) extended with GAE (sec. \ref{GAE})
\end{itemize}

In Pong, each episode is split into rounds. Each round ends with player either winning or loosing. The episode ends when the player wins or looses 21 rounds. The reward is given after each round and is +1 for winning and -1 for loosing. Therefore the maximum total reward is 21 and the minimum is -21. Note that the flag $\done$ indicating episode ending is not provided to the agent after each round but only at the end of full game (consisting of 21-41 rounds). 

The standard preprocessing for Atari games proposed in DQN \citep{mnih2013playing} was applied to the environment (see table \ref{preprocessingtable}). Thus, state space is represented by $(84, 84)$ grayscale pixels input (1 channel with domain $[0, 255]$). Action space is discrete with $|\A| = 6$ actions.

\begin{table}[h]
\setlength\extrarowheight{2pt}
\centering
\begin{tabularx}{0.85\textwidth}{|c|C|}
\hline
NoopResetEnv & Do nothing first 30 frames of games to imitate the pause between game start and real player reaction. \\\hline
MaxAndSkipEnv & Each interaction steps takes 4 frames of the game to allow less frequent switch of action. Max is taken over 4 passed frames to obtain an observation. \\\hline
FireResetEnv & Presses <<Fire>> button at first frame to launch the game, otherwise screen remains frozen.  \\\hline
WarpFrame & Turns observation to grayscale image of size 84x84. \\\hline
\end{tabularx}
\caption{Atari Pong preprocessing}
\label{preprocessingtable}
\end{table}

\begin{figure}[h]
    \centering
    \includegraphics[width=0.9\textwidth]{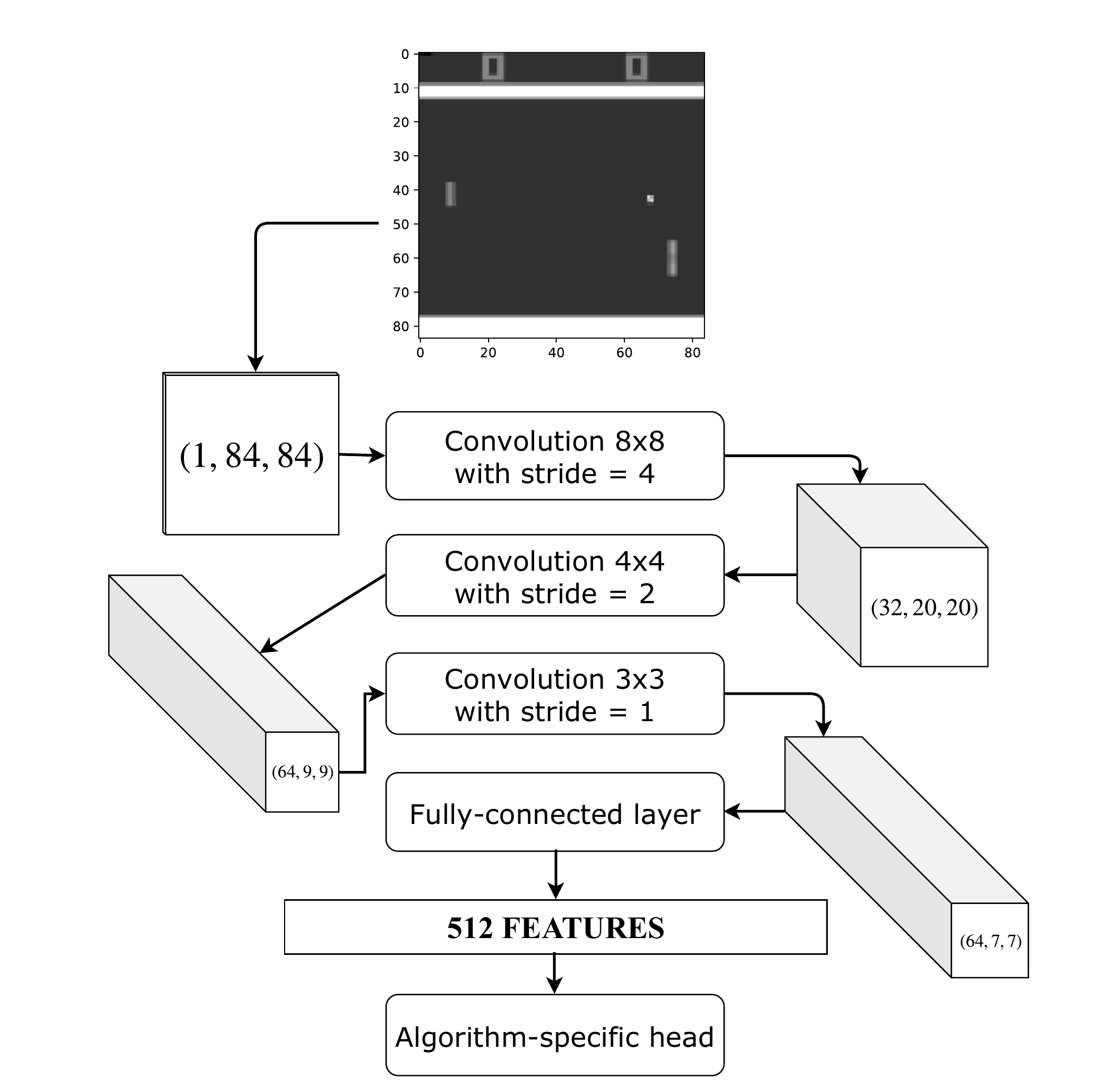}
    \caption{Network used for Atari Pong. All activation functions are ReLU. For Rainbow the fully-connected layer and all dense layers in the algorithm-specific head are substituted with noisy layers.}
    \label{PongCNN}
\end{figure}

 All algorithms were given 1\,000\,000 interaction steps to train the network with the same feature extractor presented on fig.~\ref{PongCNN}. The number of trained parameters is presented in table \ref{numofparameterstable}. All used hyperparameters are listed in table \ref{hyperparameterstable} in appendix \ref{HP}.
 
\begin{table}[h]
\setlength\extrarowheight{2pt}
\centering
\begin{tabularx}{0.5\textwidth}{|c|C|}
\hline
Algorithm & Number of trained parameters \\\hline
DQN         & 1\,681\,062  \\\hline
c51         & 1\,834\,962  \\\hline
QR-DQN      & 1\,834\,962  \\\hline
Rainbow     & 3\,650\,410  \\\hline
A2C         & 1\,681\,575  \\\hline
PPO         & 1\,681\,575  \\\hline
\end{tabularx}
\caption{Number of trained parameters in Pong experiment.}
\label{numofparameterstable}
\end{table}

\subsection{Interaction-training trade-off in value-based algorithms}

There is a common belief that policy gradient algorithms are much faster in terms of computational costs while value-based algorithms are preferable when simulation is expensive because of their sample efficiency. This follows from the nature of algorithms, as the fraction <<observations per network updates>> is extremely different for these two families: indeed, in DQN it is often assumed to perform one network update after each new transitions, while A2C collects about 32-40 observations for only one update. That makes the number of network updates performed during 1M steps interaction process substantially different and is the main reason of policy gradients speed rate.

Also policy gradient algorithms use several threads for parallel simulations (8 in our experiments) while value-based algorithms are formally single-threaded. Yet they can also enjoy multi-threaded interaction, in the simplest form by playing 1 step in all instances of environment and then performing $L$ steps of network optimization \citep{horgan2018distributed}. For consistency with single-threaded case it is reasonable to set the value of $L$ to be equal to the number of threads to maintain the same fraction <<observations per network updates>>. 

However it has been reported that lowering value of $L$ in two or four times can positively affect wall-clock time with some loss of sample efficiency, while raising batch size may mitigate this downgrade. The overall impact of such acceleration of value-based algorithms on performance properties is not well studied and may alter their behaviour.

In our experiments on Pong it became evident that value-based algorithms perform extensive amount of redundant network optimization steps, absorbing knowledge faster than novel information from new transitions comes in. This reasoning in particular follows from the success of PPO on Pong task which performs more than 10 times less network updates.

\begin{table}[h]
\setlength\extrarowheight{2pt}
\centering
\begin{tabularx}{0.75\textwidth}{c|C|C|}
            & Vanilla algorithm & Accelerated version \\\hline
Threads     & 1                 & 8   \\
Batch size  & 32                & 128 \\
$L$         & 1                 & 2   \\
\hline
\hline
Interactions per update & 1 & 4
\end{tabularx}
\caption{Setup for value-based acceleration experiment}
\label{ITtrade-off}
\end{table}

We compared two versions of value-based algorithms: \emph{vanilla version}, which is single-threaded with standard batch size (32) and $L = 1$ meaning that each observed transition is followed with one network optimization step, and \emph{accelerated version}, where 1 interaction step is performed in 8 parallel instances of environment and $L$ is set to be 2 instead of 8 which raises the fraction <<observations per training step>> in four times. To compensate this change we raised batch size in four times.

\begin{table}[h]
\setlength\extrarowheight{2pt}
\centering
\begin{tabular}{|c|c|c|c|c|}
\hline
& \multicolumn{2}{c|}{Interactions per update} & \multicolumn{2}{c|}{Average transitions per second} \\
Algorithm & vanilla & accelerated & vanilla & accelerated \\
\hline
DQN         & 1 & 4 & 55.74 & 168.43 \\\hline
c51         & 1 & 4 & 44.08 & 148.76 \\\hline
QR-DQN      & 1 & 4 & 47.46 & 155.97 \\\hline
Rainbow     & 1 & 4 & 19.30 & 70.22 \\\hline
A2C         & \multicolumn{2}{c|}{40} & \multicolumn{2}{c|}{656.25}  \\\hline
PPO         & \multicolumn{2}{c|}{10.33} & \multicolumn{2}{c|}{327.13} \\\hline
\end{tabular}
\caption{Computational efficiency of vanilla and accelerated versions.}
\label{accelerationtable}
\end{table}

\begin{figure}[h]
    \centering
    \includegraphics[width=\textwidth]{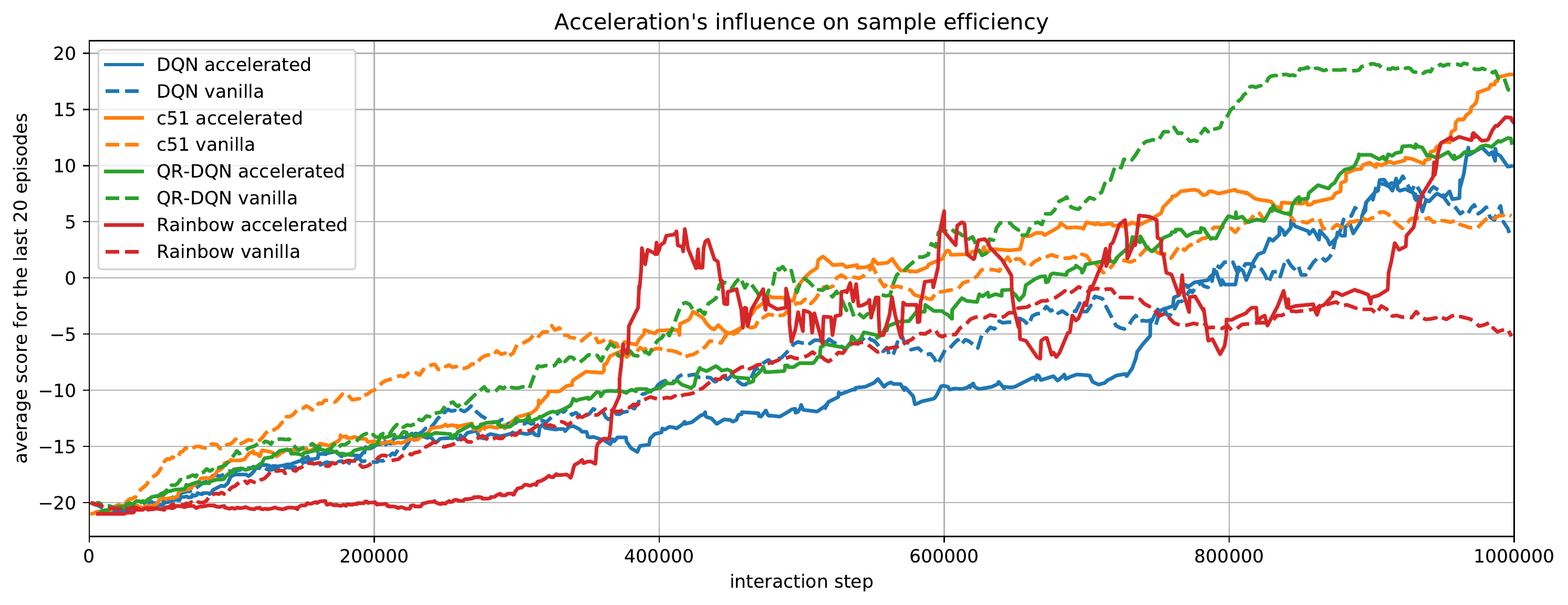}
    \caption{Training curves of vanilla and accelerated version of value-based algorithms on 1M steps of Pong. Although accelerated versions perform network updates four times less frequent, the performance degradation is not observed.}
    \label{AcceleratingResults}
\end{figure}

\begin{figure}[h]
    \centering
    \includegraphics[width=\textwidth]{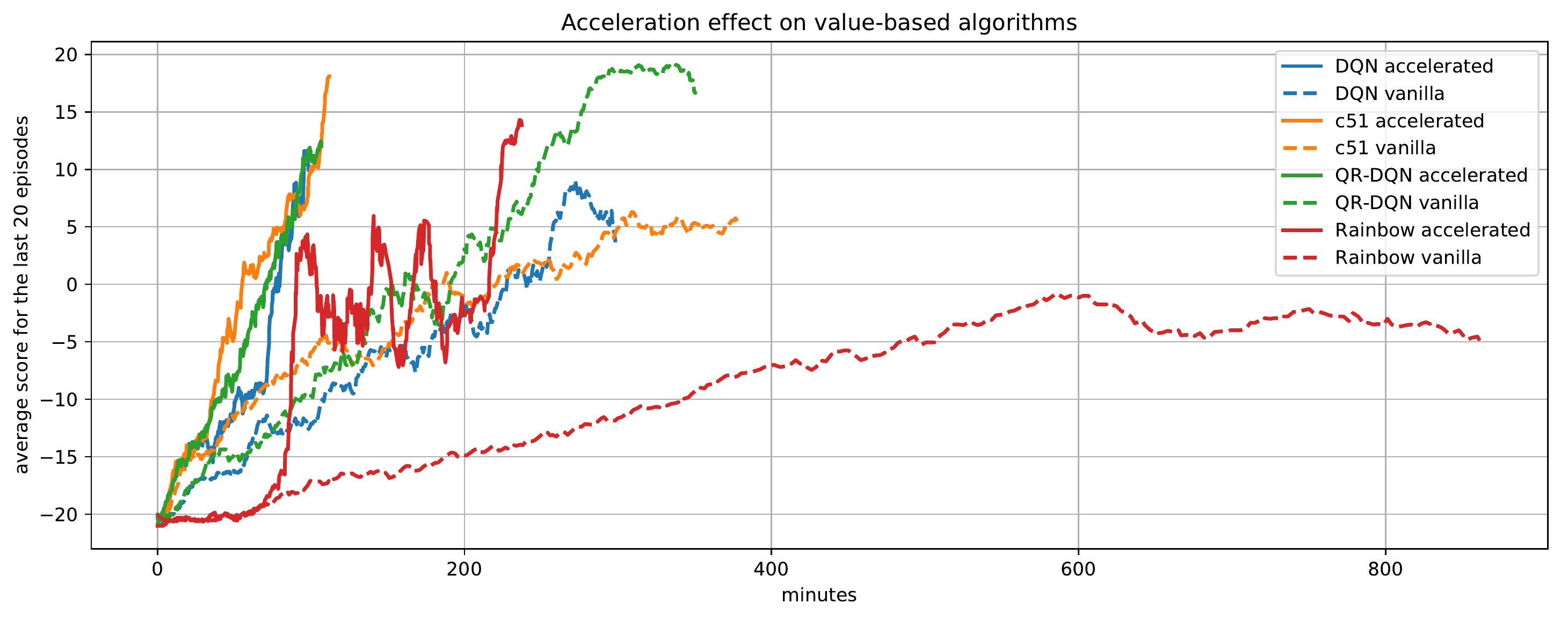}
    \caption{Training curves of vanilla and accelerated version of value-based algorithms on 1M steps of Pong from wall-clock time.}
    \label{AcceleratingClockResults}
\end{figure}

As expected, average speed of algorithms increases in approximately 3.5 times (see table \ref{accelerationtable}). We provide training curves with respect to 1M performed interaction steps on fig. \ref{AcceleratingResults} and with respect to wall-clock time on fig. \ref{AcceleratingClockResults}. The only vanilla algorithm that achieved better final score comparing to its accelerated rival is QR-DQN, while other three algorithms demonstrated both acceleration and performance improvement. The latter is probably caused by randomness as relaunch of algorithms within the same setting and hyperparameters can be strongly influenced by random seed.

It can be assumed that fraction <<observations per updates>> is an important hyperparameter of value-based algorithms which can control the trade-off between wall-clock time and sample efficiency. From our results it follows that low fraction leads to excessive network updates and may slow down learning in several times. Yet this hyperparameter can barely be tuned universally for all kinds of tasks opposed to many other hyperparameters that usually have their recommended default values. 

We stick further to the accelerated version and use its results in final comparisons. 

\subsection{Results}

\begin{figure}[h]
    \centering
    \includegraphics[width=\textwidth]{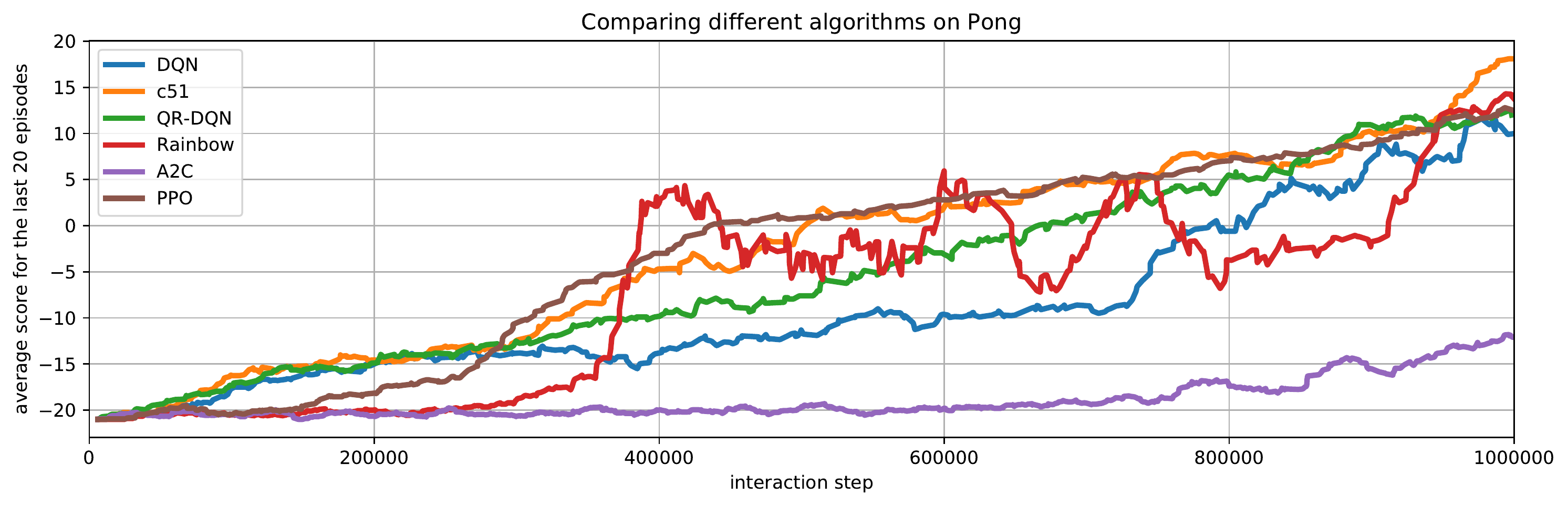}
    \caption{Training curves of all algorithms on 1M steps of Pong.}
    \label{Results}
\end{figure}

\begin{figure}[h]
    \centering
    \includegraphics[width=\textwidth]{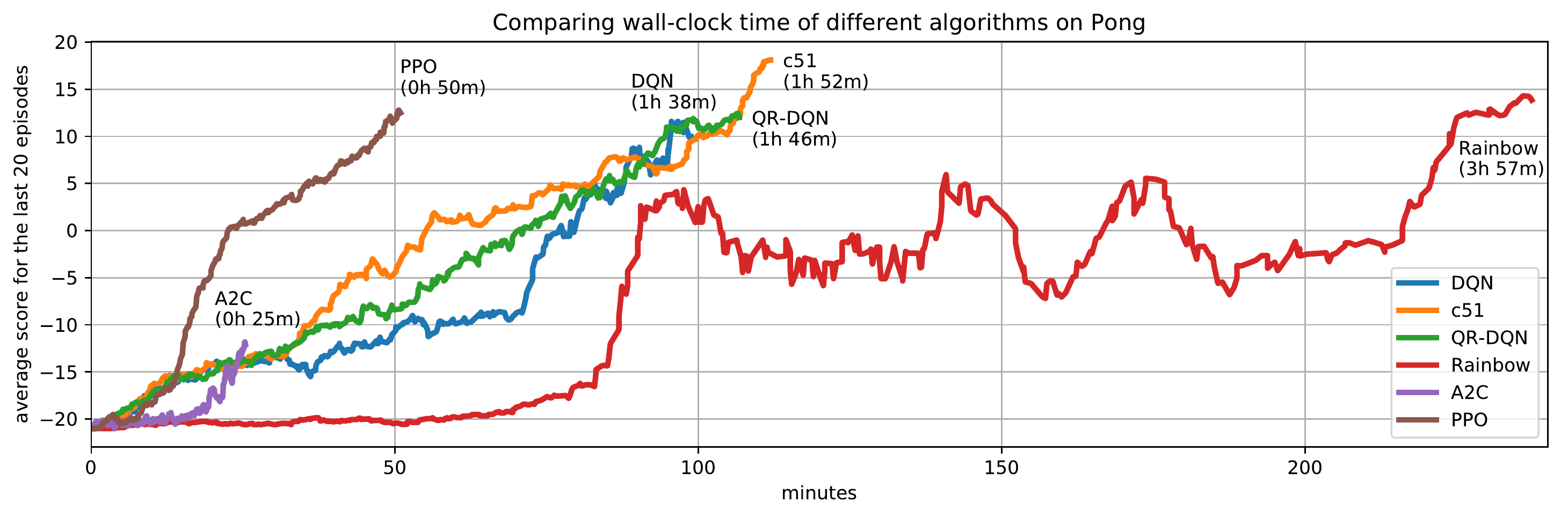}
    \caption{Training curves of all algorithms on 1M steps of Pong from wall-clock time.}
    \label{ClockResults}
\end{figure}

We compare the results of launch of six algorithms on Pong from two perspectives: sample efficiency (fig. \ref{Results}) and wall-clock time (fig. \ref{ClockResults}). We do not compare final performance of these algorithms as all six algorithms are capable to reach near-maximum final score on Pong given more iterations, while results after 1M iterations on a single launch significantly depend on chance.

All algorithms start with a warm-up session during which they try to explore the environment and learn first dependencies how the result of random behaviour can be surpassed. Epsilon-greedy with tuned parameters provides sufficient amount of exploration for DQN, c51 and QR-DQN whithout slowing down further learning while hyperparameter-free noisy networks are the main reason why Rainbow has substantially longer warm-up.

Policy gradient algorithms incorporate exploration strategy in stochasticity of learned policy but underutilization of observed samples leads to almost 1M-frames warm-up for A2C. It can be observed that PPO successfully mitigates this problem by reusing samples thrice. Nevertheless, both PPO and A2C solve Pong relatively quickly after the warm-up stage is over.

Value-based algorithm proved to be more computationally costly. QR-DQN and categorical DQN introduce more complicated loss computation, yet their slowdown compared to standard DQN is moderate. On the contrary, Rainbow is substantially slower mainly because of noise generation involvement. Furthermore, combination of noisy networks and prioritized replay results in even less stable training process.

We provide loss curves for all six algorithms and statistics for noise magnitude and prioritized replay for Rainbow in appendix \ref{TrainingstatsPong}; some additional visualizations of trained algorithms playing episodes of Pong are presented in appendix \ref{playingPong}.

\newpage

\section{Discussion}

We have concerned two main directions of universal model-free RL algorithm design and attempted to recreate several state-of-art pipelines.

While the extensions of DQN are reasonable solutions of evident DQN problems, their effect is not clearly seen on simple tasks like Pong\footnote{although it takes several hours to train, Pong is considered to be the easiest of 57 Atari games and one of the most basic testbeds for RL algorithms.}. Current state-of-art in single-threaded value-based approach, Rainbow DQN, is full of <<glue and tape>> decisions that might be not the most effective way of training process stabilization.

Distributional value-based approach is one of the cheapest in terms of resources extensions of vanilla DQN algorithm. Although it is reported to provide substantial performance improvement in empirical experiments, the reason behind this result remains unclear as expectation of return is the key quantity for agent's decision making while the rest of learned distribution does not affect his choices. One hypothesis to explain this phenomenon is that attempting to capture wider range of dependencies inside given MDP may provide auxiliary helping tasks to the algorithm, leading to better learning of expectation. Intuitively it seems that more reasonable switch of DQN to distributional setting would be learning the Bayesian uncertainty of expectation of return given observed data, but scalable practical algorithms within this orthogonal paradigm are yet to be created.

Policy gradient algorithms are aimed at direct optimization of objective and currently beat value-based approach in terms of computational costs. They tend to have less hyperparameters but are extremely sensitive to the choice of optimizer parameters and especially learning rate. We have affirmed the effectiveness of state-of-art algorithm PPO, which succeeded to solve Pong within an hour without hyperparameter tuning. Though on the one hand this algorithm was derived from TRPO theory, it essentially deviates from it and substitutes trust region updates with heuristic clipping. 

It can be observed in our results that PPO provides better gradients to the same network than DQN-based algorithms despite the absence of experience replay. While it is fair to assume that forgetting experienced transitions leads to information loss, it is also true that most observations stored in replay memory are already learned or contain no useful information. The latter makes most transitions in the sampled mini-batches insignificant, and, while prioritized replay attacks this issue, it might still be the case that current experience replay management techniques are imperfect.

There are still a lot of deviations of empirical results from theoretical perspectives. It is yet unclear which techniques are of the highest potential and what explanation lies behind many heuristic elements composing current state-of-art results. Possibly essential elements of modeling human-like reinforcement learning are yet to be unraveled as active research in this area promises substantial acceleration, generalization and stabilization of DRL algorithms.

\newpage
\bibliography{DRL}

\newpage
\begin{appendices}
\section{Implementation details}\label{implemdetails}
Here we describe several technical details of our implementation which may potentially influence the obtained results.

In most papers on value-based algorithms hyperparameters recommended for Atari games assume raw input in the range $[0, 255]$, while in various implementations of policy gradient algorithms normalized input in the range $[0, 1]$ is considered. Stepping aside from these agreements may damage the convergence speed both for value-based and policy gradient algorithms as the change of input domain requires hyperparameters retuning. 

We use MSE loss emerged in theoretical intuition for DQN while in many sources it is recommended to use Huber loss\footnote{Huber loss is defined as $$\Loss(y, \hat{y}) = \begin{cases}
(y - \hat{y})^2 & \text{if } |y - \hat{y}| < 1 \\
|y - \hat{y}| & \text{else}
\end{cases}$$} instead to stabilize learning.

In all value-based algorithms except c51 we update target network each $K$-th frame instead of exponential smoothing of its parameters as it is computationally cheaper. For c51 we remove target network heuristic as apriori limited domain prevents unbounded growth of predictions.

We do not architecturally force quantiles outputted by the network in Quantile Regression DQN to satisfy $\zeta_0 \le \zeta_1 \le \dots \le \zeta_{A-1}$. As in the original paper, we assume that all $A$ outputs of network are arbitrary real values and use a standard linear transformation as our last layer.

In dueling architectures we subtract mean of $A(s, a)$ across actions instead of theoretically assumed maximum as proposed by original paper authors.

We implement sampling from prioritized replay using SumTree data structure and in informal experiments affirmed the acceleration it provides. The importance sampling weight annealing $\beta(t)$ is represented by initial value $\beta(0) = \beta$ which is then linearly annealed to 1 during first $T_\beta$ frames; both $\beta$ and $T_\beta$ are hyperparameters.

We do not allow priorities $\Proba(T)$ to be greater than 1 by clipping as suggested in the original paper. This may mitigate the effect of prioritization replay but stabilizes the process. 

As importance sampling weights $w(T) = \frac{1}{B\Proba(T)}$ are potentially very close to zero, in original article it was proposed to normalize them on $\max w(T)$. In some implementations the maximum is taken over the whole experience replay while in others maximum is taken over current batch, which is not theoretically justified but computationally much faster. We stick to the latter option.

For noisy layers we use factorized noise sampling: for layer with $m$ inputs and $n$ outputs we sample $\eps_1 \in \R^n, \eps_2 \in \R^m$ from standard normal distributions and scale both using $f(\eps) = \operatorname{sign}(\eps)\sqrt{\eps}$. Thus we use $f(\eps_1)f(\eps_2)^T$ as our noise sample for weights matrix and $f(\eps_2)$ as noise sample for bias. All noise is shared across mini-batch. Noise is resampled on each forward pass through the network and thus is independent between evaluation, selection and interaction. Despite all these simplifications, we found noisy layers to be the most computationally expensive modification of DQN leading to substantial degradation of wall-clock time.

For policy gradient algorithms we add additional policy entropy term to the loss to force exploration. We also define actor loss as a scalar function that yields the same gradients as in the corresponding gradient estimation (\ref{ACgradient}) for A2C to compute it using PyTorch mechanics. For PPO objective (\ref{PPOobjective}) provides analogous <<actor loss>>; thus, in both policy gradient algorithms the full loss is defined as summation of actor, critic and entropy losses, with the two latter being scaled using scalar hyperparameters.

We use shared network architecture for policy gradient algorithms with one feature extractor and two heads, one for policy and one for critic.

$\KL$-penalty is not used in our PPO implementation. Also we do not normalize advantage estimations across the roll-out to zero mean and unit standard deviation as additionally done in some implementations.

We use PyTorch default initialization for linear and convolutional layers although orthogonal initialization of all layers is reported to be beneficial for policy gradient algorithms. Initial values of sigmas for noisy layers is set to be constant and equal to $\frac{\sigma_{\operatorname{init}}}{m}$ where $\sigma_{\operatorname{init}}$ is a hyperparameter and $m$ is the number of inputs 
in accordance with original paper.

We use Adam as our optimizer with default $\beta_1 = 0.9, \beta_2 = 0.999, \eps = 1\mathrm{e}{-8}$. No gradient clipping is performed.

\newpage
\section{Hyperparameters}\label{HP}

\begin{table}[h]
\setlength\extrarowheight{2pt}
\centering
\begin{tabular}{|c||c|c|c|c|c|c|}
\hline
 & DQN & QR-DQN & c51 & Rainbow & A2C & PPO \\\hline 
\hline
Reward discount factor $\gamma$ & \multicolumn{6}{c|}{0.99} \\\hline
$\eps(t)$-greedy strategy & \multicolumn{3}{c|}{$0.01 + 0.99e^{-\frac{t}{30\,000}}$} & - & \multicolumn{2}{c|}{-} \\\hline
Interactions per training step & \multicolumn{4}{c|}{4} & \multicolumn{2}{c|}{-} \\\hline
Batch size $B$ & \multicolumn{4}{c|}{128} & - & 32 \\\hline
Rollout capacity & \multicolumn{4}{c|}{-} & 40 & 1024 \\\hline
PPO number of epochs & \multicolumn{5}{c|}{-} & 3 \\\hline
Replay buffer initialization size\footnotemark & \multicolumn{4}{c|}{10\,000 transitions} & \multicolumn{2}{c|}{-}  \\\hline
Replay buffer capacity $M$ & \multicolumn{4}{c|}{1\,000\,000 transitions} & \multicolumn{2}{c|}{-}  \\\hline
Target network updates $K$ & \multicolumn{4}{c|}{each 1000-th step} & \multicolumn{2}{c|}{-}  \\\hline
Number of atoms $A$ & - & \multicolumn{3}{c|}{51} & \multicolumn{2}{c|}{-} \\\hline
$V_{\min}$, $V_{\max}$ & - & - & \multicolumn{2}{c|}{$[-10, 10]$} & \multicolumn{2}{c|}{-} \\\hline
Noisy layers std initialization & - & - & - & 0.5 & \multicolumn{2}{c|}{-} \\\hline
Multistep $N$ & - & - & - & 3 & \multicolumn{2}{c|}{-} \\\hline
Prioritization degree $\alpha$ & - & - & - & 0.5 & \multicolumn{2}{c|}{-} \\\hline
Prioritization bias correction $\beta$ & - & - & - & 0.4 & \multicolumn{2}{c|}{-} \\\hline
Unbiased prioritization after & - & - & - & 100\,000 steps & \multicolumn{2}{c|}{-} \\\hline
GAE coeff. $\lambda$ & \multicolumn{4}{c|}{-} & \multicolumn{2}{c|}{0.95} \\\hline
Critic loss weight & \multicolumn{4}{c|}{-} & \multicolumn{2}{c|}{0.5} \\\hline
Entropy loss weight & \multicolumn{4}{c|}{-} & \multicolumn{2}{c|}{0.01} \\\hline
PPO clip $\epsilon$ & \multicolumn{5}{c|}{-} & 0.1 \\\hline
Optimizer & \multicolumn{6}{c|}{Adam} \\\hline
Learning rate & \multicolumn{6}{c|}{0.0001} \\\hline
\end{tabular}
\caption{Selected hyperparameters for Atari Pong}
\label{hyperparameterstable}
\end{table}
\footnotetext{number of transitions to collect in replay memory before starting network optimization using mini-batch sampling.}

\newpage
\section{Training statistics on Pong}\label{TrainingstatsPong}

\begin{figure}[h]
    \centering
    \includegraphics[width=\textwidth]{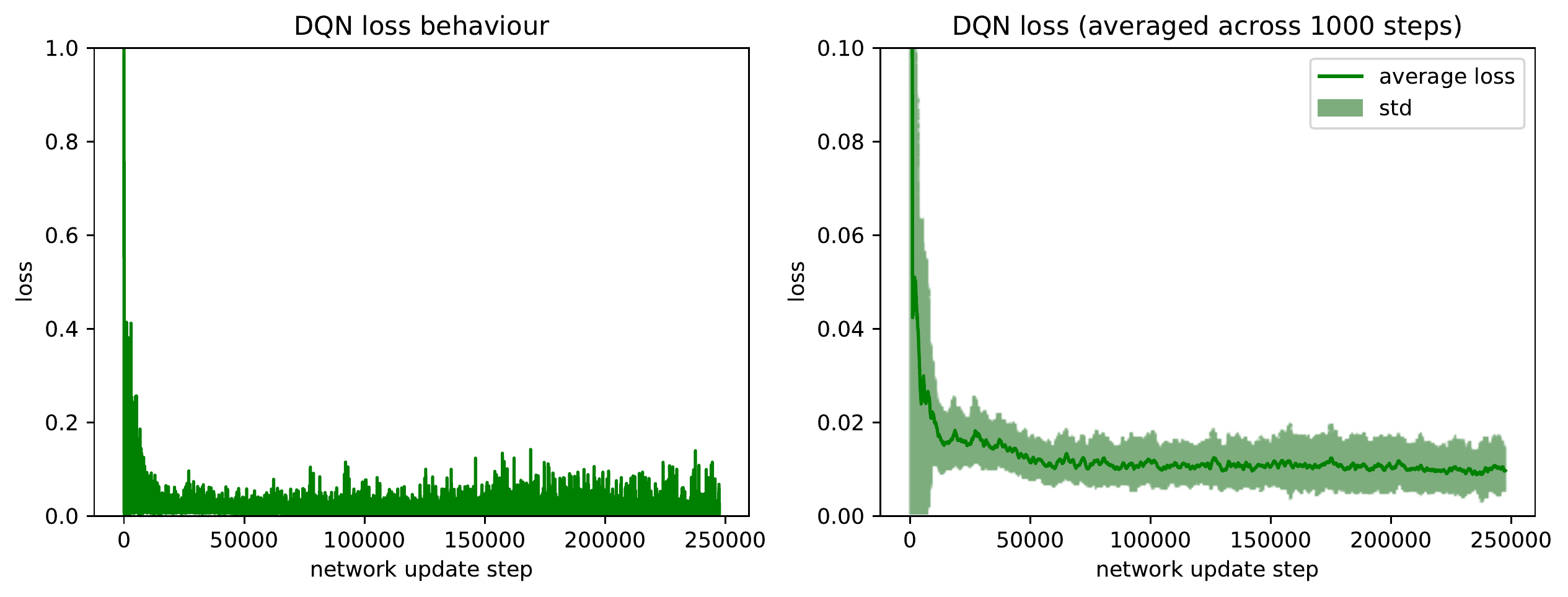}
    \caption{DQN loss behaviour during training on Pong.}
    \label{DQNloss}
\end{figure}

\begin{figure}[h]
    \centering
    \includegraphics[width=\textwidth]{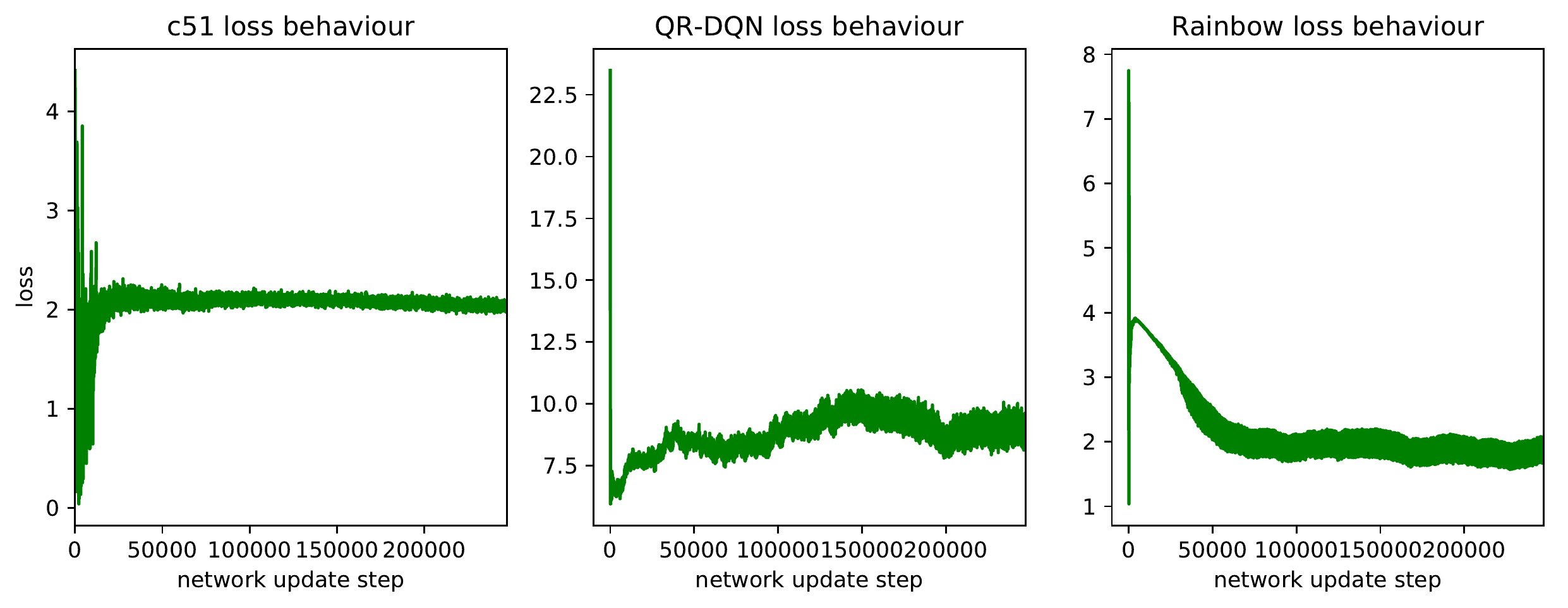}
    \caption{Loss behaviours of c51, QR-DQN and Rainbow during training on Pong.}
    \label{threelosses}
\end{figure}

\begin{figure}[h]
    \centering
    \includegraphics[width=\textwidth]{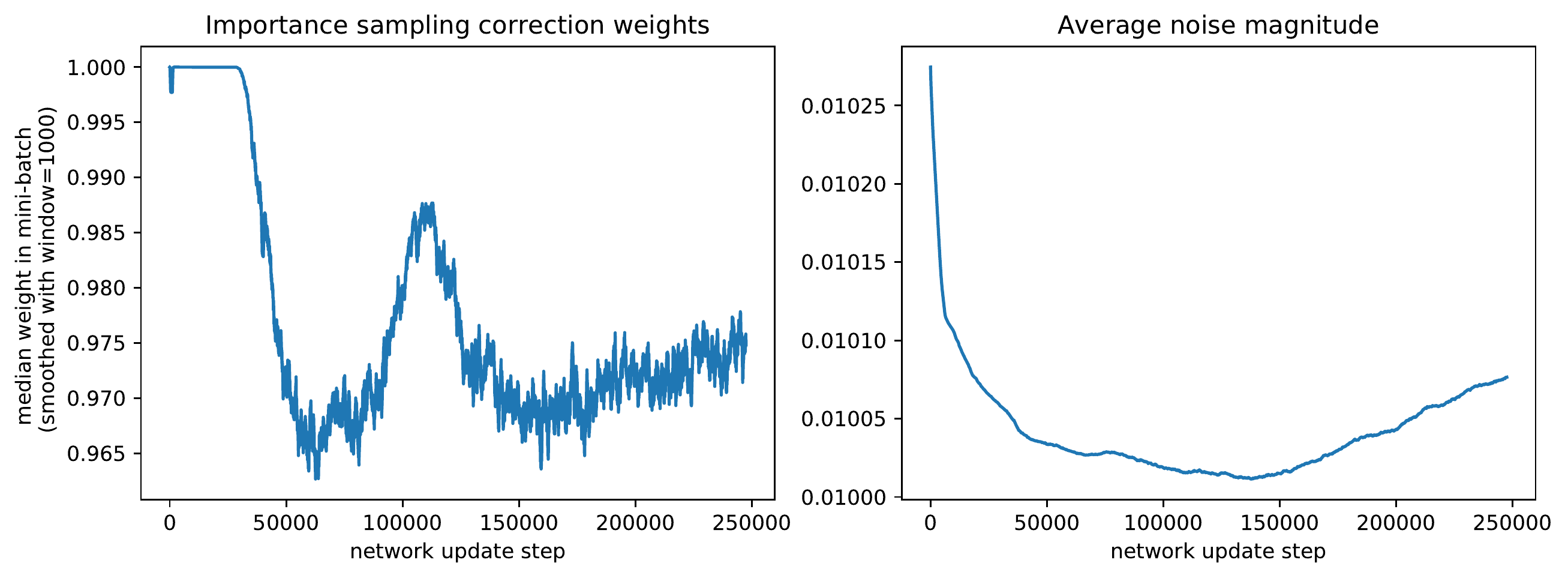}
    \caption{Rainbow statistics during training. Left: smoothed with window 1000 median of importance sampling weights from sampled mini-batches. Right: average noise magnitude logged at each 20-th step of training.}
    \label{rainbowstats}
\end{figure}

\begin{figure}[h]
    \centering
    \includegraphics[width=\textwidth]{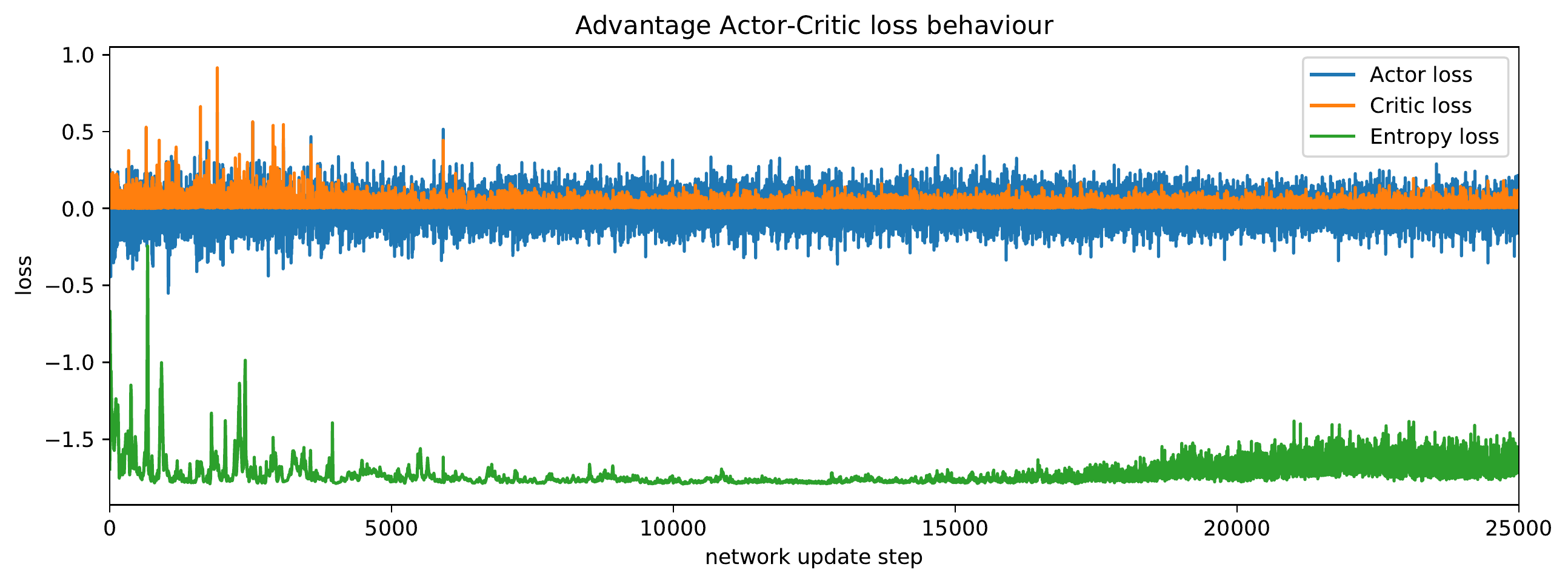}
    \caption{A2C loss behaviour during training.}
    \label{A2Closs}
\end{figure}

\begin{figure}[h]
    \centering
    \includegraphics[width=\textwidth]{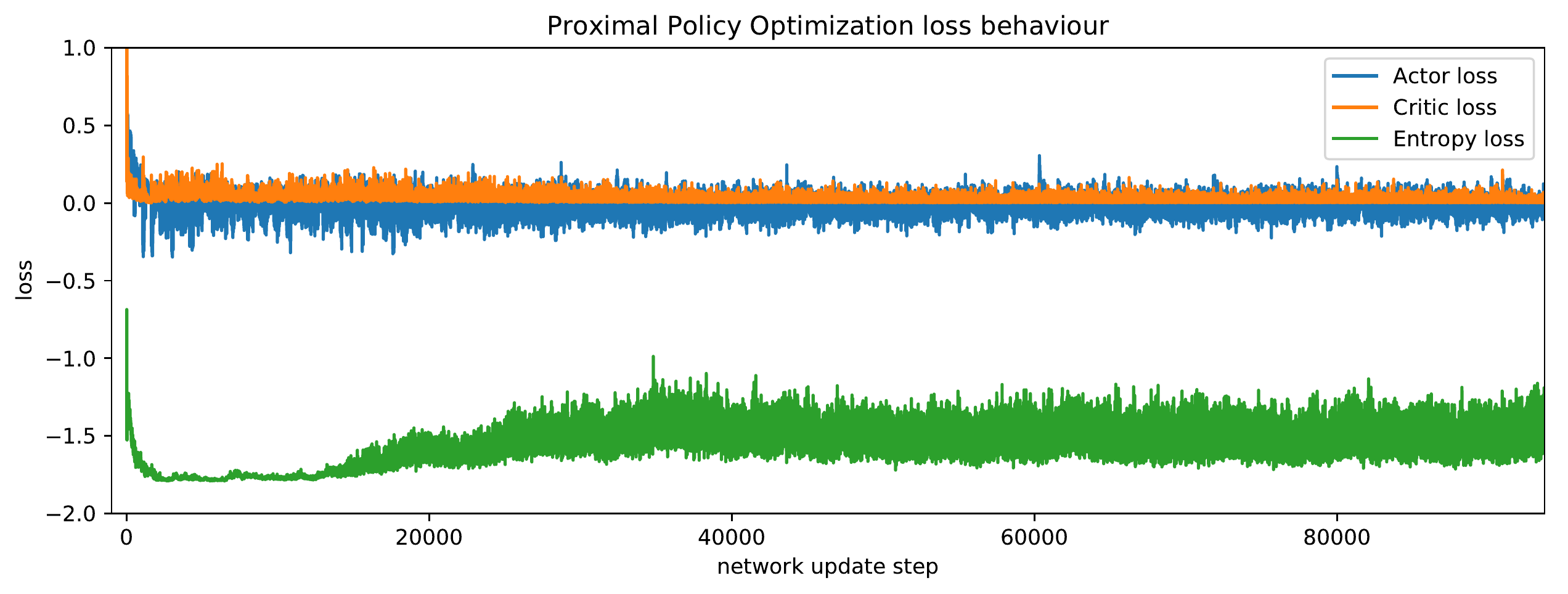}
    \caption{PPO loss behaviour during training.}
    \label{PPOloss}
\end{figure}

\newpage

\section{Playing Pong behaviour}\label{playingPong}

\begin{figure}[h]
    \centering
    \includegraphics[width=\textwidth]{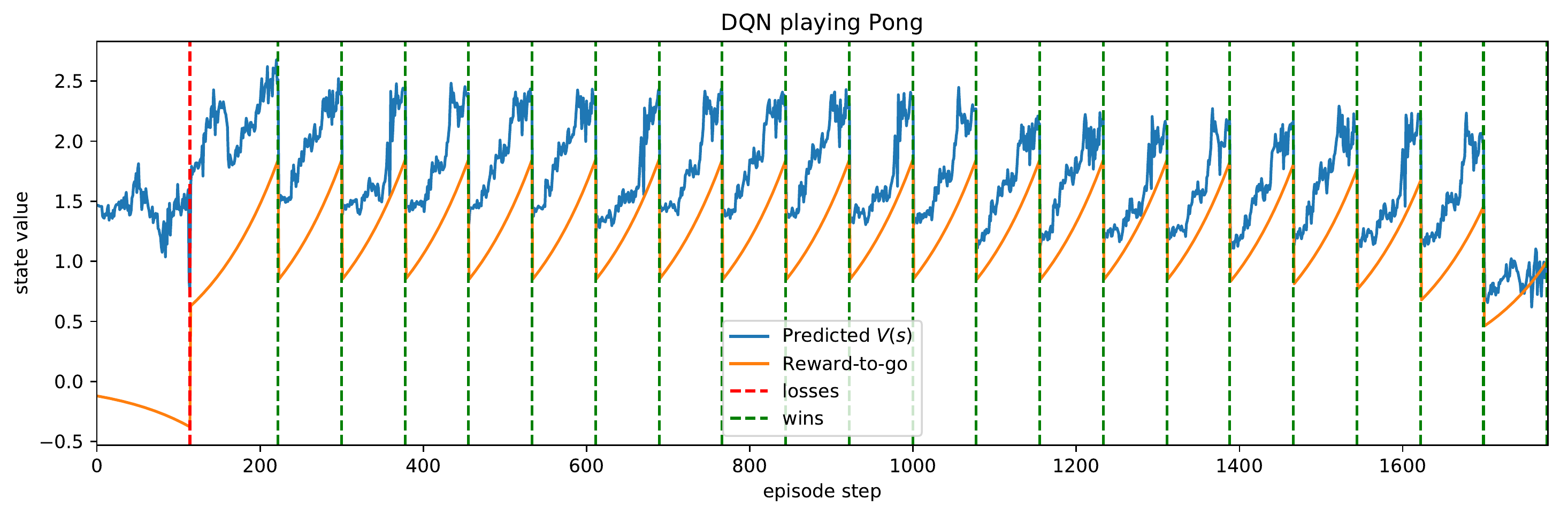}
    \caption{DQN playing one episode of Pong.}
    \label{DQNplay}
\end{figure}

\begin{figure}[h]
    \centering
    \includegraphics[width=\textwidth]{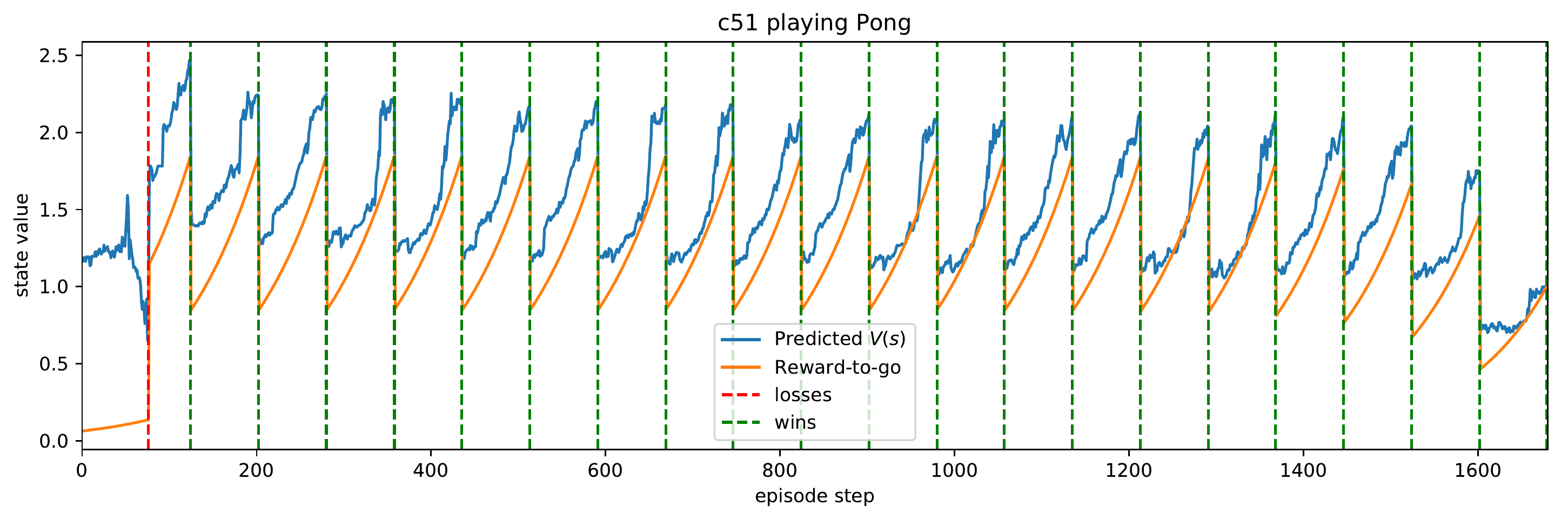}
    \caption{c51 playing one episode of Pong.}
    \label{c51play}
\end{figure}

\begin{figure}[h]
    \centering
    \includegraphics[width=\textwidth]{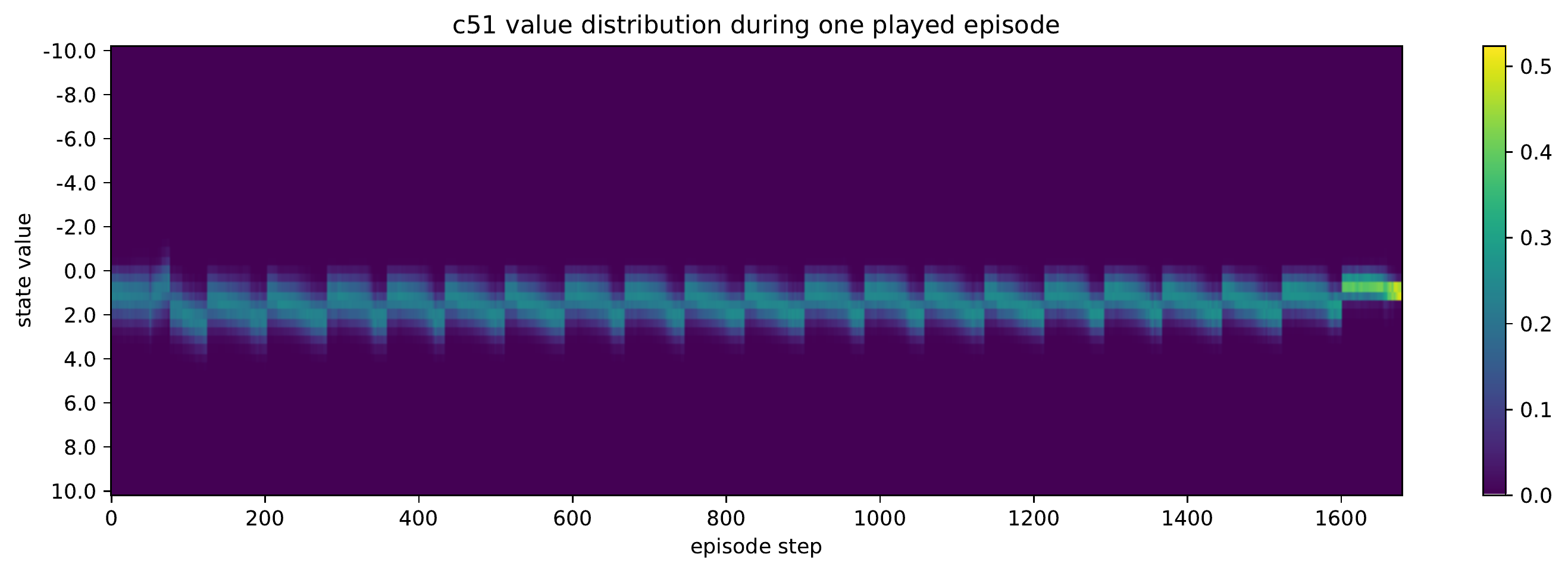}
    \caption{c51 value distribution prediction during one episode of Pong.}
    \label{c51playvd}
\end{figure}

\begin{figure}[h]
    \centering
    \includegraphics[width=\textwidth]{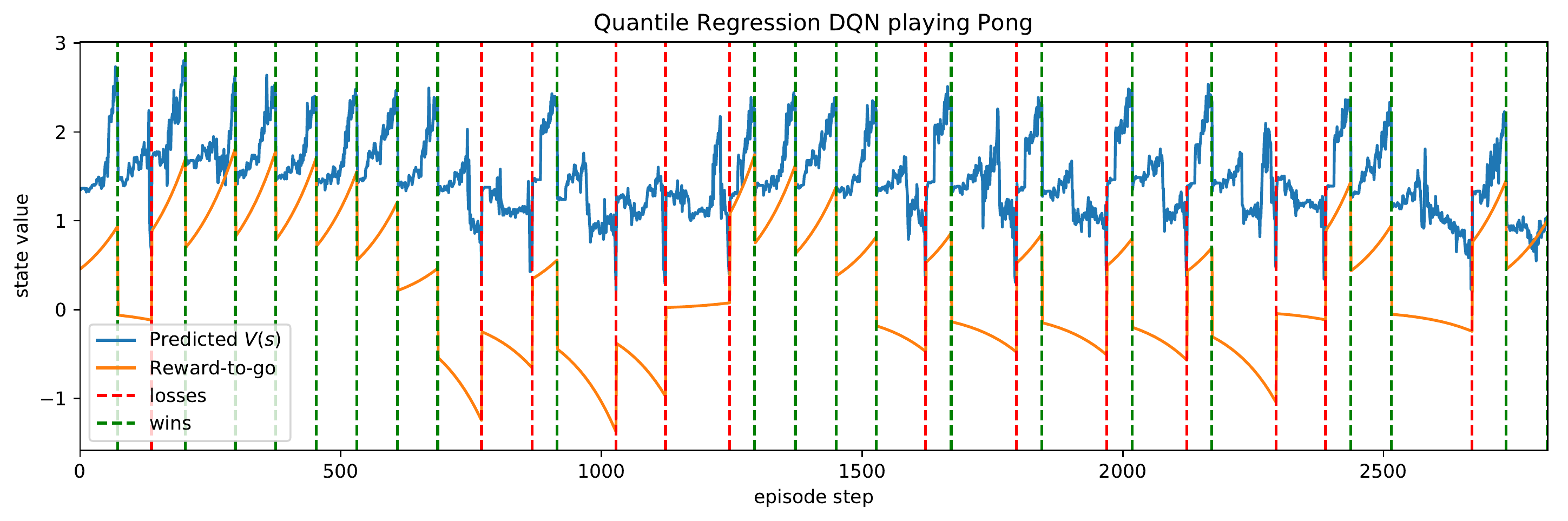}
    \caption{Quantile Regression DQN playing one episode of Pong.}
    \label{QRplay}
\end{figure}

\begin{figure}[h]
    \centering
    \includegraphics[width=\textwidth]{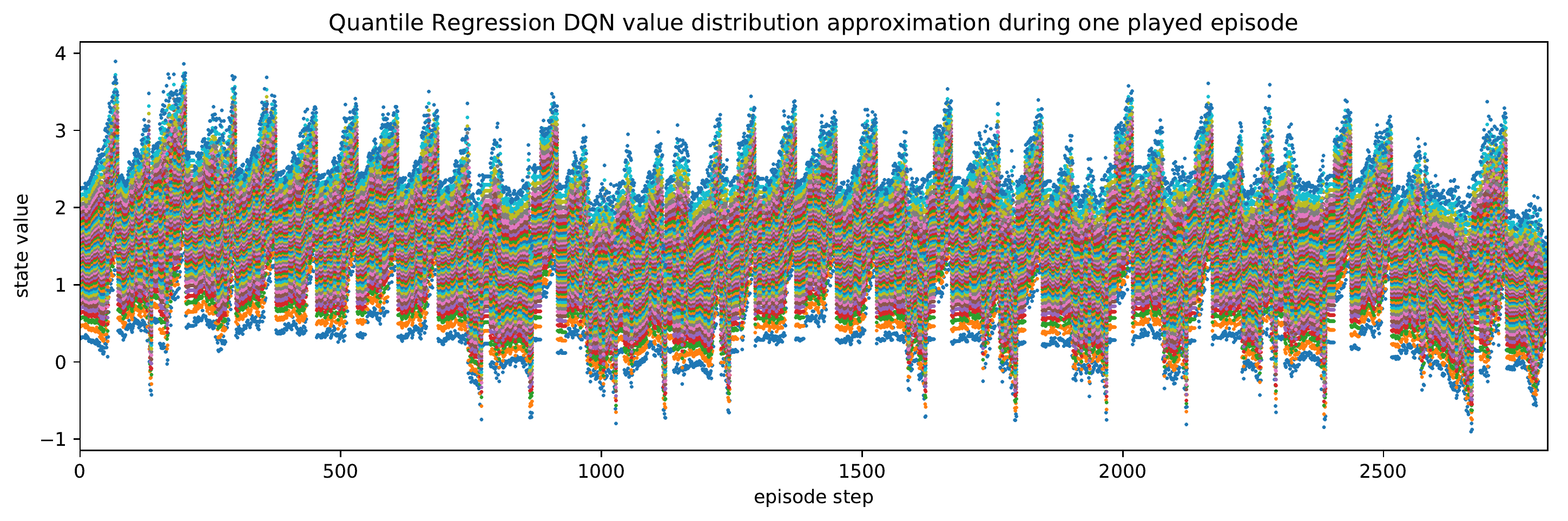}
    \caption{Quantile Regression DQN value distribution prediction during one episode of Pong.}
    \label{QRplayvd}
\end{figure}

\begin{figure}[h]
    \centering
    \includegraphics[width=\textwidth]{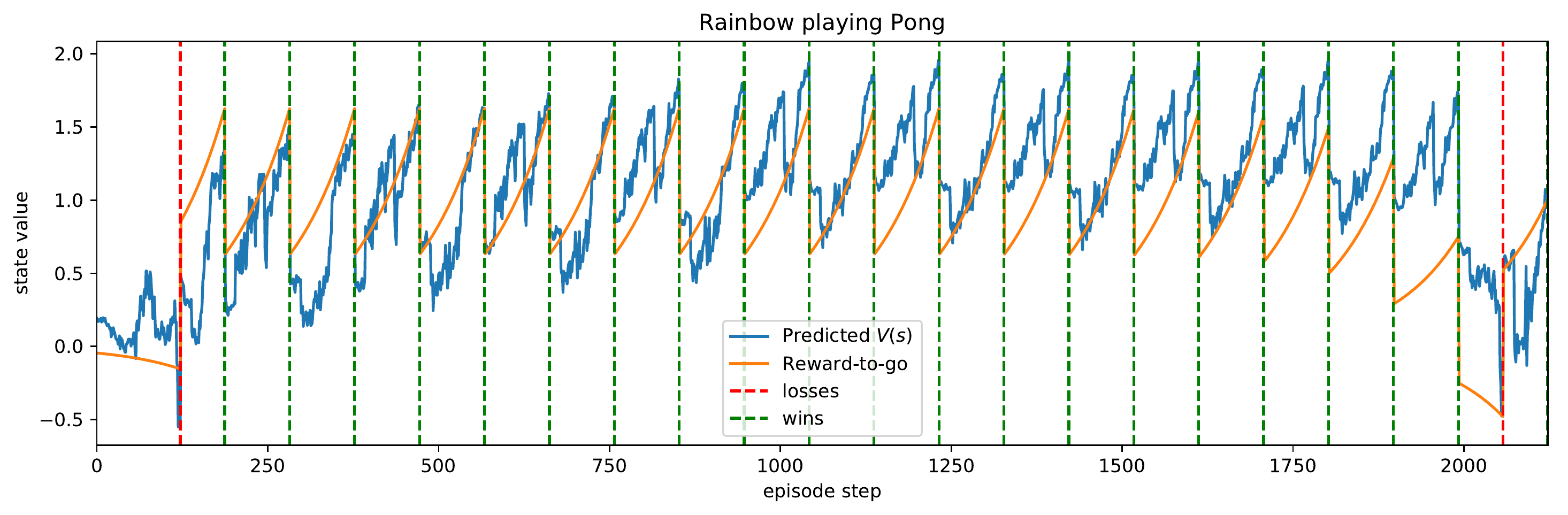}
    \caption{Rainbow playing one episode of Pong (exploration turned off, i.e. all noise samples are zero).}
    \label{Rainbowplay}
\end{figure}

\begin{figure}[h]
    \centering
    \includegraphics[width=\textwidth]{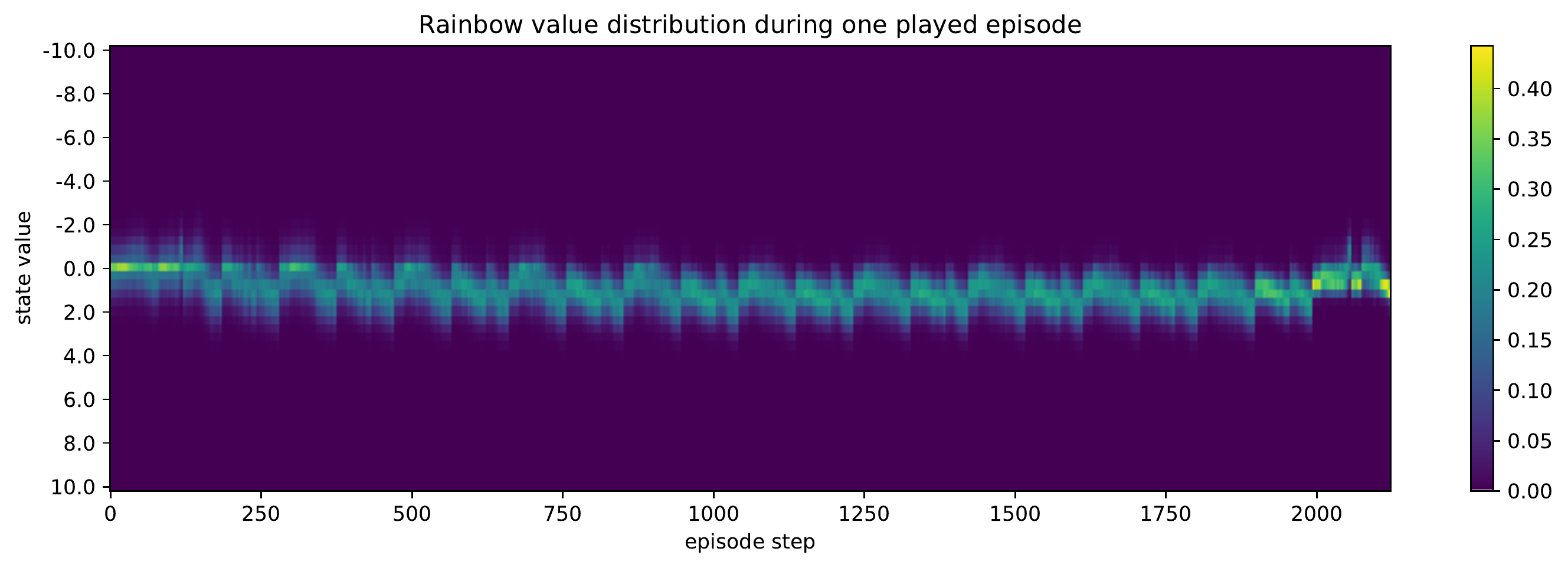}
    \caption{Rainbow value distribution prediction during one episode of Pong (exploration turned off, i.e. all noise samples are zero).}
    \label{Rainbowplayvd}
\end{figure}

\begin{figure}[h]
    \centering
    \includegraphics[width=\textwidth]{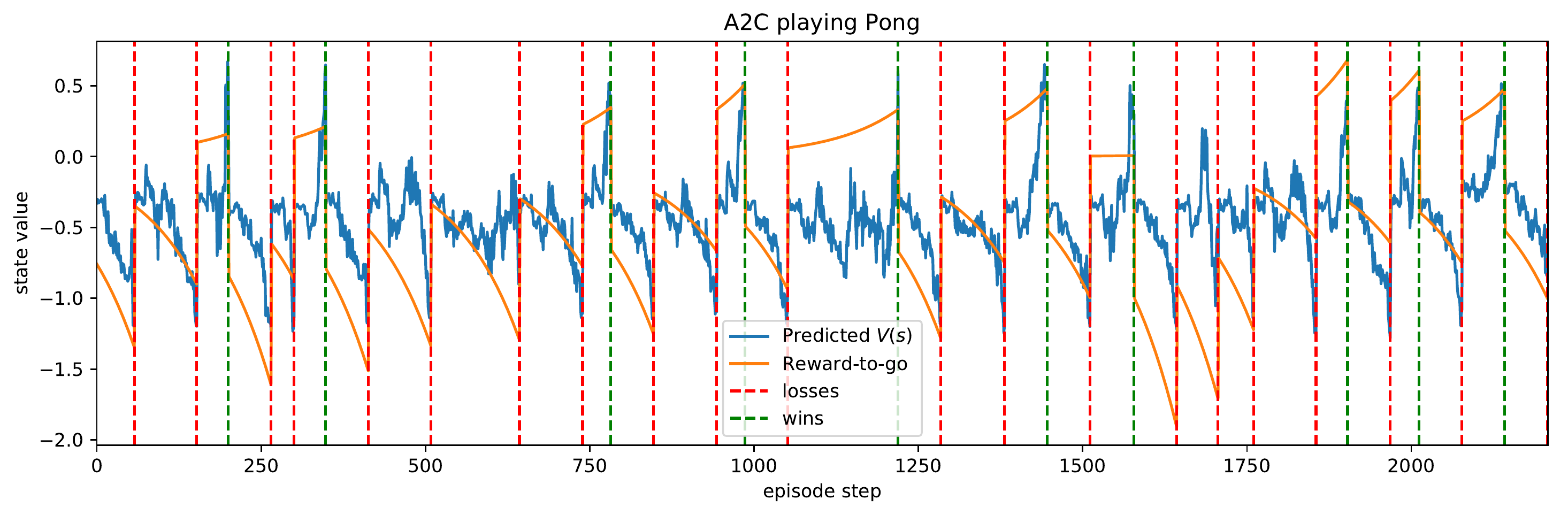}
    \caption{A2C playing one episode of Pong.}
    \label{A2Cplay}
\end{figure}

\begin{figure}[h]
    \centering
    \includegraphics[width=\textwidth]{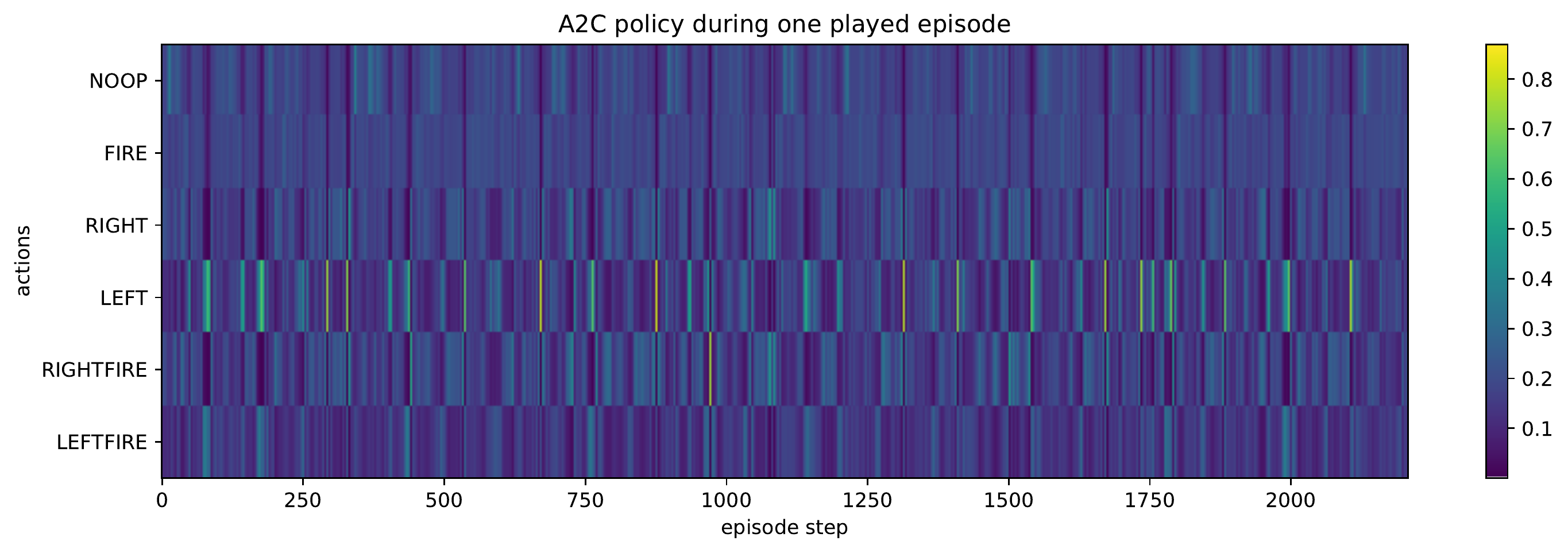}
    \caption{A2C policy distribution during one episode of Pong.}
    \label{A2Cplaypolicy}
\end{figure}

\begin{figure}[h]
    \centering
    \includegraphics[width=\textwidth]{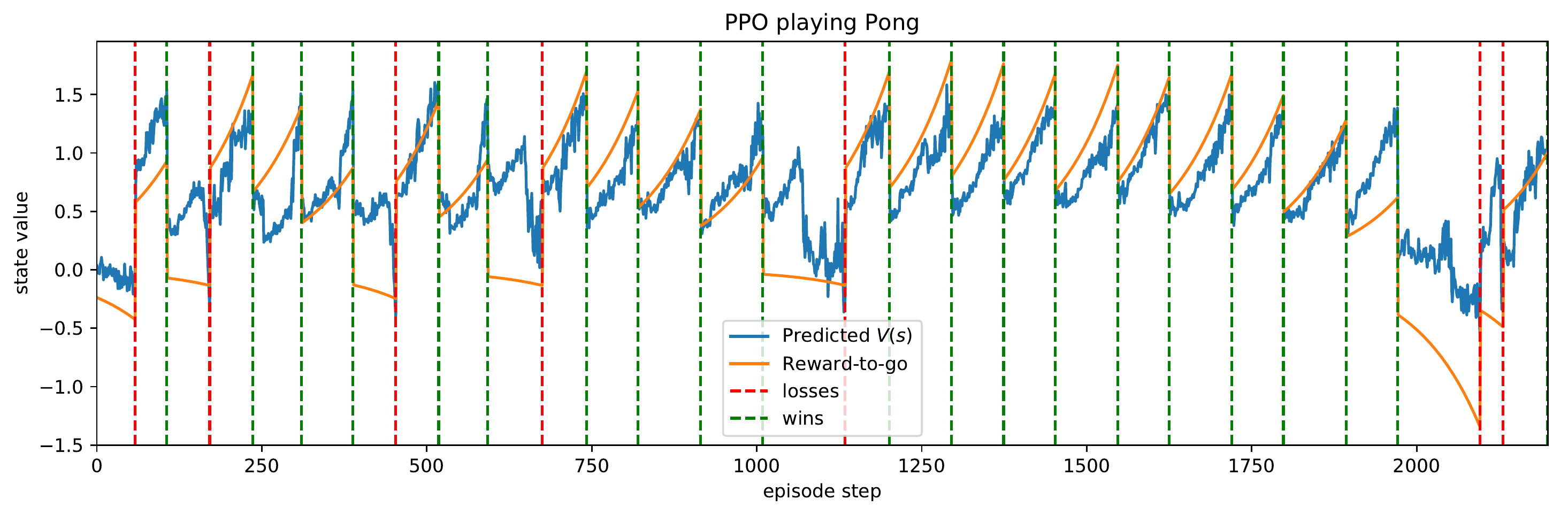}
    \caption{PPO playing one episode of Pong.}
    \label{PPOplay}
\end{figure}

\begin{figure}[h]
    \centering
    \includegraphics[width=\textwidth]{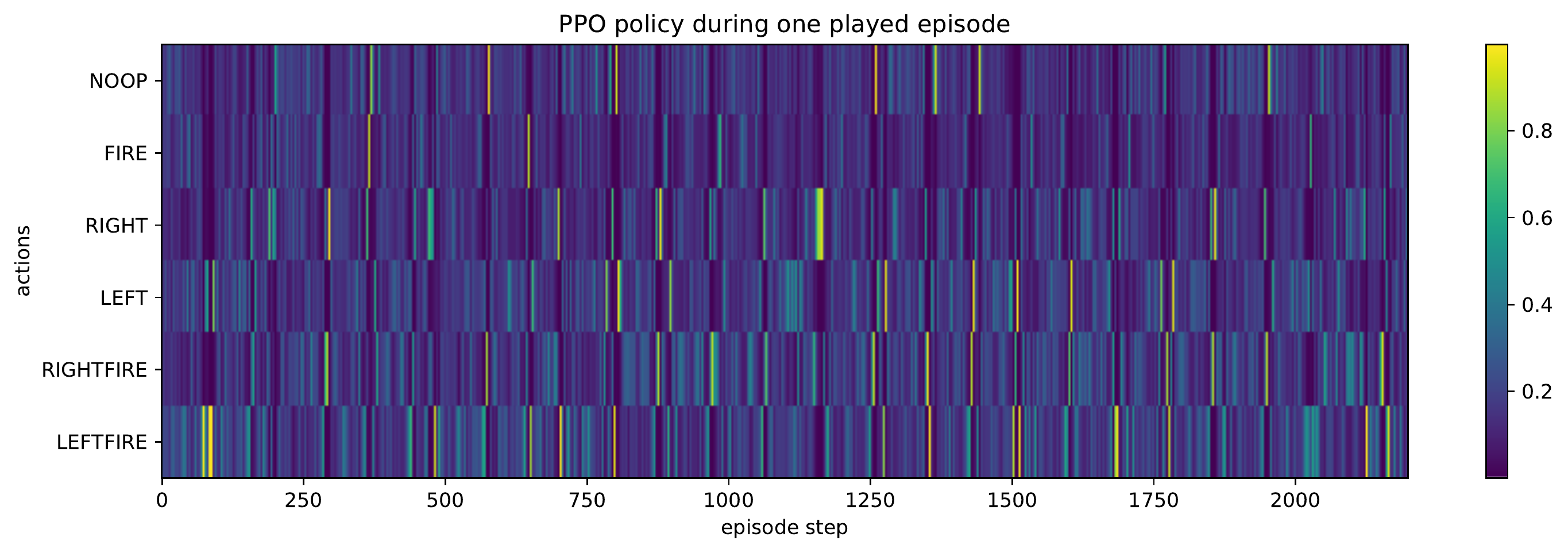}
    \caption{PPO policy distribution during one episode of Pong.}
    \label{PPOplaypolicy}
\end{figure}
\end{appendices}

\end{document}